\documentclass[logo]{googledeepmind}

\usepackage{subcaption}
\usepackage[capitalize,noabbrev]{cleveref}

\usepackage[authoryear, sort&compress, round]{natbib}
\title{Bayesian Teaching Enables Probabilistic Reasoning in Large Language Models}

\correspondingauthor{linluqiu@mit.edu, svansteenkiste@google.com, linzen@google.com}

\author[1]{Linlu Qiu}
\author[2]{Fei Sha}
\author[3,4,5]{Kelsey Allen}
\author[1]{Yoon Kim}
\author[6,7]{Tal Linzen}
\author[6]{Sjoerd van Steenkiste}

\affil[1]{MIT}
\affil[2]{Meta}
\affil[3]{Google DeepMind}
\affil[4]{University of British Columbia}
\affil[5]{Vector Institute}
\affil[6]{Google Research}
\affil[7]{New York University}

\begin{abstract}
Large language models (LLMs) are increasingly used as agents that interact with users and with the world. To do so successfully, LLMs must construct representations of the world and form probabilistic beliefs about them. To provide personalized recommendations, for example, the LLM needs to infer a user's preferences from their behavior over multiple interactions. The Bayesian inference framework lays out the optimal way for an agent to update its beliefs as it receives new information. We first show that LLMs fall far short of the standard defined by the Bayesian framework. We then show that by teaching LLMs to mimic the predictions of the normative Bayesian model, we can dramatically improve their ability to update their beliefs; this ability generalizes to new tasks. We conclude that LLMs can effectively learn reasoning skills from examples and generalize those skills to new domains.
\end{abstract}

\usepackage{xcolor}
\usepackage{xspace}
\usepackage{amssymb}
\usepackage{amsmath}
\usepackage{bbm}
\usepackage{booktabs}
\usepackage{multirow}
\usepackage{graphics}
\usepackage{ulem}
\usepackage{graphicx}
\usepackage{array}
\usepackage{tabularx}
\usepackage{makecell}
\usepackage{wrapfig}
\usepackage{adjustbox}
\usepackage{subcaption}
\usepackage{courier}
\usepackage{hyperref}       %
\usepackage{url}            %
\usepackage{booktabs}       %
\usepackage{amsfonts}       %
\usepackage{nicefrac}       %
\usepackage{microtype}      %
\usepackage{cleveref}
\usepackage{longtable}
\usepackage{soul}
\usepackage{float}

\usepackage{amsmath,amsfonts,bm}

\definecolor{boxbg}{rgb}{0.85,0.85,0.85}
\definecolor{orange}{rgb}{1,0.5,0}
\definecolor{mdred}{rgb}{0.7,0,0}
\definecolor{mdgreen}{rgb}{0.05,0.6,0.05}
\definecolor{mdblue}{rgb}{0,0,0.7}
\definecolor{dkblue}{rgb}{0,0,0.5}
\definecolor{dkgray}{rgb}{0.3,0.3,0.3}
\definecolor{slate}{rgb}{0.25,0.25,0.4}
\definecolor{gray}{rgb}{0.5,0.5,0.5}
\definecolor{ltgray}{rgb}{0.7,0.7,0.7}
\definecolor{purple}{rgb}{0.7,0,1.0}
\definecolor{lavender}{rgb}{0.65,0.55,1.0}

\newcolumntype{C}[1]{>{\centering\arraybackslash}p{#1}}

\renewcommand{\citet}{\cite}

\begin{document}

\maketitle

\section{Introduction}

Humans interact with the world based on our beliefs about it. To effectively support decision making, our beliefs need to correspond to the structure of the world as much as possible; in other words, our beliefs need to be supported by appropriate ``world models'' \citep{johnson1980mental,ha2018recurrent,lecun2022path,wong2023word}. We typically do not have perfect knowledge about the outside world; to the extent that we are uncertain about our environment, our beliefs need to be probabilistic, reflecting this uncertainty. And for these beliefs to remain relevant as the world changes, or as new information about the world becomes available, we need to update our beliefs to reflect the new information. The framework of Bayesian inference describes the normative way in which new information should trigger a change in one's beliefs so as to maximize the effectiveness of these beliefs as a foundation for acting in the world \citep{chater2006probabilistic}. %
The Bayesian framework has informed a substantial body of work in cognitive science, which has identified both areas where humans act as the framework predicts, as well as deviations from it~\citep{griffiths2024bayesian,jern2017people,josh2011how,xu2007word,baker2011bayesian,tenenbaum2006theory,chaterm06,griffithsst07,chaigneau2025breaking,rehder2018beyond,rottman2016people,sloman2015causality}.

In the last few years, artificial intelligence systems based on large language models (LLMs) have become dramatically more capable than in the past~\citep{geminiteam2024gemini15unlockingmultimodal, achiam2023gpt, claude, gemmateam2024gemma2, touvron2023llama, guo2025deepseek}. Far outgrowing their original motivation---as methods to estimate the probabilities of different word sequences---these systems are now being used for applications where they interact with users and with the outside world.
As with humans, for the LLMs' interactions with users to be effective, the LLMs' beliefs need to reflect their experience with the user and to be continuously updated as more information becomes available. Here, we ask: do LLMs act as if they have probabilistic beliefs that are updated as expected from normative Bayesian inference? To the extent that the LLMs' behavior deviates from the normative Bayesian strategy, how can we minimize these deviations? 

We begin to study these questions using a simple controlled setting: a flight recommendation task \citep{lin2022inferring}, illustrated in Fig.~\ref{fig:fig1_task}. This task involves multiple rounds of interactions between a simulated user and an LLM, where the LLM is acting as a flight booking assistant. In each round, the assistant is given a small number of flight options, and is expected to recommend one of them to the user, based on the user's preferences. The user's preferences are not directly communicated to the LLM: it only observes the choices the user makes among the flight options. To make optimal recommendations, then, the LLM must construct an implicit model of the factors that shape the user's preferences, and must reason probabilistically about those factors as it learns about the user's choices across multiple sets of flight options.

We compare the LLMs' behavior to that of a model that follows the normative Bayesian strategy, which we refer to as the \textit{Bayesian Assistant}. This model maintains a probability distribution that reflects its beliefs about the user's preferences, and uses Bayes' rule to update this distribution as new information about the user's choices becomes available. Unlike many real-life scenarios, where it is difficult to specify and  implement the Bayesian strategy computationally, in this controlled setting this strategy can be computed exactly, allowing us to precisely estimate the extent to which LLMs deviate from it. 

We use this framework to evaluate a range of LLMs and find that they all perform significantly worse than the normative Bayesian Assistant (Fig.~\ref{fig:fig2_eval_results}). Most importantly, in contrast to the Bayesian Assistant, which gradually improves its recommendations as it receives additional information about the user's choices, LLMs' performance often plateaus after a single interaction, pointing to a limited ability to adapt to new information.

We then introduce \textit{Bayesian teaching}, a strategy to teach an LLM to approximate Bayesian reasoning. We provide the LLM with examples of interactions between the user and the Bayesian Assistant, and have the LLM mimic those interactions. We find that, by leading the LLMs to gradually adapt to the user over the course of the interactions, this method substantially improves the LLMs' performance on the flight recommendation task. Crucially, teaching the LLMs to mimic the Bayesian Assistant in one task allows them to generalize to other tasks that similarly require making decisions under uncertainty; those include not only different variants of the flight recommendation task, but also a related hotel recommendation task, as well as a web shopping task with real-world products (Fig.~\ref{fig:fig1_task}), a much more complex task for which it is difficult to specify and implement a fully Bayesian model. 

Notably, while the Bayesian Assistant often makes incorrect predictions as it reasons under uncertainty, especially in the early rounds of interaction, we find that it is a more effective teacher than a teacher that directly provides the LLMs with users' choices (which we refer to as an \textit{oracle teacher}); in other words, the Bayesian model's educated guesses make for a stronger learning signal than the correct answers. Overall, we conclude that through observing the Bayesian Assistant perform a particular task, the LLMs are able to approximate transferable probabilistic reasoning skills.

To summarize our contributions: we first identify significant limitations of off-the-shelf LLMs in tasks that require forming and updating probabilistic beliefs. We then demonstrate that, by having the LLMs mimic an normative Bayesian model, we can teach them effectively to approximate probabilistic belief updates, and show that these  skills can generalize to new environments. These findings suggest that LLMs can be used in interactive settings where information is provided gradually, including complex application domains where implementing an exact Bayesian model is difficult. More generally, our results highlight a unique strength of deep learning models such as LLMs: they can learn to mimic a symbolic model and generalize its strategy to domains that are too complex to specify in a classic symbolic model.

\section{Evaluating Belief Updates via Flight Recommendations}

\begin{figure*}[t!]
\centering
\includegraphics[width=\linewidth]{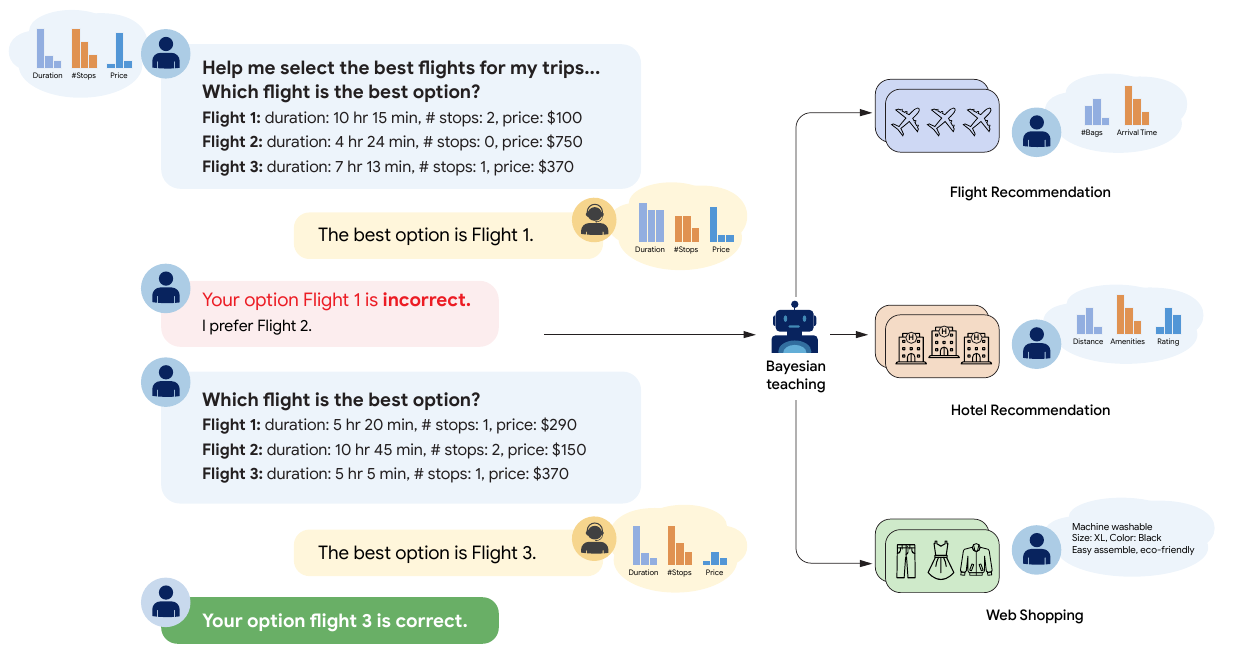}
\caption{\textbf{Evaluating and improving LLMs' probabilistic belief updates.} The flight recommendation task (left) involves multi-round interactions between a user and a flight booking assistant. In each round, the assistant is asked to recommend to the user one of three available flight options. The assistant is then shown the flight that was in fact chosen by the user (based on the user's reward function, which characterizes the user's preferences). To make good recommendations, the assistant needs to infer the user's preferences from the user's choices. To teach the LLM to reason probabilistically, we fine-tune the LLM on interactions between users and a Bayesian Assistant, which represents the normative way to update beliefs about the user's preferences. We then evaluate the fine-tuned model on the flight recommendation task as well as two new tasks (right).}
\label{fig:fig1_task}
\end{figure*}

We first describe the simplified flight recommendation task, derived that we use to evaluate the LLMs \citep{lin2022inferring}. In this task, we have the LLMs interact with a simulated user for five rounds. In each round, three flight options are presented to both the user and the assistant. Each flight is defined by a departure time, a duration, a number of stops, and a cost (see Fig.~\ref{fig:fig1_task}).
Each simulated user is characterized by a set of preferences: for each feature, they can have a strong or weak preference for high or low values of the feature (e.g., they may prefer longer or shorter flights), or no preference regarding this feature. We refer to this set of preferences as the user's \textit{reward function}. We have 624 possible users in total (see Appendix Section~\ref{app:methods}). These preferences, which determine the flights that the user chooses, are not directly revealed to the assistant. 
The goal of the assistant is to recommend the flight that matches the user's choice. At the end of each round, the user indicates to the assistant whether or not it chose correctly, and provides it with the correct answer. 

After each round, we evaluate the accuracy of the assistant's recommendations for 100 new sets of three flights that differ from the ones on which the assistant has received feedback. We do not provide any feedback to the assistant for these new flight option sets (see Appendix Fig.~\ref{fig:app_figa1_eval} for the evaluation workflow).

\subsection{The Bayesian Assistant}

Because the users' preferences are only revealed gradually, through their choices among flight options, we cannot expect the LLMs to reach perfect accuracy immediately after a single round of interaction. As an upper bound on the LLMs' performance, we define a Bayesian Assistant, which implements the strategy that optimally takes into account the evidence about the user's preferences that accumulates over rounds of interaction. This entails maintaining uncertainty about those preferences when the evidence is partial: instead of committing to a single most likely reward function, which could turn out to be incorrect in future rounds, the assistant maintains a probability distribution over possible reward functions. After each round, the Bayesian Assistant updates its distribution over reward functions using Bayes' rule: the probability of each reward function after the round (the posterior) is computed based on its probability before the round (the prior) and whether or not it was compatible with the user's choice (the likelihood). This normative model represents the best performance that we can possibly expect from any system. Because the number of possible reward functions is small, we are able to perform exact Bayesian inference (see Appendix Section~\ref{app:methods}).

This method requires us to define the Bayesian Assistant's initial prior distribution, that is, its probabilistic assumptions about which user preferences are more likely, in advance of any interaction with the user. We use an uninformed prior, where all possible sets of user preferences are equally likely (for experiments with alternative priors, see Appendix Section~\ref{sec:app_c4_whybayesian}). 

\begin{figure*}[t!]
\centering
\includegraphics[width=\linewidth]{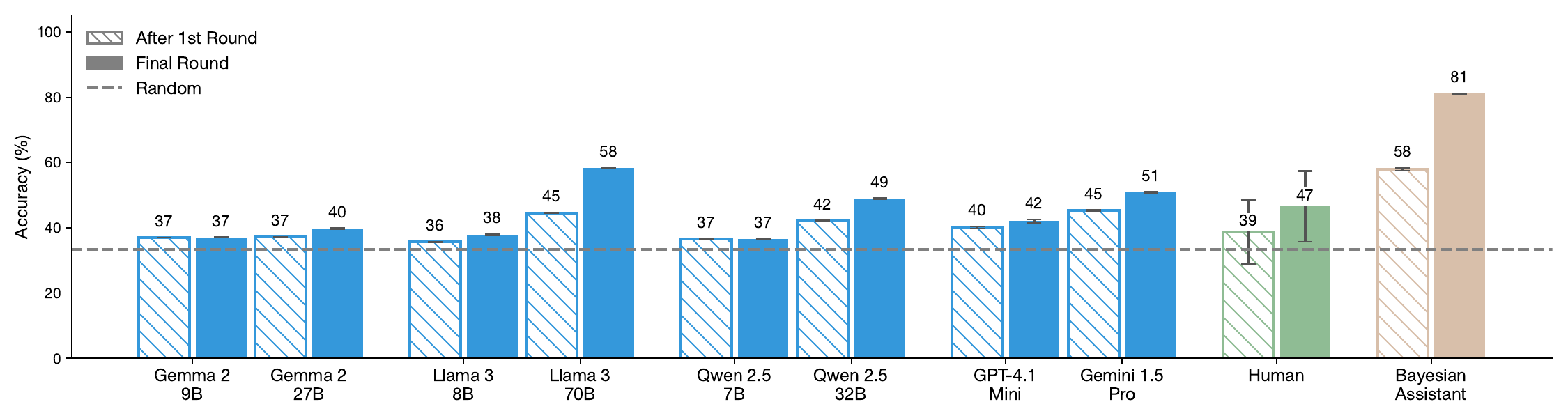}
\caption{\textbf{LLMs show limited or no improvement over multiple interactions with the user.} We show accuracy after the first round and final (fifth) round. We compare off-the-shelf LLMs from different model families to human participants and the Bayesian Assistant. For human participants, we only evaluate on a subset of 48 out of our 624 simulated users. The  LLMs perform considerably worse than the Bayesian Assistant. Human participants demonstrate a larger improvement than most LLMs as they receive more information, but they still fall short of the accuracy that characterizes the normative Bayesian strategy. For the human study, the error bars show the averaged standard error across participants; for models, they show the standard error across the three sets of interactions with each of the 624 users.}
\label{fig:fig2_eval_results}
\end{figure*}

\subsection{LLMs Show Limited Evidence of Belief Updating}

The LLMs we evaluate, like most contemporary LLMs, are first trained to predict upcoming words in a large collection of texts (``pre-training''), and are then specialized to follow user instructions provided in natural language (``instruction-tuning'') \citep{sanh2022, wei2021finetuned}. Most commercially available models are closed-weights: we can query them but we cannot access their parameters. We evaluate two such closed-weights models, Gemini 1.5 Pro~\citep{geminiteam2024gemini15unlockingmultimodal} and GPT-4.1 Mini~\citep{openai2025gpt41}, which were among the state-of-the-art LLMs at the time of writing \citep{chiang2024chatbot}. We also evaluate the following open-weights models: Gemma~2 (9B and 27B parameters)~\citep{gemmateam2024gemma2}, Llama~3 (8B and 70B parameters)~\citep{dubey2024llama3herdmodels}, and Qwen~2.5 (7B and 32B parameters)~\citep{yang2024qwen2}. We chose those models because their performance was quite competitive, and their weights are openly available, which makes it possible to perform fine-tuning (see the next section).
We provide these LLMs with English instructions explaining how to act as a flight booking assistant (see Fig.~\ref{fig:fig1_task} for an example, and Appendix Table~\ref{tab:prompt_interaction} for a detailed interaction).

We show results in Fig.~\ref{fig:fig2_eval_results}.
Overall, the accuracy of the LLMs after the five rounds of interaction is considerably lower than that of the Bayesian Assistant, and
most of the models show little improvement after the first round of interaction  (Fig.~\ref{fig:fig2_eval_results} shows results after the first and fifth round; for results after each of the five rounds, see Appendix Fig.~\ref{fig:app_figg18_eval_details}). For an exploration of how the models' performance varies across users' possible reward functions, see Appendix Section~\ref{sec:app_c2_reward_variability}. 

A range of follow-up experiments failed to produce meaningful improvement in the LLMs' behavior (for details, see Appendix Section~\ref{sec:app_b1_eval_ablations}). Those include experiments with ``chain-of-thought prompting'' \citep{wei2022chain, nye2021show, kojima2022large}, that is, instructions that are meant to encourage the LLM to reason more explicitly (Appendix Fig.~\ref{fig:app_figb3a_prompt_method}); an experiment with alternative, purely numerical representations of the flight options that we hypothesized might be easier for the LLMs to parse than the verbal ones we used for our main experiments (Appendix Fig.~\ref{fig:app_figb3b_representation}); a setting where we have 30 instead of five rounds of interaction (Appendix Fig.~\ref{fig:app_figb3c_rounds}); and experiments with models that are only pre-trained to predict upcoming words in texts, without subsequent training to follow user instructions (Appendix Fig.~\ref{fig:app_figb3f_base}).

We also had human participants act as the assistant to a subset of 48 simulated users (see Appendix Section~\ref{app:methods} and Appendix Section~\ref{sec:app_e1_human_as_assistant} for details). The human participants made recommendations for five rounds and showed a significant improvement between round 1 and 5 (p = 0.002, logistic mixed-effects model). In terms of accuracy, they perform better than small LLMs and slightly worse than larger LLMs (see Appendix Fig.~\ref{fig:app_figg18_eval_details} for performance over rounds). That being said, like all LLMs, humans also fall substantially short of the accuracy expected from the normative Bayesian strategy.

\section{Teaching LLMs to Approximate Bayesian Reasoning}
\label{sec:ft_results}

\begin{figure*}[t!]
\centering
\includegraphics[width=\linewidth]{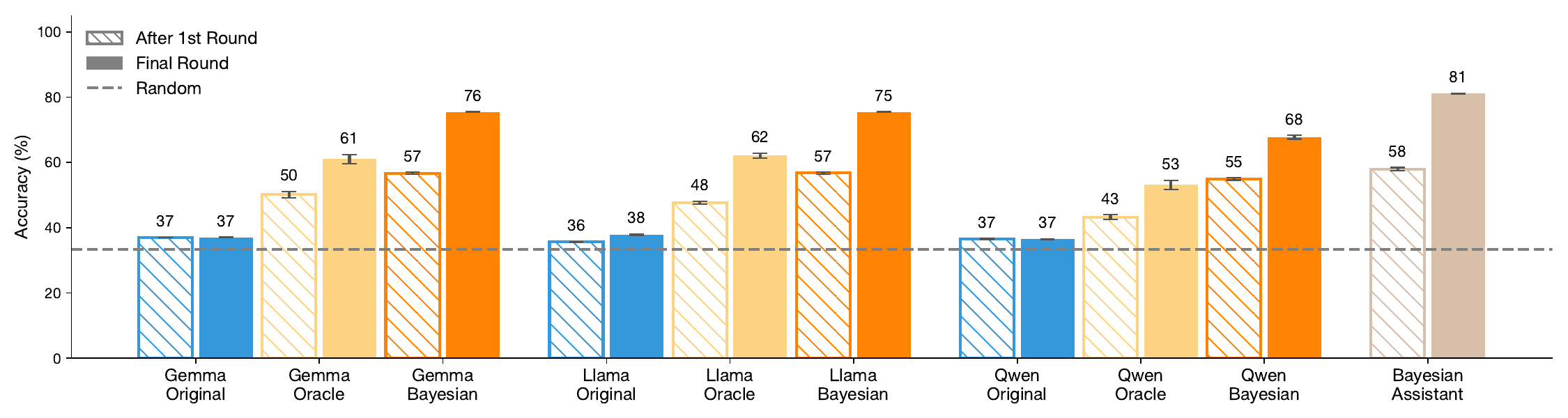}
\caption{\textbf{Supervised fine-tuning teaches LLMs to approximate probabilistic inference.} We show accuracy after the first round and final (fifth) round across different assistants. We compare the original LLMs, LLMs fine-tuned on user interactions with the Bayesian Assistant, and LLMs fine-tuned on user interactions with an oracle, which always provides the correct answer. Both types of fine-tuning significantly improve LLMs' performance, and Bayesian teaching is consistently more effectively than oracle teaching. Error bars show the standard error across three random seeds (and three training runs). All results are statistical significant, $p < 0.001$ (see Appendix Section~\ref{sec:app_f_stats}).}
\label{fig:fig3_ft_results}
\end{figure*}

We next describe the supervised fine-tuning technique we use to teach the LLM to mimic the normative Bayesian model; we show that this method substantially improves the LLM's ability to update its beliefs correctly.

From a technical perspective, supervised fine-tuning is similar to the method used to train most LLMs in the first place. The model is provided with the first words of a text and is trained to predict the upcoming word. After each example, the LLM's weights are adjusted to increase the likelihood of a correct prediction if the same example is observed again. The main difference is that while in the first phase of training the texts are typically drawn from the Internet or similar resources, in the supervised fine-tuning phase the texts are constructed in a targeted way (automatically or by human writers) so as to teach the LLM particular skills \citep{sanh2022, wei2021finetuned}; to improve arithmetic skills, for example, the model may be given the text ``the output of $1+1=\ldots$ is~$2$''. We apply supervised fine-tuning to the three medium-sized open-weights models (Gemma~2~9B, Llama~3~8B, and Qwen~2.5~7B); we do not attempt to fine-tune the larger models from these families due to computational constraints. We update all of the models' weights in fine-tuning (in Appendix Section~\ref{sec:app_b2_ft_ablations}, we show that a different training objective, Direct Preference Optimization \citep{rafailov2023direct}, produces similar results, as does a computationally cheaper fine-tuning method, LoRA \citep{hu2021lora}, which only updates a subset of the model's weights).

We explore two strategies to create supervised fine-tuning data. For both strategies, we construct 10 five-round interactions per user. These interactions follow the same format as described above (Appendix Table~\ref{tab:prompt_interaction}).
In the first strategy, which we refer to as \textit{oracle teaching}, we provide the LLM with interactions between simulated users and an ``oracle'' assistant that has perfect knowledge of the user's preferences, and as such always recommends the option that is identical to the user's choices. 

The second strategy, which we call \textit{Bayesian teaching}, provides the LLM with interactions between the user and the Bayesian Assistant. In this setting, the assistant will often choose flights that do not match the user's preferred choice, especially in early rounds where it has considerable uncertainty about the user's preferences. We hypothesize that despite this fact mimicking the Bayesian Assistant's best guesses would teach the LLM to maintain uncertainty and update its beliefs more effectively than the first strategy where the LLM is trained on the correct choices. This approach can be seen as a form of distillation, where a model is trained by learning to mimic another system \citep{hinton2015distilling, kim2016sequence, deng2023implicit, wang2023scott, li2023symbolic, jung2024impossible, yu2distilling, chen2024learning}. We use a uniform prior for the Bayesian Assistant that produces the supervised fine-tuning data. Other priors perform similarly (see Appendix Fig.~\ref{fig:app_figc10b_bayesian_prior}).

\subsection{Fine-Tuning Teaches LLMs to Adapt to Users}
Both supervised fine-tuning strategies, oracle teaching and Bayesian teaching, significantly improve the LLMs' performance on the flight recommendation task (Fig.~\ref{fig:fig3_ft_results}). Crucially, after fine-tuning, the LLMs' performance gradually improves as more information becomes available; this contrasts with the original LLMs, which plateaued after the first round (see the substantial performance improvement between the first and last round in Fig.~\ref{fig:fig3_ft_results}; for detailed results for each round, see Appendix Fig.~\ref{fig:app_figg19_ft_flight_details}). While there is still a performance gap between the fine-tuned LLMs and the normative Bayesian Assistant, this gap is much narrower than for the original LLMs. All three medium-sized LLMs, which before fine-tuning performed worse than either stronger models or our human participants, markedly outperform them after fine-tuning.

We find that Bayesian teaching leads to higher accuracy and less variability across repetitions of the experiment than oracle teaching (Fig.~\ref{fig:fig3_ft_results}). Bayesian teaching also successfully makes the LLM more Bayesian: the Bayesian-tuned LLMs' predictions agree with those of the Bayesian Assistant around 80\% of the time, significantly more often than do the predictions of the original LLMs and oracle-tuned LLMs (Fig.~\ref{fig:fig4_agreement}). In Appendix Section~\ref{sec:app_c4_whybayesian}, we show that the effectiveness of Bayesian teaching cannot be explained by two potential confounds, and conclude that the effectiveness of this method is in fact due to the Bayesian signal it provides.

The amount of information that can be gained from the user's choice for a particular option set varies from one set to another. For example, a choice between two flight options that differ in exactly one feature provides direct evidence for the user's preference for that feature; such a choice could be more informative about the user's preferences than the choice between options that differ along multiple dimensions. We expect a model with more sophisticated probabilistic skills to show greater sensitivity to this factor. Do our fine-tuned models show such sensitivity? Focusing on the Gemma models, we find that Gemma Original does not show sensitivity to option set informativity, but both fine-tuned versions of Gemma do, with Gemma Bayesian displaying considerably more sensitivity than Gemma Oracle (Appendix Section~\ref{sec:app_d_informativeness}). 

Can the fine-tuned models accurately verbalize their beliefs? To address this question, we ask the LLMs explicitly for their beliefs about the user's preferences---we have the simulated user ask them, for example, ``on a scale of 1 to 5, what is my preference for price?''. We then test for the accuracy of these verbalized beliefs by deriving flight recommendations from those beliefs, using the same decision procedure we use with the Bayesian Assistant. We find that this approach generally performs better the approach we have used so far where we directly ask for the LLMs' recommendations; that predictions based on the fine-tuned LLMs' verbalized beliefs are substantially more accurate than those based on the original LLMs' verbalized beliefs; and that the Bayesian-tuned LLMs produce more accurate beliefs than either the original LLMs or oracle-tuned ones (for additional details, see Appendix Section~\ref{sec:app_a_verbalize_beliefs}).

\begin{figure*}[t!]
\centering
\includegraphics[width=\linewidth]{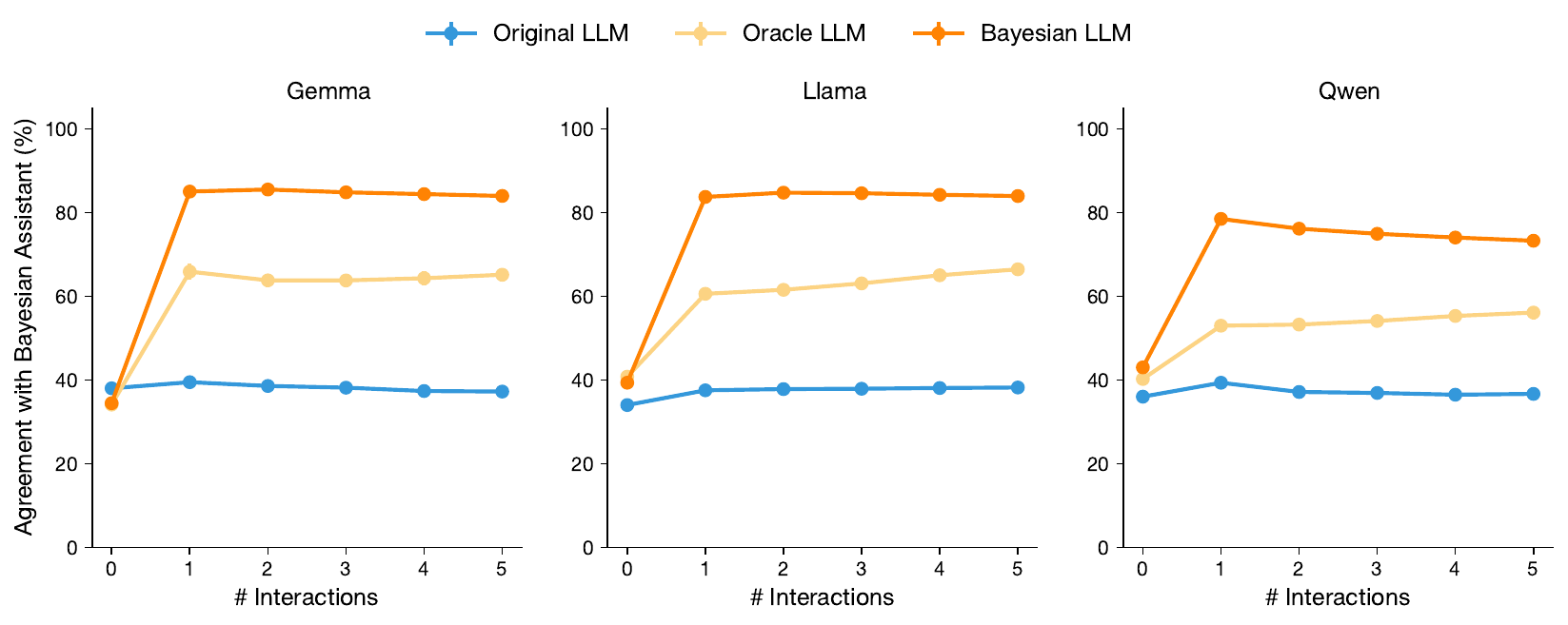}
\caption{\textbf{Fine-tuned LLMs agree more with the Bayesian Assistant.} We show agreement between the LLMs and the Bayesian Assistant, measured by the proportion of trials where the LLMs makes the same predictions as the Bayesian Assistant. Fine-tuning on the Bayesian Assistant's predictions makes the LLMs more Bayesian, with the Bayesian versions of each LLM achieving the highest agreement with the Bayesian Assistant. Error bars (too small to be visible in plot) show standard errors across three random seeds (and three training runs).}
\label{fig:fig4_agreement}
\end{figure*}

\subsection{Fine-Tuned LLMs Generalize to New Tasks}

\begin{figure*}[ht!]
    \includegraphics[width=\textwidth]{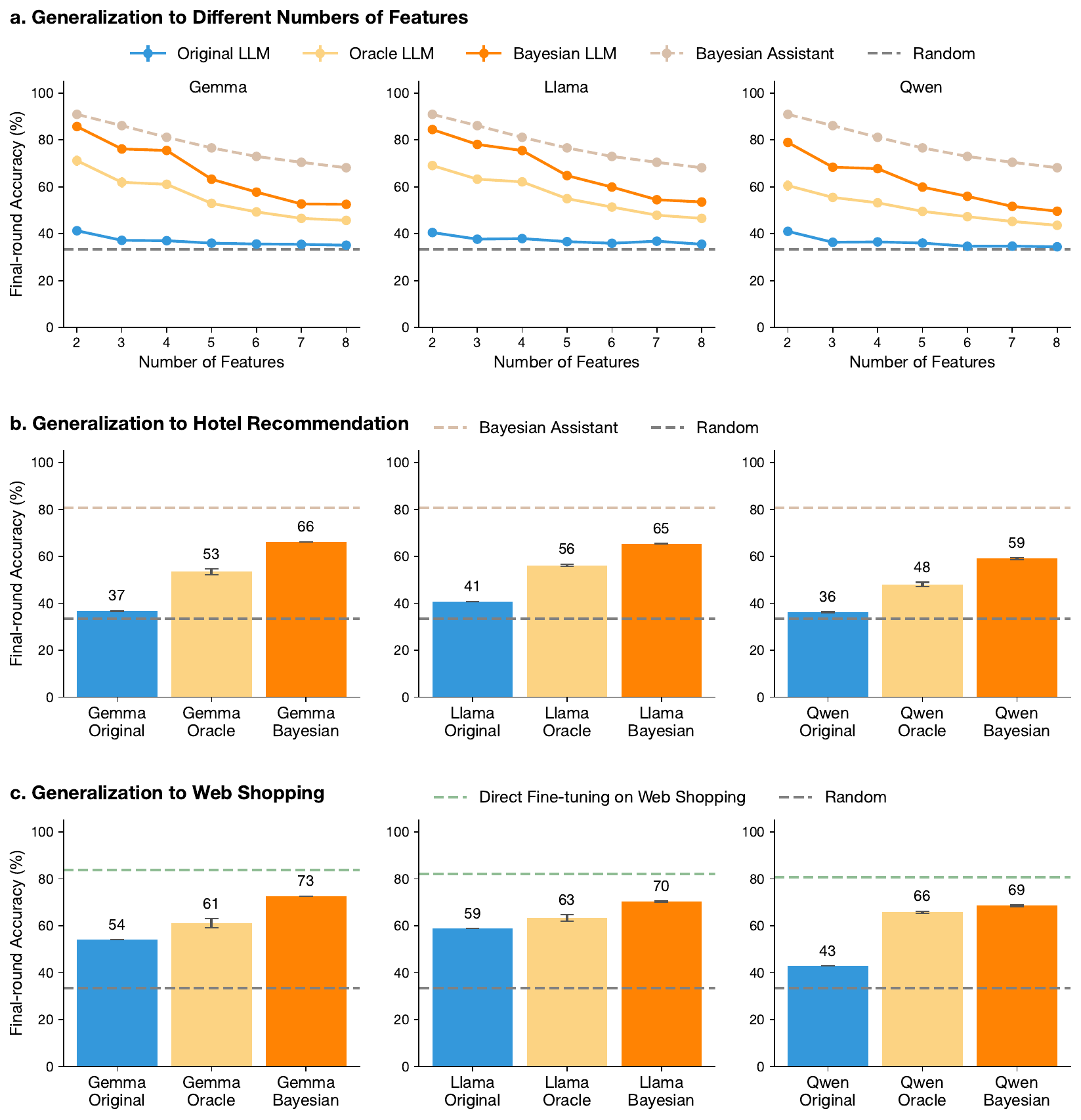}
    \begin{subfigure}[b]{0pt}
        \phantomsubcaption
        \label{fig:fig5a}
    \end{subfigure}
    \begin{subfigure}[b]{0pt}
        \phantomsubcaption
        \label{fig:fig5b}
    \end{subfigure}
    \begin{subfigure}[b]{0pt}
        \phantomsubcaption
        \label{fig:fig5c}
    \end{subfigure}
\caption{\textbf{Bayesian teaching generalizes outside the task used for fine-tuning.} \textbf{(a)} Final-round accuracy gain in fine-tuned models compared to the original LLM when varying task complexity (here the number of features is a proxy for task complexity).  \textbf{(b)} Final-round accuracy for LLMs on the hotel recommendation task, which was not seen during fine-tuning. We show the normative Bayesian Assistant's performance with brown dashed lines. \textbf{(c)} Final-round accuracy for LLMs on the web shopping domain, also unseen during fine-tuning. The green dashed line indicates the performance of the LLM when it is fine-tuned directly on web shopping data, such that no domain generalization is necessary. Error bars indicate the standard errors over three training runs (for web shopping) and additionally three random seeds (for flight recommendation and hotel recommendation).}
\label{fig:fig5_gen_results}
\end{figure*}

As a result of Bayesian teaching, the LLMs demonstrate a greatly improved ability to approximate Bayesian probabilistic inference. Is this ability specific to the particular task the models were trained on, or do the LLMs' probabilistic skills improve more broadly? To answer this question, we evaluate the fine-tuned LLMs on a set of tasks that diverge to different extents from our original flight recommendation task (see the right panel of Fig.~\ref{fig:fig1_task} for an overview). All tasks require the LLMs to infer the user's preferences from the user's choices over multiple interactions. Overall, as we show in the rest of this section, we find that fine-tuned LLMs show considerable generalization to new settings, and that, as before, Bayesian teaching is more effective than oracle teaching.

We first test the LLMs on variants of the flight recommendation task with different numbers of features: whereas in the interactions provided during fine-tuning, flights were characterized by four features, in this evaluation setting flights are described by between two and eight features. This requires the LLM to generalize to features that were not included in fine-tuning (e.g., the number of checked bags). In this setting, we find that both types of fine-tuning lead to large improvement in accuracy compared to the original LLMs. We also find that Bayesian teaching is considerably more effective than oracle teaching, as before (Fig.~\ref{fig:fig5a}). We note that as the number of features increases, the space of possible reward functions grows exponentially, and the task becomes inherently more difficult, even for the Bayesian Assistant. Despite this fact, for both fine-tuning methods, performance relative to the upper bound defined by the Bayesian Assistant drops off only moderately as the number of features increases.

The generalization experiments we have discussed so far focused on variants of the flight recommendation task. We next evaluate whether the LLMs can generalize the probabilistic skills they acquire through fine-tuning and apply them to other domains. We consider two such domains: hotel recommendations and web shopping. The hotel recommendation task is a synthetic task whose structure is similar to that of the flight recommendation task presented in fine-tuning. Here, each hotel is defined by four features: distance to downtown, price, rating, and amenities (for an example, see Appendix Table~\ref{tab:prompt_hotel}). 

The web shopping task uses real-world products from a simulated environment \citep{yao2022webshop}, and differs much more substantially from the fine-tuning task than does the hotel recommendation task.  It is difficult to construct a Bayesian Assistant for more natural scenarios like the web shopping task, where the space of user preferences is large and hard to specify formally. For this reason, successful transfer from synthetic settings like the flight recommendation task to more natural scenarios represents a particularly important application of Bayesian teaching. In the web shopping task, each user is defined by a set of randomly sampled goals that characterize the product they are interested in; for example, they might be looking for a shirt that is machine washable, or for a size XL shirt (see Appendix Table~\ref{tab:app_tabc1_webshop_example} for examples). As in the flight domain, the assistant interacts with the user for multiple rounds. In each round, a set of product options is randomly sampled from the product category (e.g., shirts), and the assistant is asked to recommend the best option.  Each product is represented by a short title along with a detailed description (see Appendix Table~\ref{tab:prompt_webshop_verbose} for an example). The user provides feedback at the end of each round, indicating whether or not the assistant's recommendation was correct. The user's preferred option is the one with the highest reward, as defined in \citet{yao2022webshop}. 
As mentioned above, it is difficult to construct a Bayesian Assistant for this task due to the large space of possible preferences. Instead, as an alternative upper bound on the transfer performance we can expect from the models fine-tuned on the flight recommendation task, we fine-tune LLMs directly on data from the shopping task.

We find that LLMs fine-tuned on the flight recommendation task generalize to both hotel recommendations and web shopping: they perform much better than the original LLMs on those tasks  (Fig.~\ref{fig:fig5b} and Fig.~\ref{fig:fig5c}).  Bayesian teaching continues to outperform oracle teaching, though the gap is smaller for web shopping than hotel recommendations. There remains a gap between the generalization performance of the LLMs fine-tuned on flight recommendations and the upper bound obtained by fine-tuning the LLMs directly on the web shopping interactions (green dashed line in Fig.~\ref{fig:fig5c}). Overall, we conclude that fine-tuning, and especially Bayesian teaching, imparts probabilistic skills that transfer substantially beyond the setting used for fine-tuning.

\subsection{Generalization to Interactions with Human Users}
The synthetically generated data we have used so far makes two simplifying assumptions: the simulated users' choices faithfully reflect the reward function that characterizes their preferences, and all reward functions are encountered equally often. In practice, these assumptions may not hold as humans' behavior could occasionally be inconsistent with their preferences, due to inattention or other biases, and some preferences may be more common in the population than others (such as a preference for lower price). To evaluate the models in a more realistic setting, we recruit human participants to act as users. Each human participant is asked to first state their preferences for each of the flight features, and then select their preferred flight out of three options, for five different sets of options. We collect data from 10 human participants each for 50 lists of flight option sets, for a total of 500 participants (see Appendix Section~\ref{app:methods}).

\begin{figure}[t!]
\centering
\includegraphics[width=\linewidth]{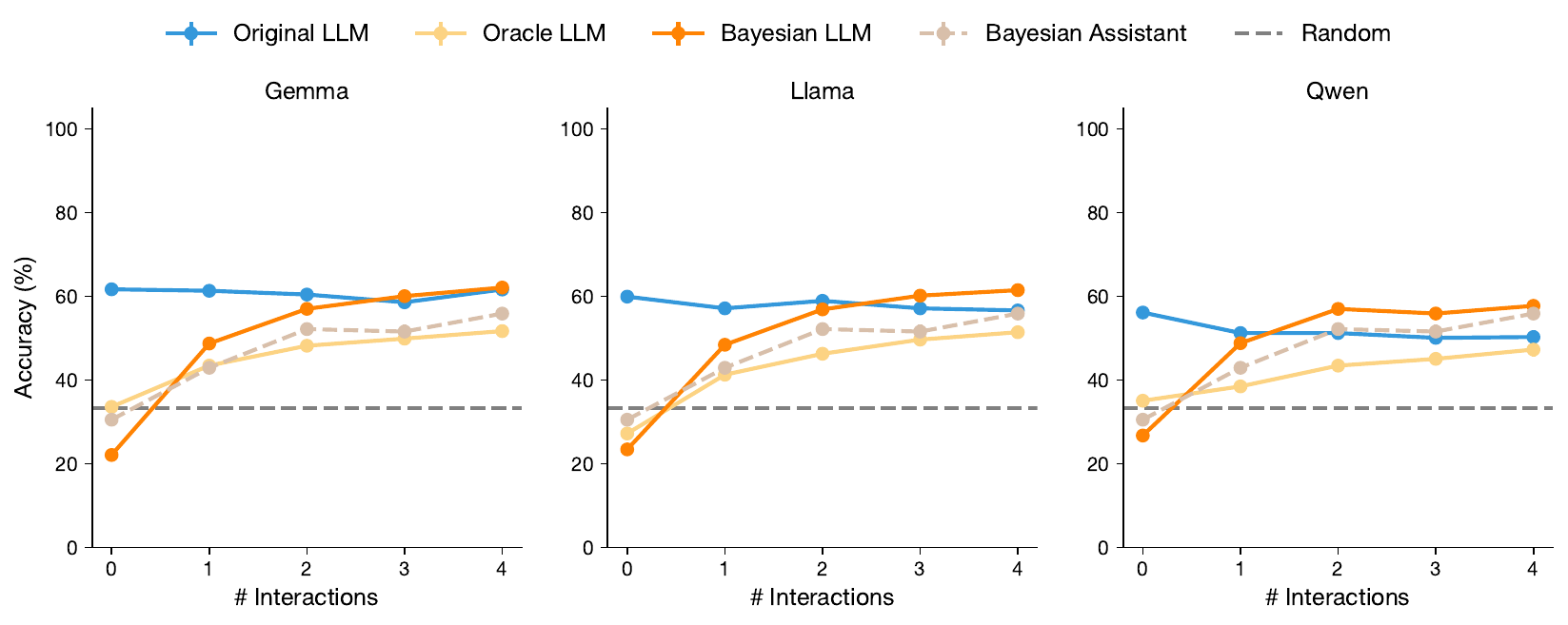}
\caption{\textbf{Bayesian teaching generalizes to human users.} We show accuracy over rounds when the user is a human participant. The original LLMs achieve strong performance but do not show any learning behavior. In contrast, fine-tuned LLMs (with both Bayesian and Oracle teachers) improve their performance over rounds, and the Bayesian LLMs consistently outperform the Oracle LLMs. Error bars show standard errors across four random seeds (and three training runs; the errors bars are not visible in the plot because they are very small).}
\label{fig:fig6_human_user}
\end{figure}

The performance of both fine-tuned models and the Bayesian Assistant for human users consistently improves over rounds (Fig.~\ref{fig:fig6_human_user}), and, as was the case for the simulated users, the Bayesian LLMs consistently outperform the Oracle LLMs; at least for some model families, the Bayesian LLMs also outperform the original LLMs. This indicates that the Bayesian LLMs generalize to human users from the simulated users on which they were fine-tuned.

All models, including the Bayesian Assistant, show substantially lower performance for humans than they did for simulated users, where accuracy after five rounds approached 80\% (Fig.~\ref{fig:fig3_ft_results}). In the Appendix Section~\ref{sec:app_e2_human_as_user}, we show that this is due to the fact that participants' choices are not always consistent with their stated preferences, and as such are impossible to predict with high accuracy (Appendix Fig.~\ref{fig:app_fige16a_human_user_consistency}).
For the subset of human users whose choices are perfectly consistent with their preferences, the Bayesian LLM performs much better than the original LLM  (Appendix Fig.~\ref{fig:app_fige15b_human_set_acc}; see also Appendix Section~\ref{sec:app_c3_nondeterministic_users}, where we study inconsistent simulated users). 

Unlike for the simulated users, for human users the original LLMs perform well even after a single interaction (although, crucially, the original LLMs do not improve over interactions). We attribute the original LLMs' surprisingly strong performance to the fact that human users have generally predictable preferences (e.g., a preference for cheaper flights), such that guesses based on the LLM's priors, without any adaptation to the individual user, can be quite effective (see Appendix Figs.~\ref{fig:app_fige14_human_preference} and~\ref{fig:app_fige15_human_set_analysis} for evidence for this hypothesis).

\section{Discussion}

To interact with the world successfully, an agent needs to adapt its behavior as it obtains additional information about the statistics of this environment. To evaluate the ability of large language models (LLMs) to do so, we introduced a simple flight recommendation task where, in order to make accurate predictions, the model needs to adapt to a user's preferences over multiple interactions with the user. We tested a range of LLMs and found that they struggle to form and update probabilistic beliefs. %
We further found that continuing the LLMs' training through exposure to interactions between users and the Bayesian Assistant---a model that implements the normative probabilistic belief update strategy---dramatically improves the LLMs' ability to approximate probabilistic reasoning. Crucially, this improvement did not only hold for the flight recommendation task the LLM was trained on, but also generalized to variants to the flight recommendation task that the LLM has not encountered before, as well as to other tasks. Across the board, this approach, which we refer to as Bayesian teaching, was more effective than a related approach where the LLM is fine-tuned directly on the correct answers, pointing to the effectiveness of the Bayesian training signal.

Our paradigm differs from those used in previous investigations of LLMs' probabilistic reasoning abilities, where LLMs were expected to compute statistics explicitly~\citep{nafar2024probabilistic, paruchuri2024odds} or provide probability judgments~\citep{zhu2024incoherent, belem2024perceptions}.
In our paradigm, probabilistic reasoning is as essential as it is in explicit reasoning tasks, but, crucially, it is implicit in the task.
Unlike in some recent studies, where the assistant is expected to ask questions to directly elicit the user's preferences \citep{li2023eliciting, handa2024bayesian, piriyakulkij2023active, andukuri2024star, pengpragmatic, aliannejadi2021building, chen2024chatshop, lin2022inferring}, our setup expects the assistant to gradually infer the user's preferences by simply observing the user's choices,  and to provide recommendations that are increasingly in line with the user's true preferences. Finally, our findings are consistent with those of concurrent work \citep{zhao2025do}, which also investigates LLMs' ability to infer user preferences from different types of dialogues, including a condition where the user accepts or rejects one or more options provided by the assistant---a setup similar to ours---where the models performed poorly.
Compared to this concurrent study, our work analyzes the LLMs' behavior through the lens of Bayesian inference, and demonstrates the benefits of mimicking a Bayesian model in fine-tuning compared to a more standard fine-tuning strategy, where the model is always provided with the correct answer (oracle teaching, in the terminology we used in the current paper).

We observed robust generalization from the synthetic flight recommendation task on which the LLMs were fine-tuned to the more natural web shopping task. While performance was even stronger when we fine-tuned the LLM directly on interactions from this task (the green dashed line in Fig.~\ref{fig:fig5c}), in practice it may be difficult or expensive to collect such data; our synthetic fine-tuning strategy provides an alternative that improves the LLM's probabilistic reasoning abilities across tasks, without requiring collecting additional data and re-training the model on the new domain.

Our proposal is related to but distinct from approaches that embed an LLM inside a neuro-symbolic framework for probabilistic reasoning~\citep{wong2023word, feng2024bird, liu2024large, piriyakulkij2024doing, grand2023loose, ying2024grounding, ellis2023human}.
In those approaches, the LLM is used to translate between natural language inputs and formal representations, which in turn serve as input to a symbolic model that can update its beliefs according to the Bayesian framework~\citep{wong2023word}. Indeed, we provide further evidence that hybrid methods can outperform the LLM-only approach in Appendix Section~\ref{sec:app_a_verbalize_beliefs}, where we describe a variation of our method where we first ask the LLM to verbalize its beliefs about the user's preferences, and then we use an external, symbolic system to make predictions based on these verbalized beliefs. The experiments described in that Appendix section show that in simple tasks where preferences can be mapped to predictions, such hybrid methods indeed outperform a direct interaction with the LLM. Our preliminary explorations of this approach can be developed in greater detail in future work.

Besides their superior performance in certain cases, neuro-symbolic methods have the benefit of greater interpretability, and their probabilistic inferences could be more robust. Crucially, however, the utility of such methods is limited to problems whose structure can be made explicit in the symbolic component of the system. By contrast, the method we propose empowers the LLM to approximate probabilistic inference on its own, such that it can apply this skill to domains that are hard to codify explicitly in a symbolic system, domains such as the web shopping task we have examined. This approach leverages LLMs' remarkable ability to generalize to new problems defined using natural language.

Notably, even in cases where the domain is simple enough for a purely symbolic model to be constructed, such models may not be consistently more accurate than LLMs. In our study, we found that while for ``well-behaved'' simulated users a moderate performance gap persisted between the fine-tuned models and the Bayesian Assistant, for human users, whose choices are not always consistent with their preferences, our Bayesian LLMs were in fact superior to the fully symbolic Bayesian Assistant, demonstrating LLMs' greater robustness to noise compared to symbolic models.

We have argued that through mimicking the Bayesian Assistant the LLMs learn to perform probabilistic inference, albeit only approximately. This hypothesis may appear to be surprising in light of the fact that LLMs' training objective does not explicitly provide supervision for this skill, and that the transformer architecture does not explicitly track probability distributions: it is trained only to predict the next word produced by the Bayesian Assistant. That being said, there is mounting evidence that in order to predict the next token successfully, LLMs can acquire sophisticated representations that match the structure of the process that generated those tokens. In the case of natural language syntax, for example, the internal representations of LLM trained solely to predict upcoming words have been shown to encode abstract features such as syntactic role and grammatical number \citep{lakretz-etal-2019-emergence,hao-linzen-2023-verb,manning2020emergent}. It would be a fruitful direction for future work to determine how probabilistic reasoning is implemented by the LLMs' internal representations, for example by using techniques such as probes and causal interventions \citep{finlayson-etal-2021-causal,ravfogel-etal-2021-counterfactual,vig2020investigating} to find internal representations of the model's probability distributions over users' preferences, or using circuit analysis \citep{wang2023} to explore the computations through which the model updates these distributions.

The success of Bayesian teaching in imparting approximate probabilistic reasoning skills to LLMs opens up a range of questions for future work. Would the benefits of Bayesian teaching extend to larger models than we were able to fine-tune in this work, or to the recent generation of models that are explicitly trained to reason in words  \citep{guo2025deepseek}? Does the benefit of Bayesian teaching extend to continuous domains and real-world applications beyond the ones we evaluated (for example, interactions whose goal goes beyond shopping)? Could we provide the models with a stronger supervision signal---for example, by instructing them to consider explicit probability distributions, by providing them with explicit supervision on the optimal way to update these distributions (for example, by supervising beliefs as in Appendix Fig.~\ref{fig:app_figb4c_data}), or by encouraging them to maintain explicit representations of users such that the probability distributions are consistent across interactions with the same user, through methods such as supervised fine-tuning or reinforcement learning?

The goal of this study was not to replicate human behavior in LLMs, but rather to identify methods that can bring LLMs' probabilistic reasoning skills closer to the normative Bayesian strategy: for most applications we expect AI assistants to be follow normative reasoning standards rather than reproduce human deviations from that standard. That being said, our comparisons between LLMs and humans point to a number of directions for future work. Our participants showed substantial deviations from the normative reasoning strategy, in line with prior work on reasoning biases \citep{eisape-etal-2024-systematic,rottman2016people,chaigneau2025breaking,tversky1974judgment}. To what extent can people be taught to follow the normative strategy more closely? Can participants' apparent biases be explained as consequences of resource limitations \citep{simon1955behavioral}? How consistent are participants' choices with their stated preferences? Do people's deviations from the normative strategy align with those of LLMs \citep{eisape-etal-2024-systematic}, and what properties of an LLM lead to closer alignment with humans?

While our findings from our first experiment point to the limitations of particular LLMs, the positive findings of our subsequent fine-tuning experiments can be viewed as a demonstration of the strength of the LLM ``post-training'' paradigm more generally: by training the LLMs on demonstrations of the normative strategy to perform the task, we were able to improve their performance considerably, suggesting that they learned to approximate the probabilistic reasoning strategy illustrated by the demonstrations. The LLMs were able to generalize this strategy to domains where it is difficult to encode it explicitly in a symbolic model, demonstrating the power of distilling a classic symbolic model into a neural network. We hypothesize that this generalization ability is, in part, responsible for LLMs' remarkable empirical success.

\section*{Acknowledgments}
We thank Stephanie Chan, Andrew Lampinen, Michael Mozer, Peter Shaw, and Zhaofeng Wu for helpful discussions.

\section*{Author Contributions}

L.Q., F.S., T.L., and S.V.S. co-led the project. S.V.S. conceptualized the project direction. L.Q. conducted the experiments and analysis. L.Q., F.S., T.L., and S.V.S. framed, analyzed and designed experiments, with inputs from K.A. and Y.K. L.Q., T.L., and S.V.S. wrote the paper with help from F.S., K.A., and Y.K.

\bibliography{reference}

\begin{thebibliography}{84}
\providecommand{\natexlab}[1]{#1}
\providecommand{\url}[1]{\texttt{#1}}
\expandafter\ifx\csname urlstyle\endcsname\relax
  \providecommand{\doi}[1]{doi: #1}\else
  \providecommand{\doi}{doi: \begingroup \urlstyle{rm}\Url}\fi

\bibitem[Achiam et~al.(2023)Achiam, Adler, Agarwal, Ahmad, Akkaya, Aleman, Almeida, Altenschmidt, Altman, Anadkat, et~al.]{achiam2023gpt}
J.~Achiam, S.~Adler, S.~Agarwal, L.~Ahmad, I.~Akkaya, F.~L. Aleman, D.~Almeida, J.~Altenschmidt, S.~Altman, S.~Anadkat, et~al.
\newblock {GPT-4} technical report.
\newblock \emph{ArXiv preprint}, abs/2303.08774, 2023.

\bibitem[Aliannejadi et~al.(2021)Aliannejadi, Kiseleva, Chuklin, Dalton, and Burtsev]{aliannejadi2021building}
M.~Aliannejadi, J.~Kiseleva, A.~Chuklin, J.~Dalton, and M.~Burtsev.
\newblock Building and evaluating open-domain dialogue corpora with clarifying questions.
\newblock In \emph{Proceedings of the 2021 Conference on Empirical Methods in Natural Language Processing}, 2021.

\bibitem[Andukuri et~al.(2024)Andukuri, Fr{\"a}nken, Gerstenberg, and Goodman]{andukuri2024star}
C.~Andukuri, J.-P. Fr{\"a}nken, T.~Gerstenberg, and N.~Goodman.
\newblock {ST}a{R-GATE}: Teaching language models to ask clarifying questions.
\newblock In \emph{First Conference on Language Modeling}, 2024.

\bibitem[Anthropic(2024)]{claude}
Anthropic.
\newblock Claude 3, 2024.

\bibitem[Baker et~al.(2011)Baker, Saxe, and Tenenbaum]{baker2011bayesian}
C.~Baker, R.~Saxe, and J.~Tenenbaum.
\newblock Bayesian theory of mind: Modeling joint belief-desire attribution.
\newblock In \emph{Proceedings of the annual meeting of the cognitive science society}, volume~33, 2011.

\bibitem[Bel{\'e}m et~al.(2024)Bel{\'e}m, Kelly, Steyvers, Singh, and Smyth]{belem2024perceptions}
C.~G. Bel{\'e}m, M.~Kelly, M.~Steyvers, S.~Singh, and P.~Smyth.
\newblock Perceptions of linguistic uncertainty by language models and humans.
\newblock In \emph{Proceedings of the 2024 Conference on Empirical Methods in Natural Language Processing}, 2024.

\bibitem[Brown et~al.(2020)Brown, Mann, Ryder, Subbiah, Kaplan, Dhariwal, Neelakantan, Shyam, Sastry, Askell, Agarwal, Herbert{-}Voss, Krueger, Henighan, Child, Ramesh, Ziegler, Wu, Winter, Hesse, Chen, Sigler, Litwin, Gray, Chess, Clark, Berner, McCandlish, Radford, Sutskever, and Amodei]{brown2020language}
T.~B. Brown, B.~Mann, N.~Ryder, M.~Subbiah, J.~Kaplan, P.~Dhariwal, A.~Neelakantan, P.~Shyam, G.~Sastry, A.~Askell, S.~Agarwal, A.~Herbert{-}Voss, G.~Krueger, T.~Henighan, R.~Child, A.~Ramesh, D.~M. Ziegler, J.~Wu, C.~Winter, C.~Hesse, M.~Chen, E.~Sigler, M.~Litwin, S.~Gray, B.~Chess, J.~Clark, C.~Berner, S.~McCandlish, A.~Radford, I.~Sutskever, and D.~Amodei.
\newblock Language models are few-shot learners.
\newblock In \emph{Advances in Neural Information Processing Systems 33: Annual Conference on Neural Information Processing Systems 2020, NeurIPS 2020, December 6-12, 2020, virtual}, 2020.

\bibitem[Chaigneau et~al.(2025)Chaigneau, Marchant, and Rehder]{chaigneau2025breaking}
S.~Chaigneau, N.~Marchant, and B.~Rehder.
\newblock Breaking the chains of independence: A bayesian uncertainty model of normative violations in human causal probabilistic reasoning.
\newblock \emph{OSF}, 2025.

\bibitem[Chater and Manning(2006)]{chaterm06}
N.~Chater and C.~D. Manning.
\newblock Probabilistic models of language processing and acquisition.
\newblock \emph{Trends in Cognitive Sciences}, 10, 2006.

\bibitem[Chater et~al.(2006)Chater, Tenenbaum, and Yuille]{chater2006probabilistic}
N.~Chater, J.~B. Tenenbaum, and A.~Yuille.
\newblock Probabilistic models of cognition: Conceptual foundations.
\newblock \emph{Trends in Cognitive Sciences}, 10\penalty0 (7), 2006.

\bibitem[Chen et~al.(2024{\natexlab{a}})Chen, Wiseman, and Dhingra]{chen2024chatshop}
S.~Chen, S.~Wiseman, and B.~Dhingra.
\newblock Chatshop: Interactive information seeking with language agents.
\newblock \emph{ArXiv preprint}, abs/2404.09911, 2024{\natexlab{a}}.

\bibitem[Chen et~al.(2024{\natexlab{b}})Chen, Huang, Gao, Wang, Zhao, and Ding]{chen2024learning}
X.~Chen, H.~Huang, Y.~Gao, Y.~Wang, J.~Zhao, and K.~Ding.
\newblock Learning to maximize mutual information for chain-of-thought distillation.
\newblock In \emph{Findings of the Association for Computational Linguistics: ACL 2024}, 2024{\natexlab{b}}.

\bibitem[Chiang et~al.(2024)Chiang, Zheng, Sheng, Angelopoulos, Li, Li, Zhu, Zhang, Jordan, Gonzalez, and Stoica]{chiang2024chatbot}
W.~Chiang, L.~Zheng, Y.~Sheng, A.~N. Angelopoulos, T.~Li, D.~Li, B.~Zhu, H.~Zhang, M.~I. Jordan, J.~E. Gonzalez, and I.~Stoica.
\newblock Chatbot arena: An open platform for evaluating llms by human preference.
\newblock In \emph{Forty-first International Conference on Machine Learning, {ICML} 2024, Vienna, Austria, July 21-27, 2024}, 2024.

\bibitem[Christiano et~al.(2017)Christiano, Leike, Brown, Martic, Legg, and Amodei]{christiano2017deep}
P.~F. Christiano, J.~Leike, T.~B. Brown, M.~Martic, S.~Legg, and D.~Amodei.
\newblock Deep reinforcement learning from human preferences.
\newblock In \emph{Advances in Neural Information Processing Systems 30: Annual Conference on Neural Information Processing Systems 2017, December 4-9, 2017, Long Beach, CA, {USA}}, 2017.

\bibitem[Deng et~al.(2023)Deng, Prasad, Fernandez, Smolensky, Chaudhary, and Shieber]{deng2023implicit}
Y.~Deng, K.~Prasad, R.~Fernandez, P.~Smolensky, V.~Chaudhary, and S.~Shieber.
\newblock Implicit chain of thought reasoning via knowledge distillation.
\newblock \emph{ArXiv preprint}, abs/2311.01460, 2023.

\bibitem[Eisape et~al.(2024)Eisape, Tessler, Dasgupta, Sha, Steenkiste, and Linzen]{eisape-etal-2024-systematic}
T.~Eisape, M.~Tessler, I.~Dasgupta, F.~Sha, S.~Steenkiste, and T.~Linzen.
\newblock A systematic comparison of syllogistic reasoning in humans and language models.
\newblock In \emph{Proceedings of the 2024 Conference of the North American Chapter of the Association for Computational Linguistics: Human Language Technologies (Volume 1: Long Papers)}, 2024.

\bibitem[Ellis(2023)]{ellis2023human}
K.~Ellis.
\newblock Human-like few-shot learning via bayesian reasoning over natural language.
\newblock In \emph{Advances in Neural Information Processing Systems 36: Annual Conference on Neural Information Processing Systems 2023, NeurIPS 2023, New Orleans, LA, USA, December 10 - 16, 2023}, 2023.

\bibitem[Feng et~al.(2024)Feng, Zhou, Lin, and Roth]{feng2024bird}
Y.~Feng, B.~Zhou, W.~Lin, and D.~Roth.
\newblock {BIRD}: A trustworthy bayesian inference framework for large language models.
\newblock In \emph{The Thirteenth International Conference on Learning Representations}, 2024.

\bibitem[Finlayson et~al.(2021)Finlayson, Mueller, Gehrmann, Shieber, Linzen, and Belinkov]{finlayson-etal-2021-causal}
M.~Finlayson, A.~Mueller, S.~Gehrmann, S.~Shieber, T.~Linzen, and Y.~Belinkov.
\newblock Causal analysis of syntactic agreement mechanisms in neural language models.
\newblock In \emph{Proceedings of the 59th Annual Meeting of the Association for Computational Linguistics and the 11th International Joint Conference on Natural Language Processing (Volume 1: Long Papers)}, 2021.

\bibitem[Grand et~al.(2023)Grand, Pepe, Andreas, and Tenenbaum]{grand2023loose}
G.~Grand, V.~Pepe, J.~Andreas, and J.~Tenenbaum.
\newblock Loose lips sink ships: Asking questions in battleship with language-informed program sampling.
\newblock In \emph{Proceedings of the Annual Meeting of the Cognitive Science Society}, volume~46, 2023.

\bibitem[Grattafiori et~al.(2024)Grattafiori, Dubey, Jauhri, Pandey, Kadian, Al-Dahle, Letman, Mathur, Schelten, Vaughan, et~al.]{dubey2024llama3herdmodels}
A.~Grattafiori, A.~Dubey, A.~Jauhri, A.~Pandey, A.~Kadian, A.~Al-Dahle, A.~Letman, A.~Mathur, A.~Schelten, A.~Vaughan, et~al.
\newblock The llama 3 herd of models, 2024.

\bibitem[Griffiths et~al.(2007)Griffiths, Steyvers, and Tenenbaum]{griffithsst07}
T.~L. Griffiths, M.~Steyvers, and J.~B. Tenenbaum.
\newblock Topics in semantic association.
\newblock \emph{Psychological Review}, 114, 2007.

\bibitem[Griffiths et~al.(2024)Griffiths, Chater, and Tenenbaum]{griffiths2024bayesian}
T.~L. Griffiths, N.~Chater, and J.~B. Tenenbaum.
\newblock \emph{Bayesian Models of Cognition: Reverse Engineering the Mind}.
\newblock The MIT Press, 2024.
\newblock ISBN 9780262049412.

\bibitem[Guo et~al.(2025)Guo, Yang, Zhang, Song, Zhang, Xu, Zhu, Ma, Wang, Bi, et~al.]{guo2025deepseek}
D.~Guo, D.~Yang, H.~Zhang, J.~Song, R.~Zhang, R.~Xu, Q.~Zhu, S.~Ma, P.~Wang, X.~Bi, et~al.
\newblock Deepseek-r1: Incentivizing reasoning capability in {LLMs} via reinforcement learning.
\newblock \emph{Nature}, 645, 2025.

\bibitem[Ha and Schmidhuber(2018)]{ha2018recurrent}
D.~Ha and J.~Schmidhuber.
\newblock Recurrent world models facilitate policy evolution.
\newblock \emph{Advances in Neural Information Processing Systems}, 31, 2018.

\bibitem[Handa et~al.(2024)Handa, Gal, Pavlick, Goodman, Andreas, Tamkin, and Li]{handa2024bayesian}
K.~Handa, Y.~Gal, E.~Pavlick, N.~Goodman, J.~Andreas, A.~Tamkin, and B.~Z. Li.
\newblock Bayesian preference elicitation with language models.
\newblock \emph{ArXiv preprint}, abs/2403.05534, 2024.

\bibitem[Hao and Linzen(2023)]{hao-linzen-2023-verb}
S.~Hao and T.~Linzen.
\newblock Verb conjugation in transformers is determined by linear encodings of subject number.
\newblock In \emph{Findings of the Association for Computational Linguistics: EMNLP 2023}, 2023.

\bibitem[Hinton et~al.(2015)Hinton, Vinyals, and Dean]{hinton2015distilling}
G.~Hinton, O.~Vinyals, and J.~Dean.
\newblock Distilling the knowledge in a neural network.
\newblock \emph{stat}, 1050:\penalty0 9, 2015.

\bibitem[Hu et~al.(2022)Hu, Shen, Wallis, Allen{-}Zhu, Li, Wang, Wang, and Chen]{hu2021lora}
E.~J. Hu, Y.~Shen, P.~Wallis, Z.~Allen{-}Zhu, Y.~Li, S.~Wang, L.~Wang, and W.~Chen.
\newblock Lora: Low-rank adaptation of large language models.
\newblock In \emph{The Tenth International Conference on Learning Representations, {ICLR} 2022, Virtual Event, April 25-29, 2022}, 2022.

\bibitem[Hu and Levy(2023)]{hu2023prompting}
J.~Hu and R.~Levy.
\newblock Prompting is not a substitute for probability measurements in large language models.
\newblock In \emph{Proceedings of the 2023 Conference on Empirical Methods in Natural Language Processing}, 2023.

\bibitem[J.~Koehler and James(2010)]{koehler2010probability}
D.~J.~Koehler and G.~James.
\newblock Probability matching and strategy availability.
\newblock \emph{Memory \& Cognition}, 38\penalty0 (6), 2010.

\bibitem[Jern et~al.(2017)Jern, Lucas, and Kemp]{jern2017people}
A.~Jern, C.~G. Lucas, and C.~Kemp.
\newblock People learn other people’s preferences through inverse decision-making.
\newblock \emph{Cognition}, 168, 2017.
\newblock ISSN 0010-0277.

\bibitem[Johnson-Laird(1980)]{johnson1980mental}
P.~N. Johnson-Laird.
\newblock Mental models in cognitive science.
\newblock \emph{Cognitive Science}, 4\penalty0 (1), 1980.

\bibitem[Jung et~al.(2024)Jung, West, Jiang, Brahman, Lu, Fisher, Sorensen, and Choi]{jung2024impossible}
J.~Jung, P.~West, L.~Jiang, F.~Brahman, X.~Lu, J.~Fisher, T.~Sorensen, and Y.~Choi.
\newblock Impossible distillation for paraphrasing and summarization: How to make high-quality lemonade out of small, low-quality model.
\newblock In \emph{Proceedings of the 2024 Conference of the North American Chapter of the Association for Computational Linguistics: Human Language Technologies (Volume 1: Long Papers)}, 2024.

\bibitem[Kim and Rush(2016)]{kim2016sequence}
Y.~Kim and A.~M. Rush.
\newblock Sequence-level knowledge distillation.
\newblock In \emph{Proceedings of the 2016 Conference on Empirical Methods in Natural Language Processing}, 2016.

\bibitem[Kojima et~al.(2022)Kojima, Gu, Reid, Matsuo, and Iwasawa]{kojima2022large}
T.~Kojima, S.~S. Gu, M.~Reid, Y.~Matsuo, and Y.~Iwasawa.
\newblock Large language models are zero-shot reasoners.
\newblock In \emph{Advances in Neural Information Processing Systems 35: Annual Conference on Neural Information Processing Systems 2022, NeurIPS 2022, New Orleans, LA, USA, November 28 - December 9, 2022}, 2022.

\bibitem[Kotha et~al.(2024)Kotha, Springer, and Raghunathan]{kotha2023understanding}
S.~Kotha, J.~M. Springer, and A.~Raghunathan.
\newblock Understanding catastrophic forgetting in language models via implicit inference.
\newblock In \emph{The Twelfth International Conference on Learning Representations, {ICLR} 2024, Vienna, Austria, May 7-11, 2024}, 2024.

\bibitem[Lakretz et~al.(2019)Lakretz, Kruszewski, Desbordes, Hupkes, Dehaene, and Baroni]{lakretz-etal-2019-emergence}
Y.~Lakretz, G.~Kruszewski, T.~Desbordes, D.~Hupkes, S.~Dehaene, and M.~Baroni.
\newblock The emergence of number and syntax units in {LSTM} language models.
\newblock In \emph{Proceedings of the 2019 Conference of the North {A}merican Chapter of the Association for Computational Linguistics: Human Language Technologies, Volume 1 (Long and Short Papers)}, 2019.

\bibitem[LeCun(2022)]{lecun2022path}
Y.~LeCun.
\newblock A path towards autonomous machine intelligence.
\newblock \emph{Open Review}, 62\penalty0 (1), 2022.

\bibitem[Li et~al.(2023{\natexlab{a}})Li, Tamkin, Goodman, and Andreas]{li2023eliciting}
B.~Z. Li, A.~Tamkin, N.~Goodman, and J.~Andreas.
\newblock Eliciting human preferences with language models.
\newblock In \emph{The Thirteenth International Conference on Learning Representations}, 2023{\natexlab{a}}.

\bibitem[Li et~al.(2023{\natexlab{b}})Li, Hessel, Yu, Ren, Chang, and Choi]{li2023symbolic}
L.~H. Li, J.~Hessel, Y.~Yu, X.~Ren, K.-W. Chang, and Y.~Choi.
\newblock Symbolic chain-of-thought distillation: Small models can also ``think'' step-by-step.
\newblock In \emph{Proceedings of the 61st Annual Meeting of the Association for Computational Linguistics (Volume 1: Long Papers)}, 2023{\natexlab{b}}.

\bibitem[Lin et~al.(2022)Lin, Fried, Klein, and Dragan]{lin2022inferring}
J.~Lin, D.~Fried, D.~Klein, and A.~Dragan.
\newblock Inferring rewards from language in context.
\newblock In \emph{Proceedings of the 60th Annual Meeting of the Association for Computational Linguistics (Volume 1: Long Papers)}, 2022.

\bibitem[Lin et~al.(2024)Lin, Lin, Xiong, Diao, Liu, Zhang, Pan, Wang, Hu, Zhang, Dong, Pi, Zhao, Jiang, Ji, Yao, and Zhang]{lin2023mitigating}
Y.~Lin, H.~Lin, W.~Xiong, S.~Diao, J.~Liu, J.~Zhang, R.~Pan, H.~Wang, W.~Hu, H.~Zhang, H.~Dong, R.~Pi, H.~Zhao, N.~Jiang, H.~Ji, Y.~Yao, and T.~Zhang.
\newblock Mitigating the alignment tax of {RLHF}.
\newblock In \emph{Proceedings of the 2024 Conference on Empirical Methods in Natural Language Processing}, 2024.

\bibitem[Liu et~al.(2024)Liu, Geng, Peterson, Sucholutsky, and Griffiths]{liu2024large}
R.~Liu, J.~Geng, J.~Peterson, I.~Sucholutsky, and T.~L. Griffiths.
\newblock Large language models assume people are more rational than we really are.
\newblock In \emph{The Thirteenth International Conference on Learning Representations}, 2024.

\bibitem[Manning et~al.(2020)Manning, Clark, Hewitt, Khandelwal, and Levy]{manning2020emergent}
C.~D. Manning, K.~Clark, J.~Hewitt, U.~Khandelwal, and O.~Levy.
\newblock Emergent linguistic structure in artificial neural networks trained by self-supervision.
\newblock \emph{Proceedings of the National Academy of Sciences}, 117\penalty0 (48), 2020.

\bibitem[Nafar et~al.(2025)Nafar, Venable, and Kordjamshidi]{nafar2024probabilistic}
A.~Nafar, K.~B. Venable, and P.~Kordjamshidi.
\newblock Reasoning over uncertain text by generative large language models.
\newblock In \emph{Proceedings of the AAAI Conference on Artificial Intelligence}, 2025.

\bibitem[Nye et~al.(2021)Nye, Andreassen, Gur-Ari, Michalewski, Austin, Bieber, Dohan, Lewkowycz, Bosma, Luan, et~al.]{nye2021show}
M.~Nye, A.~J. Andreassen, G.~Gur-Ari, H.~Michalewski, J.~Austin, D.~Bieber, D.~Dohan, A.~Lewkowycz, M.~Bosma, D.~Luan, et~al.
\newblock Show your work: Scratchpads for intermediate computation with language models.
\newblock In \emph{Deep Learning for Code Workshop}, 2021.

\bibitem[OpenAI(2025)]{openai2025gpt41}
OpenAI.
\newblock Introducing {GPT}-4.1 in the {API}, 2025.

\bibitem[Ouyang et~al.(2022)Ouyang, Wu, Jiang, Almeida, Wainwright, Mishkin, Zhang, Agarwal, Slama, Ray, Schulman, Hilton, Kelton, Miller, Simens, Askell, Welinder, Christiano, Leike, and Lowe]{ouyang2022training}
L.~Ouyang, J.~Wu, X.~Jiang, D.~Almeida, C.~L. Wainwright, P.~Mishkin, C.~Zhang, S.~Agarwal, K.~Slama, A.~Ray, J.~Schulman, J.~Hilton, F.~Kelton, L.~Miller, M.~Simens, A.~Askell, P.~Welinder, P.~F. Christiano, J.~Leike, and R.~Lowe.
\newblock Training language models to follow instructions with human feedback.
\newblock In \emph{Advances in Neural Information Processing Systems 35: Annual Conference on Neural Information Processing Systems 2022, NeurIPS 2022, New Orleans, LA, USA, November 28 - December 9, 2022}, 2022.

\bibitem[Palan and Schitter(2018)]{palan2018prolific}
S.~Palan and C.~Schitter.
\newblock Prolific.ac---{A} subject pool for online experiments.
\newblock \emph{Journal of Behavioral and Experimental Finance}, 17, 2018.

\bibitem[Paruchuri et~al.(2024)Paruchuri, Garrison, Liao, Hernandez, Sunshine, Althoff, Liu, and McDuff]{paruchuri2024odds}
A.~Paruchuri, J.~Garrison, S.~Liao, J.~B. Hernandez, J.~Sunshine, T.~Althoff, X.~Liu, and D.~McDuff.
\newblock What are the odds? language models are capable of probabilistic reasoning.
\newblock In \emph{Proceedings of the 2024 Conference on Empirical Methods in Natural Language Processing}, 2024.

\bibitem[Peng et~al.(2024)Peng, Sun, Shu, and Abel]{pengpragmatic}
A.~Peng, Y.~Sun, T.~Shu, and D.~Abel.
\newblock Pragmatic feature preferences: Learning reward-relevant preferences from human input.
\newblock In \emph{Forty-first International Conference on Machine Learning, {ICML} 2024, Vienna, Austria, July 21-27, 2024}, 2024.

\bibitem[Piriyakulkij et~al.(2023)Piriyakulkij, Kuleshov, and Ellis]{piriyakulkij2023active}
T.~Piriyakulkij, V.~Kuleshov, and K.~Ellis.
\newblock Active preference inference using language models and probabilistic reasoning.
\newblock In \emph{NeurIPS 2023 Foundation Models for Decision Making Workshop}, 2023.

\bibitem[Piriyakulkij et~al.(2024)Piriyakulkij, Langenfeld, Le, and Ellis]{piriyakulkij2024doing}
T.~Piriyakulkij, C.~Langenfeld, T.~A. Le, and K.~Ellis.
\newblock Doing experiments and revising rules with natural language and probabilistic reasoning.
\newblock In \emph{Advances in Neural Information Processing Systems 38: Annual Conference on Neural Information Processing Systems 2024, NeurIPS 2024, Vancouver, BC, Canada, December 10 - 15, 2024}, 2024.

\bibitem[Rafailov et~al.(2023)Rafailov, Sharma, Mitchell, Manning, Ermon, and Finn]{rafailov2023direct}
R.~Rafailov, A.~Sharma, E.~Mitchell, C.~D. Manning, S.~Ermon, and C.~Finn.
\newblock Direct preference optimization: Your language model is secretly a reward model.
\newblock In \emph{Advances in Neural Information Processing Systems 36: Annual Conference on Neural Information Processing Systems 2023, NeurIPS 2023, New Orleans, LA, USA, December 10 - 16, 2023}, 2023.

\bibitem[Ravfogel et~al.(2021)Ravfogel, Prasad, Linzen, and Goldberg]{ravfogel-etal-2021-counterfactual}
S.~Ravfogel, G.~Prasad, T.~Linzen, and Y.~Goldberg.
\newblock Counterfactual interventions reveal the causal effect of relative clause representations on agreement prediction.
\newblock In \emph{Proceedings of the 25th Conference on Computational Natural Language Learning}, 2021.

\bibitem[Rehder(2018)]{rehder2018beyond}
B.~Rehder.
\newblock Beyond {M}arkov: Accounting for independence violations in causal reasoning.
\newblock \emph{Cognitive Psychology}, 103, 2018.

\bibitem[Rottman and Hastie(2016)]{rottman2016people}
B.~M. Rottman and R.~Hastie.
\newblock Do people reason rationally about causally related events? {M}arkov violations, weak inferences, and failures of explaining away.
\newblock \emph{Cognitive Psychology}, 87, 2016.

\bibitem[Sanh et~al.(2022)Sanh, Webson, Raffel, Bach, Sutawika, Alyafeai, Chaffin, Stiegler, Raja, Dey, Bari, Xu, Thakker, Sharma, Szczechla, Kim, Chhablani, Nayak, Datta, Chang, Jiang, Wang, Manica, Shen, Yong, Pandey, Bawden, Wang, Neeraj, Rozen, Sharma, Santilli, F{\'{e}}vry, Fries, Teehan, Scao, Biderman, Gao, Wolf, and Rush]{sanh2022}
V.~Sanh, A.~Webson, C.~Raffel, S.~H. Bach, L.~Sutawika, Z.~Alyafeai, A.~Chaffin, A.~Stiegler, A.~Raja, M.~Dey, M.~S. Bari, C.~Xu, U.~Thakker, S.~S. Sharma, E.~Szczechla, T.~Kim, G.~Chhablani, N.~V. Nayak, D.~Datta, J.~Chang, M.~T. Jiang, H.~Wang, M.~Manica, S.~Shen, Z.~X. Yong, H.~Pandey, R.~Bawden, T.~Wang, T.~Neeraj, J.~Rozen, A.~Sharma, A.~Santilli, T.~F{\'{e}}vry, J.~A. Fries, R.~Teehan, T.~L. Scao, S.~Biderman, L.~Gao, T.~Wolf, and A.~M. Rush.
\newblock Multitask prompted training enables zero-shot task generalization.
\newblock In \emph{The Tenth International Conference on Learning Representations, {ICLR} 2022, Virtual Event, April 25-29, 2022}, 2022.

\bibitem[Simon(1955)]{simon1955behavioral}
H.~A. Simon.
\newblock A behavioral model of rational choice.
\newblock \emph{The Quarterly Journal of Economics}, 1955.

\bibitem[Sloman and Lagnado(2015)]{sloman2015causality}
S.~A. Sloman and D.~Lagnado.
\newblock Causality in thought.
\newblock \emph{Annual Review of Psychology}, 66\penalty0 (1), 2015.

\bibitem[Stiennon et~al.(2020)Stiennon, Ouyang, Wu, Ziegler, Lowe, Voss, Radford, Amodei, and Christiano]{stiennon2020learning}
N.~Stiennon, L.~Ouyang, J.~Wu, D.~M. Ziegler, R.~Lowe, C.~Voss, A.~Radford, D.~Amodei, and P.~F. Christiano.
\newblock Learning to summarize with human feedback.
\newblock In \emph{Advances in Neural Information Processing Systems 33: Annual Conference on Neural Information Processing Systems 2020, NeurIPS 2020, December 6-12, 2020, virtual}, 2020.

\bibitem[Team(2024{\natexlab{a}})]{geminiteam2024gemini15unlockingmultimodal}
Team.
\newblock Gemini 1.5: Unlocking multimodal understanding across millions of tokens of context, 2024{\natexlab{a}}.

\bibitem[Team(2024{\natexlab{b}})]{gemmateam2024gemma2}
G.~Team.
\newblock Gemma 2: Improving open language models at a practical size.
\newblock \emph{ArXiv preprint}, abs/2408.00118, 2024{\natexlab{b}}.

\bibitem[Tenenbaum et~al.(2006)Tenenbaum, Griffiths, and Kemp]{tenenbaum2006theory}
J.~B. Tenenbaum, T.~L. Griffiths, and C.~Kemp.
\newblock Theory-based bayesian models of inductive learning and reasoning.
\newblock \emph{Trends in Cognitive Sciences}, 10\penalty0 (7), 2006.
\newblock ISSN 1364-6613.
\newblock Special issue: Probabilistic models of cognition.

\bibitem[Tenenbaum et~al.(2011)Tenenbaum, Kemp, Griffiths, and Goodman]{josh2011how}
J.~B. Tenenbaum, C.~Kemp, T.~L. Griffiths, and N.~D. Goodman.
\newblock How to grow a mind: Statistics, structure, and abstraction.
\newblock \emph{Science}, 331\penalty0 (6022), 2011.

\bibitem[Touvron et~al.(2023)Touvron, Martin, Stone, Albert, Almahairi, Babaei, Bashlykov, Batra, Bhargava, Bhosale, et~al.]{touvron2023llama}
H.~Touvron, L.~Martin, K.~Stone, P.~Albert, A.~Almahairi, Y.~Babaei, N.~Bashlykov, S.~Batra, P.~Bhargava, S.~Bhosale, et~al.
\newblock Llama 2: Open foundation and fine-tuned chat models.
\newblock \emph{ArXiv preprint}, abs/2307.09288, 2023.

\bibitem[Tversky and Kahneman(1974)]{tversky1974judgment}
A.~Tversky and D.~Kahneman.
\newblock Judgment under uncertainty: Heuristics and biases: Biases in judgments reveal some heuristics of thinking under uncertainty.
\newblock \emph{Science}, 185\penalty0 (4157), 1974.

\bibitem[Vaswani et~al.(2017)Vaswani, Shazeer, Parmar, Uszkoreit, Jones, Gomez, Kaiser, and Polosukhin]{vaswani2017attention}
A.~Vaswani, N.~Shazeer, N.~Parmar, J.~Uszkoreit, L.~Jones, A.~N. Gomez, L.~Kaiser, and I.~Polosukhin.
\newblock Attention is all you need.
\newblock In \emph{Advances in Neural Information Processing Systems 30: Annual Conference on Neural Information Processing Systems 2017, December 4-9, 2017, Long Beach, CA, {USA}}, 2017.

\bibitem[Vig et~al.(2020)Vig, Gehrmann, Belinkov, Qian, Nevo, Singer, and Shieber]{vig2020investigating}
J.~Vig, S.~Gehrmann, Y.~Belinkov, S.~Qian, D.~Nevo, Y.~Singer, and S.~M. Shieber.
\newblock Investigating gender bias in language models using causal mediation analysis.
\newblock In \emph{Advances in Neural Information Processing Systems 33: Annual Conference on Neural Information Processing Systems 2020, NeurIPS 2020, December 6-12, 2020, virtual}, 2020.

\bibitem[Wang et~al.(2023{\natexlab{a}})Wang, Variengien, Conmy, Shlegeris, and Steinhardt]{wang2023}
K.~R. Wang, A.~Variengien, A.~Conmy, B.~Shlegeris, and J.~Steinhardt.
\newblock Interpretability in the wild: a circuit for indirect object identification in {GPT-2} small.
\newblock In \emph{The Eleventh International Conference on Learning Representations, {ICLR} 2023, Kigali, Rwanda, May 1-5, 2023}, 2023{\natexlab{a}}.

\bibitem[Wang et~al.(2023{\natexlab{b}})Wang, Wang, Li, Gao, Yin, and Ren]{wang2023scott}
P.~Wang, Z.~Wang, Z.~Li, Y.~Gao, B.~Yin, and X.~Ren.
\newblock {SCOTT}: Self-consistent chain-of-thought distillation.
\newblock In \emph{Proceedings of the 61st Annual Meeting of the Association for Computational Linguistics (Volume 1: Long Papers)}, 2023{\natexlab{b}}.

\bibitem[Wang et~al.(2024)Wang, Bai, Peng, and Hsieh]{wang2024loss}
Y.~Wang, A.~Bai, N.~Peng, and C.-J. Hsieh.
\newblock On the loss of context-awareness in general instruction fine-tuning.
\newblock \emph{ArXiv preprint}, abs/2411.02688, 2024.

\bibitem[Wei et~al.(2022{\natexlab{a}})Wei, Bosma, Zhao, Guu, Yu, Lester, Du, Dai, and Le]{wei2021finetuned}
J.~Wei, M.~Bosma, V.~Y. Zhao, K.~Guu, A.~W. Yu, B.~Lester, N.~Du, A.~M. Dai, and Q.~V. Le.
\newblock Finetuned language models are zero-shot learners.
\newblock In \emph{The Tenth International Conference on Learning Representations, {ICLR} 2022, Virtual Event, April 25-29, 2022}, 2022{\natexlab{a}}.

\bibitem[Wei et~al.(2022{\natexlab{b}})Wei, Wang, Schuurmans, Bosma, Ichter, Xia, Chi, Le, and Zhou]{wei2022chain}
J.~Wei, X.~Wang, D.~Schuurmans, M.~Bosma, B.~Ichter, F.~Xia, E.~H. Chi, Q.~V. Le, and D.~Zhou.
\newblock Chain-of-thought prompting elicits reasoning in large language models.
\newblock In \emph{Advances in Neural Information Processing Systems 35: Annual Conference on Neural Information Processing Systems 2022, NeurIPS 2022, New Orleans, LA, USA, November 28 - December 9, 2022}, 2022{\natexlab{b}}.

\bibitem[Wong et~al.(2023)Wong, Grand, Lew, Goodman, Mansinghka, Andreas, and Tenenbaum]{wong2023word}
L.~Wong, G.~Grand, A.~K. Lew, N.~D. Goodman, V.~K. Mansinghka, J.~Andreas, and J.~B. Tenenbaum.
\newblock From word models to world models: Translating from natural language to the probabilistic language of thought.
\newblock \emph{ArXiv preprint}, abs/2306.12672, 2023.

\bibitem[Xu and Tenenbaum(2007)]{xu2007word}
F.~Xu and J.~B. Tenenbaum.
\newblock Word learning as {B}ayesian inference.
\newblock \emph{Psychological Review}, 114\penalty0 (2), 2007.

\bibitem[Yang et~al.(2024{\natexlab{a}})Yang, Yang, Zhang, Hui, Zheng, Yu, Li, Liu, Huang, Wei, et~al.]{yang2024qwen2}
A.~Yang, B.~Yang, B.~Zhang, B.~Hui, B.~Zheng, B.~Yu, C.~Li, D.~Liu, F.~Huang, H.~Wei, et~al.
\newblock Qwen2. 5 technical report.
\newblock \emph{arXiv e-prints}, 2024{\natexlab{a}}.

\bibitem[Yang et~al.(2024{\natexlab{b}})Yang, Zhang, Xu, Lu, Heng, and Lam]{yang2024unveiling}
H.~Yang, Y.~Zhang, J.~Xu, H.~Lu, P.-A. Heng, and W.~Lam.
\newblock Unveiling the generalization power of fine-tuned large language models.
\newblock In \emph{Proceedings of the 2024 Conference of the North American Chapter of the Association for Computational Linguistics: Human Language Technologies (Volume 1: Long Papers)}, 2024{\natexlab{b}}.

\bibitem[Yao et~al.(2022)Yao, Chen, Yang, and Narasimhan]{yao2022webshop}
S.~Yao, H.~Chen, J.~Yang, and K.~Narasimhan.
\newblock Webshop: Towards scalable real-world web interaction with grounded language agents.
\newblock In \emph{Advances in Neural Information Processing Systems 35: Annual Conference on Neural Information Processing Systems 2022, NeurIPS 2022, New Orleans, LA, USA, November 28 - December 9, 2022}, 2022.

\bibitem[Ying et~al.(2024)Ying, Zhi-Xuan, Wong, Mansinghka, and Tenenbaum]{ying2024grounding}
L.~Ying, T.~Zhi-Xuan, L.~Wong, V.~Mansinghka, and J.~Tenenbaum.
\newblock Grounding language about belief in a bayesian theory-of-mind.
\newblock In \emph{Proceedings of the Annual Meeting of the Cognitive Science Society}, volume~46, 2024.

\bibitem[Yu et~al.(2024)Yu, Xu, Weston, and Kulikov]{yu2distilling}
P.~Yu, J.~Xu, J.~E. Weston, and I.~Kulikov.
\newblock Distilling system 2 into system 1.
\newblock In \emph{The First Workshop on System-2 Reasoning at Scale, NeurIPS'24}, 2024.

\bibitem[Zhao et~al.(2025)Zhao, Hong, Liu, Hazarika, and Lin]{zhao2025do}
S.~Zhao, M.~Hong, Y.~Liu, D.~Hazarika, and K.~Lin.
\newblock Do {LLM}s recognize your preferences? evaluating personalized preference following in {LLM}s.
\newblock In \emph{The Thirteenth International Conference on Learning Representations}, 2025.

\bibitem[Zhu and Griffiths(2024)]{zhu2024incoherent}
J.-Q. Zhu and T.~Griffiths.
\newblock Incoherent probability judgments in large language models.
\newblock In \emph{Proceedings of the Annual Meeting of the Cognitive Science Society}, volume~46, 2024.

\end{thebibliography}

\pagebreak
\appendix

\section{Experimental Details}
\label{app:methods}
\subsection{Simulated Users in the Flight Recommendation Task}
In each round, we presented a set of $k$ flight options $\mathcal{O} = \{o_1, ..., o_k\}$ to both the simulated user and the assistant (typically $k=3$).
Each flight has a departure time, a duration, a number of stops, and a cost; these four features are encoded in a vector $\bm{\phi}(o) \in \mathbb{R}^4$. For each flight option, each feature can take one of 11 values uniformly distributed between 0 and 1, except for the number of stops, which has 3 values. This defines $3 \times 11^3$ unique flight options. We converted these four numbers into a textual description illustrated in Fig. 1.

The user's preferences are defined by a reward function $\bm{\theta}$ parameterized by four numbers, which indicate the user's preferences for the aforementioned features. The space $\Theta$ of reward functions includes all four-dimensional vectors with the values $\{-1, -0.5, 0, 0.5, 1\}$, where $-1$ corresponds to a preference for low values of this feature (e.g., short flights) and $1$ to a preference for high values (e.g., long flights). Given a set of flight options $\mathcal{O}$, the user computes the reward $r(o; \bm{\theta}) = \bm{\theta^{T}}\bm{\phi}(o)$ of each flight $o$, and chooses the flight with the highest reward: 
\begin{align}
    o^*(\mathcal{O}, \bm{\theta}) = \textrm{argmax}_{o \in \mathcal{O}} r(\mathcal{O}; \bm{\theta}).
\label{eq:user_argmax}
\end{align}

When there was a tie between multiple options, we randomly selected one of the options that had the highest reward.
We excluded the reward function $(0, 0, 0, 0)$, that is, the completely indifferent user. This results in a total of $5^4 - 1 = 624$ possible reward functions, corresponding to 624 simulated users. We note that these simulated users are highly simplified and are not meant to capture the full complexity of humans: humans do not always choose the option that maximizes their utility \citep{koehler2010probability}, and their preferences may evolve over time.

\subsection{The Bayesian Assistant} Since the space of reward functions is relatively small, we were able perform exact Bayesian updates. In each round, given options $\mathcal{O}$ and the user's preferred option $o^*$, the Bayesian Assistant updates its posterior as follows: 
\begin{align}
q^{i+1}_{B}(\bm{\theta} | \mathcal{O}^{i+1}, o^{*i+1}) = \frac{p(o^{*i+1} | \bm{\theta}, \mathcal{O}^{i+1})q^{i}_{B}(\bm{\theta})}{p(o^{*i+1} | \mathcal{O}^{i+1})}, 
\label{eq:bayes_update}
\end{align}
where the likelihood function indicates whether the reward function is consistent with the user's choice: 
\begin{align}
p(o^* | \bm{\theta}, \mathcal{O}) = \mathbbm{1} \big [  \max_{o \in \mathcal{O}} r(o; \bm{\theta}) = o^* ]. 
\end{align}

The Bayesian Assistant then makes flight recommendations based on its reward posterior mean, $\hat{\bm{\theta}} = \mathbb{E}_{q(\bm{\theta})} [\bm{\theta}]$, following Equation~\ref{eq:user_argmax}. 
In most experiments, we used the uniform prior (for experiments with other priors, see Supplementary Fig. C10b).

\subsection{LLMs}
Our main experiments focus on the instruction-tuned versions of open-weights models, including models from the Gemma 2~\citep{gemmateam2024gemma2}, Llama 3~\citep{dubey2024llama3herdmodels}, and Qwen 2.5 ~\citep{yang2024qwen2} families. We used Gemma 2 models with 9B parameters (\url{https://huggingface.co/google/gemma-2-9b-it}) and 27B parameters (\url{https://huggingface.co/google/gemma-2-27b-it}), Llama 3 models with 8B parameters (\url{https://huggingface.co/meta-llama/Llama-3.1-8B-Instruct}) and 70B parameters (\url{https://huggingface.co/meta-llama/Llama-3.3-70B-Instruct}), and Qwen 2.5 models with 7B paramters (\url{https://huggingface.co/Qwen/Qwen2.5-7B-Instruct}) and 32B parameters (\url{https://huggingface.co/Qwen/Qwen2.5-32B-Instruct}).
We also evaluated Gemini 1.5 Pro~\citep{geminiteam2024gemini15unlockingmultimodal} and GPT-4.1 Mini~\citep{openai2025gpt41}, which can only be accessed through an API, as representatives of stronger models whose weights are not accessible. All of the models we use are based on the Transformer neural network architecture \citep{vaswani2017attention}. We used greedy decoding (temperature of 0) for all experiments. 

\subsection{Generalization Tasks}

For the variants of the flight recommendation task (see ``Fine-tuned LLMs generalize to new tasks''), we varied the number of flight features, ranging from two to eight features. The full flight features include the following features, in addition to the above four: arrival time, layover duration, cancellation policy, and number of bags. As the number of possible reward functions grows exponentially with the number of features, we randomly sampled up to 1,000 reward functions (simulated users) for each number of features.

For the hotel recommendation task, the hotel features include distance to downtown, price, rating, and amenities. For each hotel option, the distance to downtown and price take one of 11 values uniformly distributed between 0 and 1, while rating and amenities take one of 5 values uniformly distributed between 0 and 1, resulting in $5 \times 5 \times 11^2$ unique hotel options. We evaluated $624$ different simulated users, as in the flight recommendation task.

For the web shopping task, we used real-world products that are publicly available at \url{https://webshop-pnlp.github.io}. We chose the 100 categories with the most products. Each product is described by a title and bullet point descriptions, whose length is limited to 800 characters. The reward of a user for a product was calculated based on text-matching heuristics on product attributes and options, following \citet{yao2022webshop}. For each category, we randomly sampled 10 users, each consisting of five-round interactions. Performance was evaluated on 100 held-out option sets within the same category. 

To reduce the sensitivity of the results to the specific randomly selected option sets, we averaged all experiments over three random seeds for flight and hotel recommendations, and over all categories for web shopping. In each case, we report the mean and the standard error across runs and evaluation seeds.

\subsection{LLM Fine-Tuning}
We used the instruction-tuned version of Gemma~2 9B, Llama 3 8B, and Qwen 2.5 7B for all fine-tuning experiments. For each reward function, we generated 10 user--assistant interactions, resulting in $624 \times 10 = 6,240$ fine-tuning examples, each with five-round interactions. We experimented with fine-tuning on more examples but did not observe any significant improvement.  The interactions were formatted as shown in Supplementary Table H3. 

We used full fine-tuning (i.e. all parameters were updated) with a learning rate of \mbox{2e-6}, a batch size of 128, and a maximum sequence length of 2048, for 1 epoch. The models were fine-tuned using the standard language modeling objective, i.e., the cross-entropy loss between the model's predicted token probabilities and the ground-truth tokens in the training data. The loss was only computed on the model's responses. For each setup, we trained three models with different random seeds. We conducted all fine-tuning experiments using 4 $\times$ H100 GPUs based on the standard recipe (\url{https://github.com/huggingface/alignment-handbook}). Fine-tuning Gemma 2 9B, Llama 3 8B and Qwen 2.5 7B required about an hour for each model.

\subsection{Human Annotations}
We collected two sets of human annotations for the flight recommendation task: one where the annotators act as assistants and one where they act as users. The human annotators were recruited online and paid the market rate of \$12 an hour, as suggested by the Prolific platform~\citep{palan2018prolific} we used to recruit participants. See details in Supplementary Section E.

The annotation setup for the assistant role follows the evaluation setup we used for LLMs. In each round, the annotator was asked to make recommendations from three flight options, with each represented in the same format shown to the LLMs. After making their recommendation, the annotator received feedback indicating whether their choice was correct. They were then directed to a preference questionnaire, where they provided their estimates of the user's preferences for each individual feature (see annotation interface in Supplementary Fig. G17).
We sampled 48 reward functions by first grouping them based on the L2 distance between their four-dimensional parameter vector and the origin, then sampling from each group proportionally to its size. We had 15 separate participants provide annotations for each of the 48 simulated users (720 human participants in total).

When the annotator serves in the user role, we first asked them to rate their own preferences for different flight features; this serves as their reward function. Then, the annotator was asked to select their preferred option out of three flight options based on their preferences; this was repeated for five rounds. We constructed 50 such lists of five rounds of flights options, and had 10 annotators produce annotations for each of these 50 lists (500 human participants in total). We then produced three randomly shuffled variants of each of the interactions, for a total of 2000 interactions (500 original interactions and $3 \times 500$ shuffled interactions). This ensures that a particular option set is not consistently at a particular point in the interaction (for example, at the end of the interaction, where the participants may be paying less attention). To ensure quality, we required annotators to think for at least 30 seconds before making their selection.

\section{Can LLMs Accurately Verbalize Their Beliefs?}
\label{sec:app_a_verbalize_beliefs}

The results of the fine-tuning experiments described in the main text suggest that fine-tuned models are able to infer the user's preferences, at least implicitly. Here, we test if the LLMs can verbalize their beliefs about the user's preferences, based on the user's previous booking history which is provided as context. 

\subsection{Eliciting Beliefs About User Preferences}
\label{sec:app_a1_elicit_beliefs}

We elicit beliefs in one of two ways. For the open-weights models (Gemma 2, Llama 3, and Qwen 2.5), for which we have access to the probability distribution over upcoming words, we employ continuation scoring, as follows. After interacting with the LLM for one or more rounds, the user asks the LLM for its beliefs about the user's preferences, for example, ``on a scale of 1 to 5, what is my preference for price?'', where $1$~indicates a strong preference for cheaper flights, $3$~indicates no strong preference, and $5$~indicates a strong preference for expensive flights.
We score the numbers 1, 2, 3, 4, and 5 as possible continuations of the current text and re-normalize them to form a probability distribution over these five numbers (see Table~\ref{tab:prompt_reward} for a detailed example). 

For closed-weights models (Gemini 1.5 Pro and GPT-4.1 Mini), where the LLM's underlying probability distribution over upcoming words is not made accessible to researchers, we ask the LLM to explicitly provide a probability distribution over each of the five points on the scale (see Table~\ref{tab:prompt_reward_generation}; for a comparison between the scoring and explicit probability judgment method in Gemma~2, which finds that scoring yields more accurate estimates, see Section \ref{sec:app_reward_ablation}). For our human participants, we use a similar survey to the one we use for closed-weights models. 

We then approximate the distribution over reward functions as a factorization of these feature distributions:
\begin{align}
q_{LLM}^{i}(\bm{\theta} | \mathcal{O}^{i}, o^{*i}) \approx \prod_{j} q_{LLM}^{i}(\bm{\theta}_j | \mathcal{O}^{i}, o^{*i}, c_j^{i}).
\label{eq:lm_posterior}
\end{align}
where $q_{LLM}^{i}(\bm{\theta}_j | \mathcal{O}^{i}, o^{*i}, c_j^{i})$ is the probability that the LLM assigns to each of the user's preferences for feature $j$ given the current context in the prompt $c_j^{i}$, using either scoring or explicit probability judgement.
This makes an independence assumption, whereby the preference for one feature does not interact with the preference for another; because this assumption is quite strong, we cannot guarantee that it provides a complete picture of the LLM's beliefs over all possible reward functions. We elicit the LLM's beliefs by prompting it; it is possible that other techniques, such as probing, where a classifier is trained to decode the model's internal activations, could yield different results. We leave a more systematic study of this question for future work.

\subsection{Evaluating the LLM’s Verbalized Beliefs}
\label{sec:app_a2_evaluate_beliefs}
\begin{figure}[t!]
    \centering
    \includegraphics[width=0.5\linewidth]{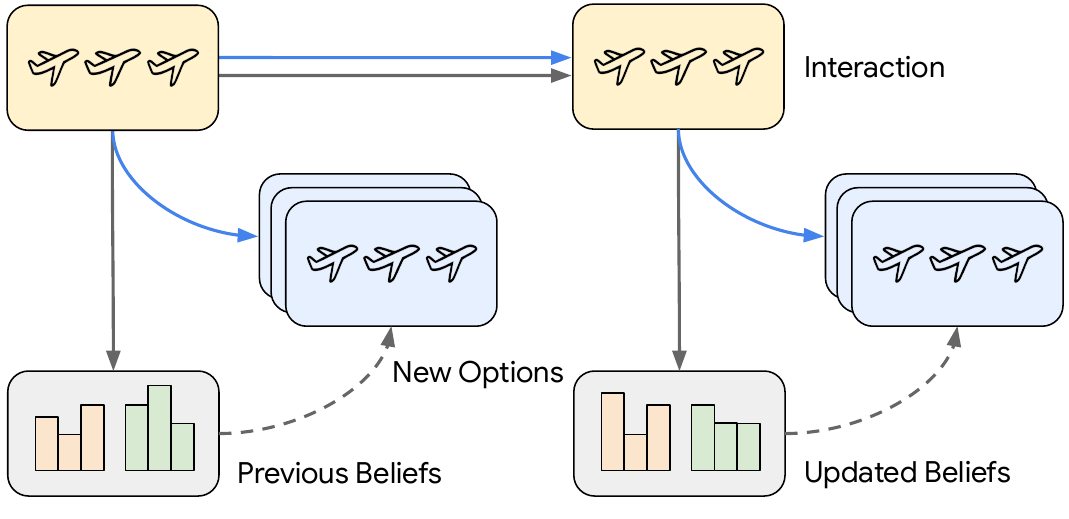}
    \vspace{+1em}
    \caption{\textbf{Experimental design for LLM evaluation.} At the end of each round, we evaluate the LLM using new option sets for which it has not received feedback. The evaluation branches out from the main interactions (that is, the evaluation performed after the first round is not included in the context of the second round). The LLM's direct evaluation, where we ask the LLM directly to choose a flight, follows the blue lines; the belief-based evaluation, where we first assess the LLM's beliefs about the user's preferences and then use them to choose the flight, follows the gray lines. The dashed lines indicate the deterministic conversion of the LLM's beliefs into flight recommendations.}
    \label{fig:app_figa1_eval}
\end{figure}

To determine whether the LLM can accurately verbalize its beliefs about the user's preferences, we derive flight recommendations from the LLM's verbalized beliefs, using the same procedure that the Bayesian Assistant uses to make recommendations based on its explicit beliefs, and evaluate the accuracy of these recommendations. We follow the same evaluation setup as our main experiments, except at the end of each round, we query the LLM's beliefs about the user's preferences. Importantly, this process branches out from the main dialogue, ensuring the interaction remains unaffected (Fig.~\ref{fig:app_figa1_eval}).
We also examine whether the recommendations produced in this way are consistent with the recommendations made by the LLM directly. High consistency between these two measures would suggest that the LLM's verbalized beliefs align with the implicit internal beliefs used by the LLM to make predictions in the original setup.

We also perform an analogous analysis for the experiment where human participants acted as the assistant to a simulated user. Recall that after each round we asked the participants what they thought the user's preferences were. We use these verbalized beliefs about the user's preferences as input to the same computation we used for the LLMs' verbalized beliefs. As with the LLMs, we can compute the consistency between the flight choices derived in this way and the participants' choices in the experiment. We only evaluated on the five-round interactions of the 48 simulated users for which we have human annotations.

\begin{figure*}[t!]
    \centering
    \includegraphics[width=\linewidth]{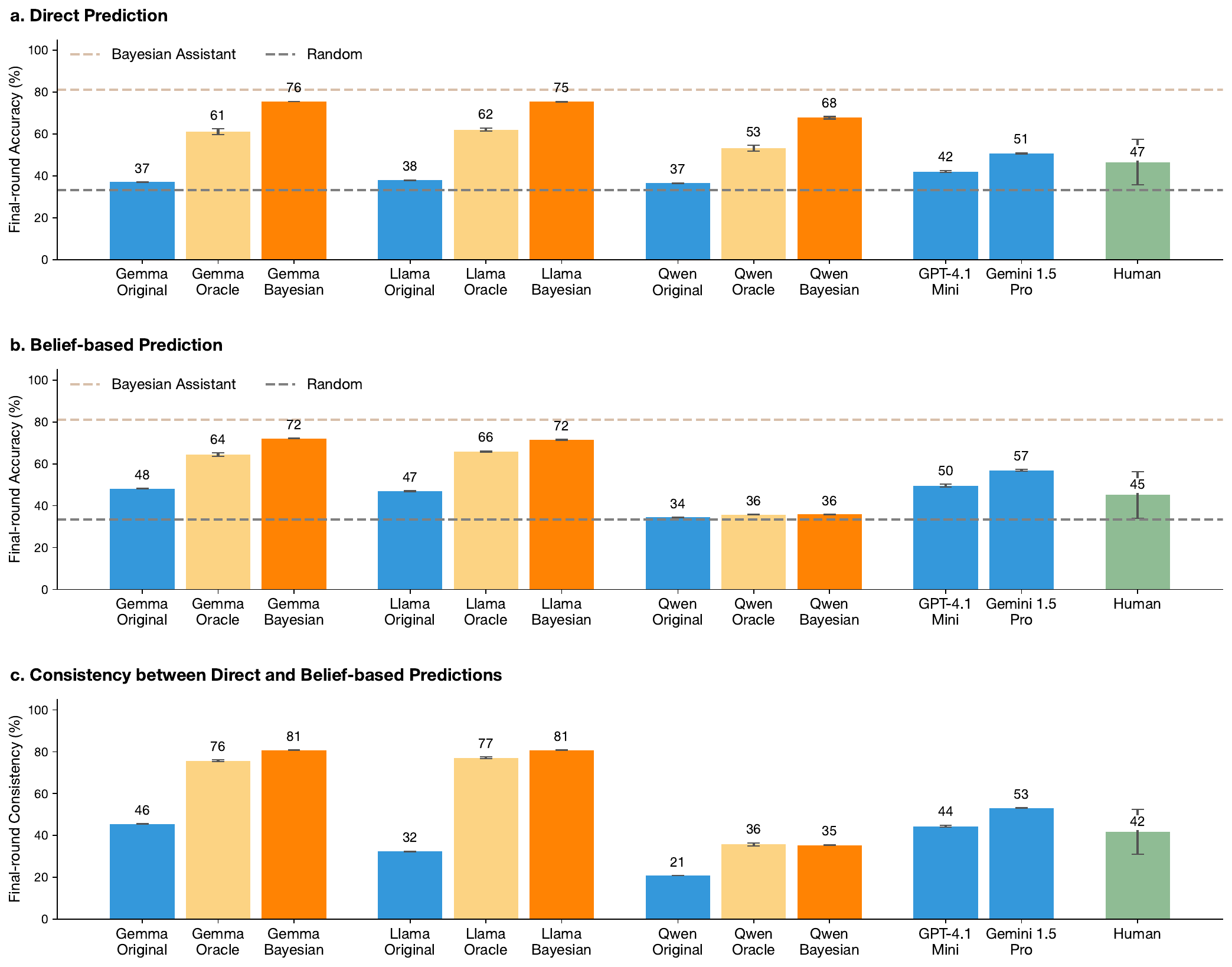}
    \begin{subfigure}[b]{0pt}
        \phantomsubcaption
        \label{fig:app_figa2a_direct_acc}
    \end{subfigure}
    \begin{subfigure}[b]{0pt}
        \phantomsubcaption
        \label{fig:app_figa2b_belief_acc}
    \end{subfigure}
    \begin{subfigure}[b]{0pt}
        \phantomsubcaption
        \label{fig:app_figa2c_belief_consistency}
    \end{subfigure}
    \vspace{-1em}
    \caption{\textbf{Comparison of direct accuracy and belief-based accuracy.} We show final-round accuracy based on \textbf{(a)} the LLM's or human's direct predictions and \textbf{(b)} predictions derived from their verbalized beliefs about the user's preferences. The gray dashed line indicates random performance, while the brown dashed line indicates the performance of the Bayesian Assistant. For human participants, we only evaluate on a subset of our evaluation data, which includes 48 different simulated users. \textbf{(c)} Final-round consistency between two predictions: the prediction directly provided by the LLM or human participants and the prediction derived from their beliefs about the user's preferences. Fine-tuned LLMs show better consistency than the original LLMs, with Bayesian LLMs achieving the highest consistency. Error bars show standard error across participants for humans and across three random seeds (and three training runs) for LLMs.}
    \label{fig:app_figa2_belief_vs_direct}
\end{figure*}

\subsection{Results}
\label{sec:app_a3_results}
For the original LLMs, we find that the approach described in this section---where we first estimate the LLMs' beliefs about the user's preferences by explicitly querying the LLMs, and then use a decision-making component that is external to the LLM---performs better than directly using the LLMs' predictions (Fig.~\ref{fig:app_figa2b_belief_acc} vs. Fig.~\ref{fig:app_figa2a_direct_acc}, original LLMs). 
We also find that the original LLMs' direct predictions are often inconsistent with the belief-based predictions (those derived from the beliefs elicited from the LLMs), with less than 50\% alignment between the two sets of predictions (Fig.~\ref{fig:app_figa2c_belief_consistency}). Human participants similarly show high inconsistency between the two types of predictions.

Predictions based on the fine-tuned LLMs' verbalized beliefs are substantially more accurate than those based on the original LLM's verbalized beliefs, except for Qwen 2.5 models (Fig.~\ref{fig:app_figa2a_direct_acc} and Fig.~\ref{fig:app_figa2b_belief_acc}, Oracle LLMs and Bayesian LLMs). This suggests that both training methods teach the LLM to infer the user's preferences and update them as more information becomes available, even though neither method provides the model with direct access to users' preferences. 
For both Gemma 2 and Llama 3, the Bayesian variant of the LLMs produces more accurate estimates of the user's beliefs than the Oracle one. Likewise, compared to the Oracle variants, the Bayesian variants achieve higher consistency between the predictions directly provided by the LLM and those derived from the LLM's verbalized beliefs. The difference in overall accuracy between these models' direct predictions and belief-based predictions is much smaller after fine-tuning. This trend, however, does not hold for Qwen 2.5 model: while direct prediction accuracy improves after fine-tuning, belief-based prediction accuracy remains unchanged from the original LLM. This suggests that for Gemma~2 and Llama~3, prompt-based prediction elicitation may tap into a representation that is shared with the computation used to make direct predictions, but that is not the case for Qwen~2.5.

\section{Additional Results}
\label{sec:app_b_additional_results}

\subsection{The Original LLMs' Poor Performance is Robust to Evaluation Setup}
\label{sec:app_b1_eval_ablations}

\begin{figure*}[t!]
    \centering
    \includegraphics[width=\linewidth]{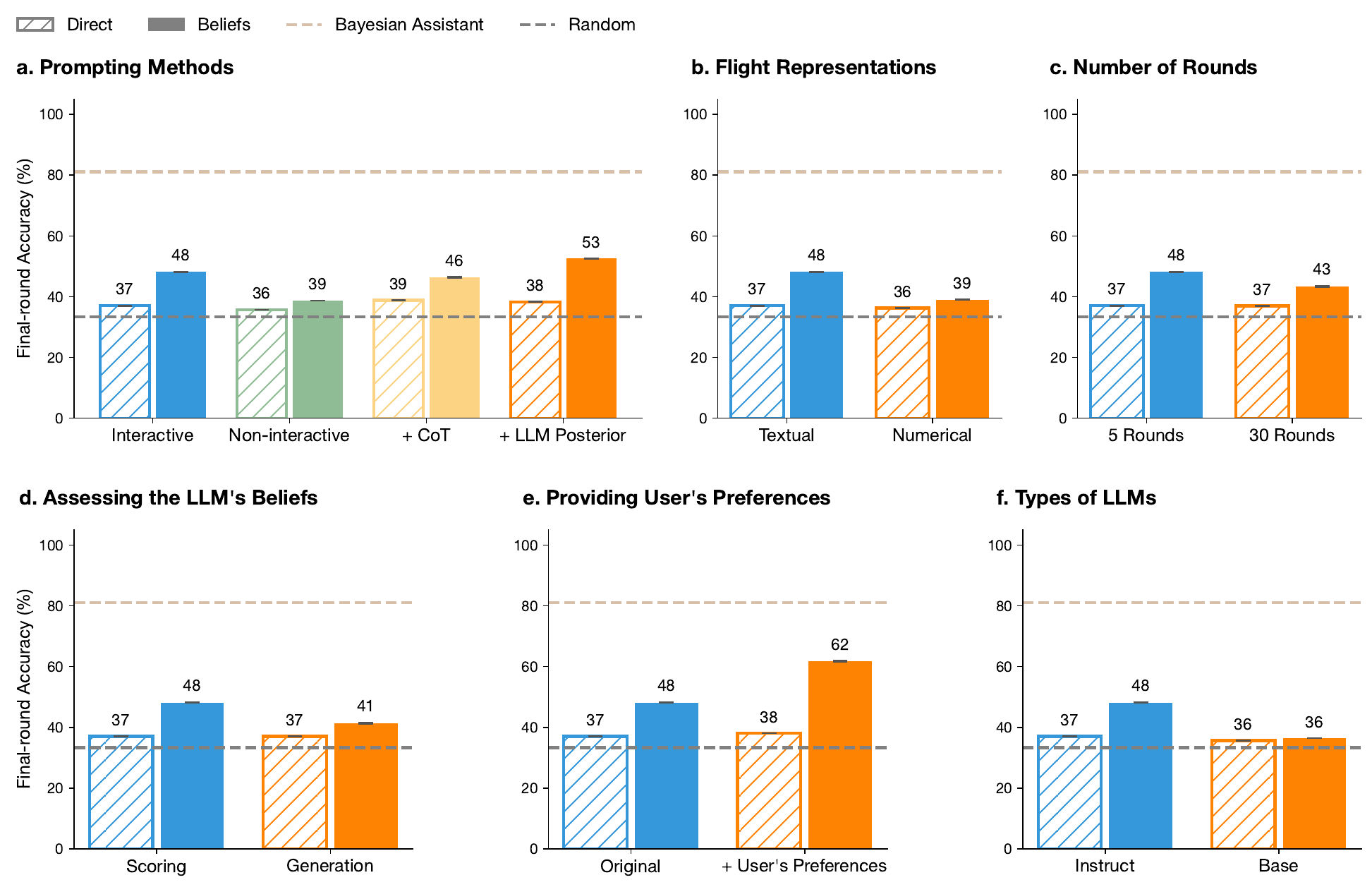}
    \begin{subfigure}[b]{0pt}
        \phantomsubcaption
        \label{fig:app_figb3a_prompt_method}
        \end{subfigure}
    \begin{subfigure}[b]{0pt}
        \phantomsubcaption
        \label{fig:app_figb3b_representation}
    \end{subfigure}
    \begin{subfigure}[b]{0pt}
        \phantomsubcaption
        \label{fig:app_figb3c_rounds}
    \end{subfigure}
    \begin{subfigure}[b]{0pt}
        \phantomsubcaption
        \label{fig:app_figb3d_reward_type}
    \end{subfigure}
    \begin{subfigure}[b]{0pt}
        \phantomsubcaption
        \label{fig:app_figb3e_bayesian_posterior}
    \end{subfigure}
    \begin{subfigure}[b]{0pt}
        \phantomsubcaption
        \label{fig:app_figb3f_base}
    \end{subfigure}
    \vspace{-1em}
    \caption{\textbf{Final-round accuracy of Gemma Original under different variations of our experimental setup.} We report both the model's direct predictions (hatched bars) and the predictions derived from the model's verbalized beliefs (solid bars; Supplementary Section~\ref{sec:app_a_verbalize_beliefs}). \textbf{(a)} We compare the original interactive setting, where we directly ask the LLM to generate predictions and provide it with feedback, with other common techniques: non-interactive prompting, where we always provide correct examples; chain-of-thought (CoT) prompting, which encourages the LLM to think step-by-step; and methods that incorporate the LLM's verbalized reward posterior distribution in the context. \textbf{(b)} The textual representation of the flight options, which uses natural language descriptions deterministically generated from the feature values, compared to the numerical representation, which directly uses the feature values. \textbf{(c)} 5-round interactions between the user and the LLM compared to 30-round interactions. \textbf{(d)} The scoring method, which assesses the LLM's beliefs by scoring possible continuations, compared to the generation method, where we explicitly ask the LLM to generate probability judgments. \textbf{(e)} Performance without versus with the verbalized user's preferences from the Bayesian model. \textbf{(f)} Instruction-tuned models versus their pre-trained base models. Error bars show standard errors across three random seeds.}
    \label{fig:app_figb3_eval_ablation}
\end{figure*}

In light of the poor performance of the original LLMs (before fine-tuning), we considered various modifications to our evaluation setting. These include prompting-based methods, that is, modifications to the instructions provided to the LLM; an alternative, numerical representation of the flight options; and a greater number of interactions. We also ablate methods that access the LLM's verbalized beliefs, explore whether providing the user's preferences improves performance, and compare the instructed version of the models with their corresponding pre-trained versions. These robustness analyses focus on Gemma~2 9B.
Overall, we do not observe significant differences across these evaluations; the only methods that we find effectively improved model performance involved fine-tuning (see Section \ref{sec:app_b2_ft_ablations}).

\paragraph{Advanced prompting methods do not improve accuracy.} 

Our main experiments evaluate the LLM in an interactive setting, where the user provides it with feedback indicating whether the LLM's choice is correct. In this case, the user's feedback is always based on the LLM's prediction. We first experiment with an alternative non-interactive setting, where the context for the assistant includes all previous rounds and the option chosen by the assistant in these context rounds is always correct, a setting that better approximates the standard few-shot or in-context learning setup for LLMs (\citet{brown2020language}; see Table~\ref{tab:prompt_icl} for an example). While performance on direct prediction remains similar, we observe a performance drop when evaluating with predictions derived from the LLM's beliefs (Fig.~\ref{fig:app_figb3a_prompt_method}, ``Non-interactive').

Chain-of-thought (CoT) prompting ~\citep{wei2022chain, nye2021show, kojima2022large}, which encourages the model to generate step-by-step reasoning chains, has been shown to be effective on many reasoning tasks. We evaluate the LLM using this strategy by explicitly including reasoning hints and the phrase ``Let's think step by step'' in the instruction (see Table~\ref{tab:prompt_cot} for an example prompt). We find that LLMs prompted with CoT perform similarly to those prompted in the way described in the main text (Fig.~\ref{fig:app_figb3a_prompt_method}, ``$+$~CoT'').

Since inferring user's preferences based on current information before making recommendations is crucial in our task, we further evaluate another CoT-style two-stage prompting method, where we allow the LLM to explicitly reason over the posterior distribution over reward functions. Specifically, we verbalize the LLM's reward posterior distribution using natural language and add it to the LLM's context (see Table~\ref{tab:prompt_posterior} for an example). 
Explicitly encouraging the LLM to reason over its own reward posterior distribution improves the predictions derived from its verbalized beliefs. However, direct prediction accuracy remains similar (Fig.~\ref{fig:app_figb3a_prompt_method}, ``$+$~LLM Posterior'').

Though additional prompt engineering and advanced prompting techniques could potentially yield different results; in particular, some prompts may more effectively extract the model's beliefs than the ones we used. 
For the moment, however, our preliminary findings suggest that it is challenging to significantly improve the LLM's performance purely through prompting.

\paragraph{The LLMs' poor performance is not due to inability to parse the flight representations.} Our main experiments use a representation that deterministically maps the feature value of each flight to a textual description (e.g., the departure time may be \texttt{02:00 PM} and the duration \texttt{2 hr 30 min}). While this textual representation is closer to realistic scenarios, and may therefore better align with the LLM's training distribution, this setup introduces a potential confounder that complicates the interpretation of our results: the LLM's poor performance in the flight recommendation task could be due to its inability to translate the text description into the feature space required for probabilistic reasoning.
To control for this factor, we investigate an alternative numerical representation of the flight options, where we directly provide the LLM with numerical feature values in the same format we provide them to the Bayesian Assistant (e.g., the duration may be 0.9 instead of \texttt{16 hr 6 min}; see Table~\ref{tab:prompt_interaction_num} and Table~\ref{tab:prompt_reward_num} for examples).  We find that, if anything, the textual representation outperforms its numerical counterpart in both accuracy metrics (Fig.~\ref{fig:app_figb3b_representation}). This suggests that the LLM's poor performance cannot be attributed to an inability to parse the textual format to a numerical value.

\paragraph{Increasing the number of interactions does not improve performance.}
Our previous experiments include only five rounds of interactions between the user and the LLM. To investigate the possibility that LLMs do in fact extract information from the interaction and update their beliefs, but do so more slowly than the Bayesian Assistant, we increase the number of interactions to 30. 
We find that Gemma Original still shows similar performance; if anything, its performance is slightly worse compared to our main experiments (Fig.~\ref{fig:app_figb3c_rounds}). This indicates that simply increasing the number of interactions is unlikely to significantly improve the LLM's performance.

\paragraph{Assessing the LLM's beliefs: Scoring continuations vs. explicit probability judgments.}
\label{sec:app_reward_ablation}

In the main experiment, for the open-weights LLMs where we have access to the probabilities the LLM assigns to upcoming words, we estimate the LLM's distribution over reward functions by asking it to rate individual features and scoring the possible continuations; for flight duration, for example, we might ask it what the user's preference is on a scale of 1 to 5. We refer to this method as ``scoring''. 
Here, we compare this method to one where we instruct the LLM to assign a probability to each of the five ratings on each scale; we refer to this method as ``generation'' (see Table~\ref{tab:prompt_reward_generation} for an example). The generation method is also used for experiments with the closed-weights models, as we do not have access to these LLM's probabilities. As in the scoring method, we renormalize the probabilities to ensure that they sum to~1 (although we find that this step is typically not necessary as they already sum to~1). Overall, we find the scoring-based reward distribution, which we use in the main text for the open-weights models, is closer than the generation-based one to the ground truth distribution (Fig.~\ref{fig:app_figb3d_reward_type}; for related results, see \citet{hu2023prompting}). %

\paragraph{Can the LLM make recommendations given the user's preferences?}

The flight recommendation task requires two capabilities: inferring the user's preferences and making predictions based on these preferences. We previously showed that the original LLM performs poorly at inferring the user's preferences (Fig.~\ref{fig:app_figa2b_belief_acc}). Here, we investigate its performance in the latter one. Specifically, we provide the LLM with the verbalized reward posterior distribution from the normative Bayesian model (see Table~\ref{tab:prompt_posterior} for an example). In this case, the LLM only needs to make recommendations based on the provided preferences. We find that having access to the optimal reward posterior distribution improves belief-based accuracy; however, the direct prediction accuracy remains similar (Fig.~\ref{fig:app_figb3e_bayesian_posterior}). Although our method of presenting user's preferences to the LLM may not be optimal, these results suggest that the LLM struggles to make correct recommendations even when the user's preferences are explicitly provided.

\paragraph{Types of LLMs: Instructed model vs. pre-trained base model.}
\label{sec:app_instructed_vs_base}

We use instruction-tuned LLMs for our main experiments. As these models are trained using an additional post-training alignment stage, their behavior is likely to differ from their pre-trained base model counterparts  ~\citep{lin2023mitigating, yang2024unveiling, wang2024loss, kotha2023understanding}, though because we expect instruction-tuning to improve the models' interactive capabilities, we hypothesize that the base version of Gemma would not perform better than the instruction-tuned one. As base models are not well-suited to interactive evaluation, we evaluate them using the non-interactive setting by providing them with in-context examples (see earlier in this section). We find that the base model performs comparably to the instruction-tuned one (Fig.~\ref{fig:app_figb3f_base}); we omit the results for Llama~3 and Qwen~2.5, which were similar.

\subsection{Modifications to Training Setup}
\label{sec:app_b2_ft_ablations}

This Supplementary describes variants on the methods we used to fine-tune the LLMs on interactions with users. We only explore these variants for Bayesian teaching, which was consistently more effective than oracle teaching). We use Gemma 2 9B for all of the follow-up experiments reported in this section.

\begin{figure*}[t!]
    \centering
    \includegraphics[width=\linewidth]{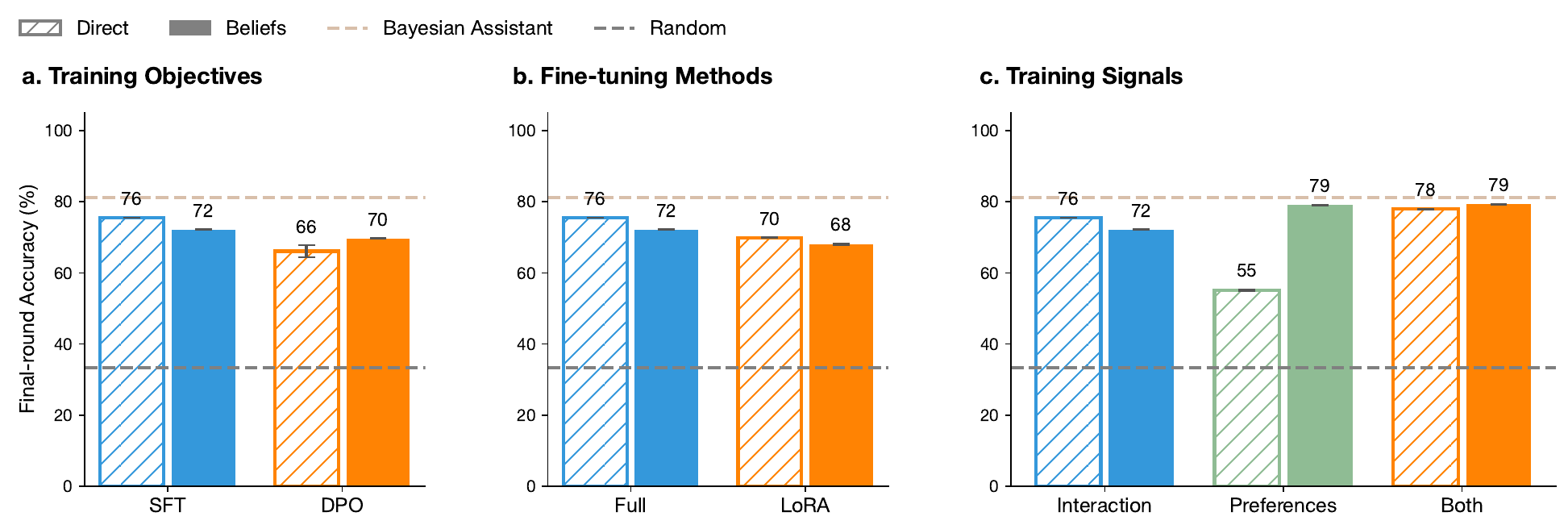}
    \begin{subfigure}[b]{0pt}
    \phantomsubcaption
    \label{fig:app_figb4a_objective}
    \end{subfigure}
    \begin{subfigure}[b]{0pt}
    \phantomsubcaption
    \label{fig:app_figb4b_ft_method}
    \end{subfigure}
    \begin{subfigure}[b]{0pt}
    \phantomsubcaption
    \label{fig:app_figb4c_data}
    \end{subfigure}
    \vspace{-1em}
    \caption{\textbf{Final-round accuracy of LLMs fine-tuned with different training strategies on the flight recommendation task}. We use Bayesian teaching (i.e. users' interactions with the Bayesian Assistant) for all experiments. \textbf{(a)} Comparison of training objectives: supervised fine-tuning (SFT) vs. direct preference optimization (DPO). \textbf{(b)} Fine-tuning methods: full fine-tuning, which updates all model parameters, vs. LoRA fine-tuning, a parameter-efficient method that only updates partial parameters. \textbf{(c)} Training only on interactions between users and assistants, as in our other experiments, compared to training on the Bayesian Assistant's estimate of the user's preferences, as well as training on both interactions and the estimated preferences. Error bars show standard errors across three random seeds and three training runs.}
    \label{fig:app_figb4_ft_ablation}
\end{figure*}

\paragraph{Training objective: Supervised fine-tuning vs. Direct preference optimization.} In most of our experiments, we use supervised fine-tuning (SFT) to provide the oracle and Bayesian predictions. In this method, the LLM is trained to predict the upcoming token in the interaction, the same objective used during pre-trainining. Here, we examine the utility of reinforcement learning from human feedback (RLHF; \citet{christiano2017deep, ouyang2022training, stiennon2020learning}), another common practice for adapting LLMs' behavior following pre-training, in which the LLM is instead provided with an explicit signal indicating whether an output is preferable. 
In particular, we use direct preference optimization (DPO; \citet{rafailov2023direct}), where the model is trained to assign higher probability to the preferred response than to the less preferred one. We investigate the DPO training objective by treating the Bayesian Assistant's prediction as the preferred one while using a different random recommendation as the less preferred one. We train the model with the DPO objective with a learning rate of 2e-6 and $\beta=0.1$. We find that training on Bayesian predictions works comparably for both SFT (used in main experiments) and DPO objectives (Fig.~\ref{fig:app_figb4a_objective}), indicating that the approach is robust to the choice of training objective.

\paragraph{Full vs. parameter-efficient fine-tuning.} 
In our main experiments, we update all model parameters. As this approach becomes less feasible as the model size grows, a common strategy to improve training efficiency relies on parameter-efficient fine-tuning, where only a subset of parameters is updated. We evaluate this approach using Low-Rank Adaptation (LoRA; \citet{hu2021lora}), a method that injects trainable rank decomposition matrices while keeping the original model weights frozen. We perform LoRA fine-tuning with a learning rate of 2e-5. While LoRA fine-tuning performs slightly worse than full fine-tuning (Fig.~\ref{fig:app_figb4b_ft_method}), it achieves comparable performance while significantly reducing training costs. This demonstrates that our fine-tuning strategy can be effectively applied in computationally efficient settings, which is particularly beneficial for larger LLMs.

\paragraph{Providing Bayesian preference estimates in fine-tuning.}
We have shown in the main text that fine-tuning the LLMs to make better recommendations also significantly improves its ability to infer the user's preferences, even though their supervision does not explicitly include the user's preferences. Here, we investigate a complementary setup, where we explicitly train the model to match the Bayesian Assistant's estimates of the user's preferences, but not to make flight recommendations. The Bayesian Assistant produces a posterior probability distribution over all reward functions after each round; we select the reward function with the highest posterior probability and provide it to the LLM, formatted as in Table~\ref{tab:prompt_reward}. We find that, like training on interactions, providing the user's preferences as a fine-tuning signal improves both accuracy measures, compared to Gemma Original, but the gain in direct prediction accuracy on is smaller than when we fine-tune on interactions (Fig.~\ref{fig:app_figb4c_data}). We also explore a setting where both the interactions and the preferences are provided during training; this setting leads to the best performance on both metrics, with accuracy approaching the accuracy of the Bayesian Assistant.

\section{Additional Analyses}
\label{sec:app_c_analysis}

\subsection{LLM Priors} 
\label{sec:app_c1_llm_priors}

\begin{figure*}[t!]
\centering
\includegraphics[width=\linewidth]{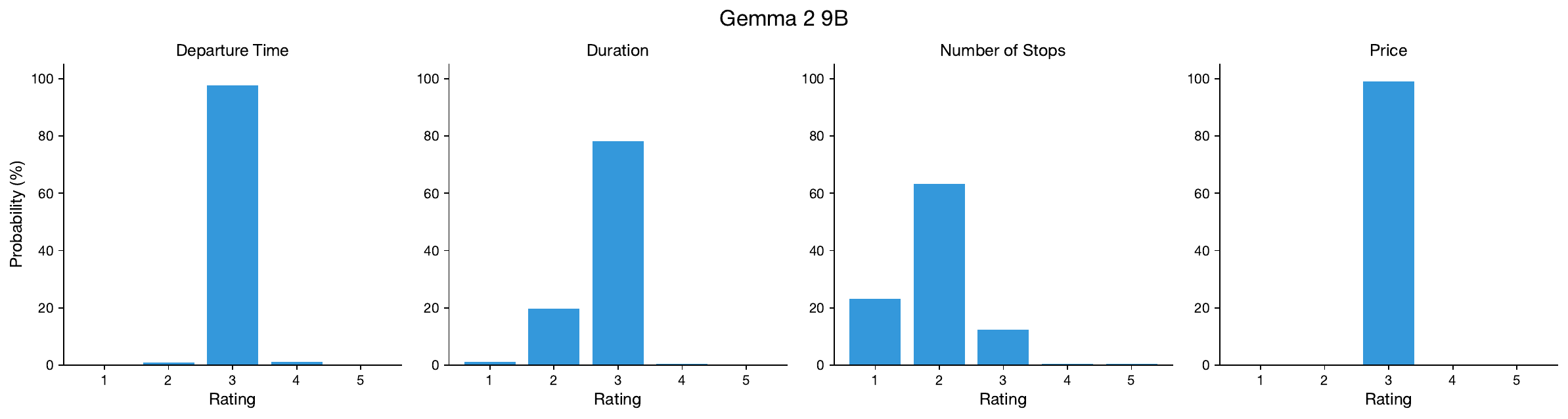}
\includegraphics[width=\linewidth]{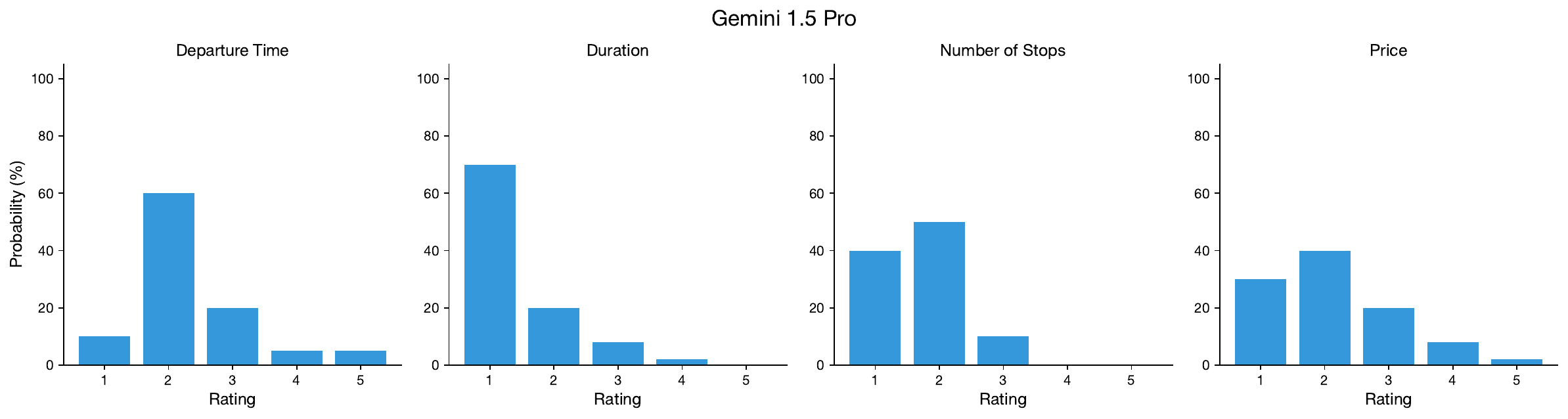}
\caption{\textbf{Priors of Gemma 2 9B Original and Gemini 1.5 Pro for each flight feature}. We obtain these priors via the prompting-based elicitation method (Supplementary~\ref{sec:app_a_verbalize_beliefs}). A rating of 1 indicates a strongest preference for the earliest departure time, the shortest duration, the fewest number of stops, and the lowest price, while a rating of 5 indicates the opposite. A rating of 3 indicates no preference.}
\label{fig:app_figc5_prior}
\end{figure*}

In the section \textit{Generalization to interactions with human users}, we find that the original LLMs, before fine-tuning, were able to provide recommendations with an accuracy substantially higher than chance even before their first interaction with the user, suggesting that the LLMs' priors are aligned with human preferences. In this section, we test this hypothesis by asking two models, Gemma~2 and Gemini~1.5, for their verbalized beliefs in advance of any interaction with a particular user. Fig.~\ref{fig:app_figc5_prior} shows the results. For Gemma~2~9B, the hypothesis is only partly supported: the prior derived from this model assigns a high probability to ``no preference'' for most of the features, with the exception of the number of stops, where it reflects a moderate preference for fewer stops. By contrast, Gemini 1.5 Pro has a more diffuse prior over these features, which favors cheaper and shorter flights, as well as flights that leave earlier in the day, plausibly reflecting the preferences of most flyers. We note that the interpretation of this pattern of results is complicated by the fact that Gemma's verbalized prior beliefs may not faithfully reflect the underlying prior it uses to make recommendations before having interacted with a user.

\begin{figure*}[t!]
    \centering
    \includegraphics[width=\linewidth]{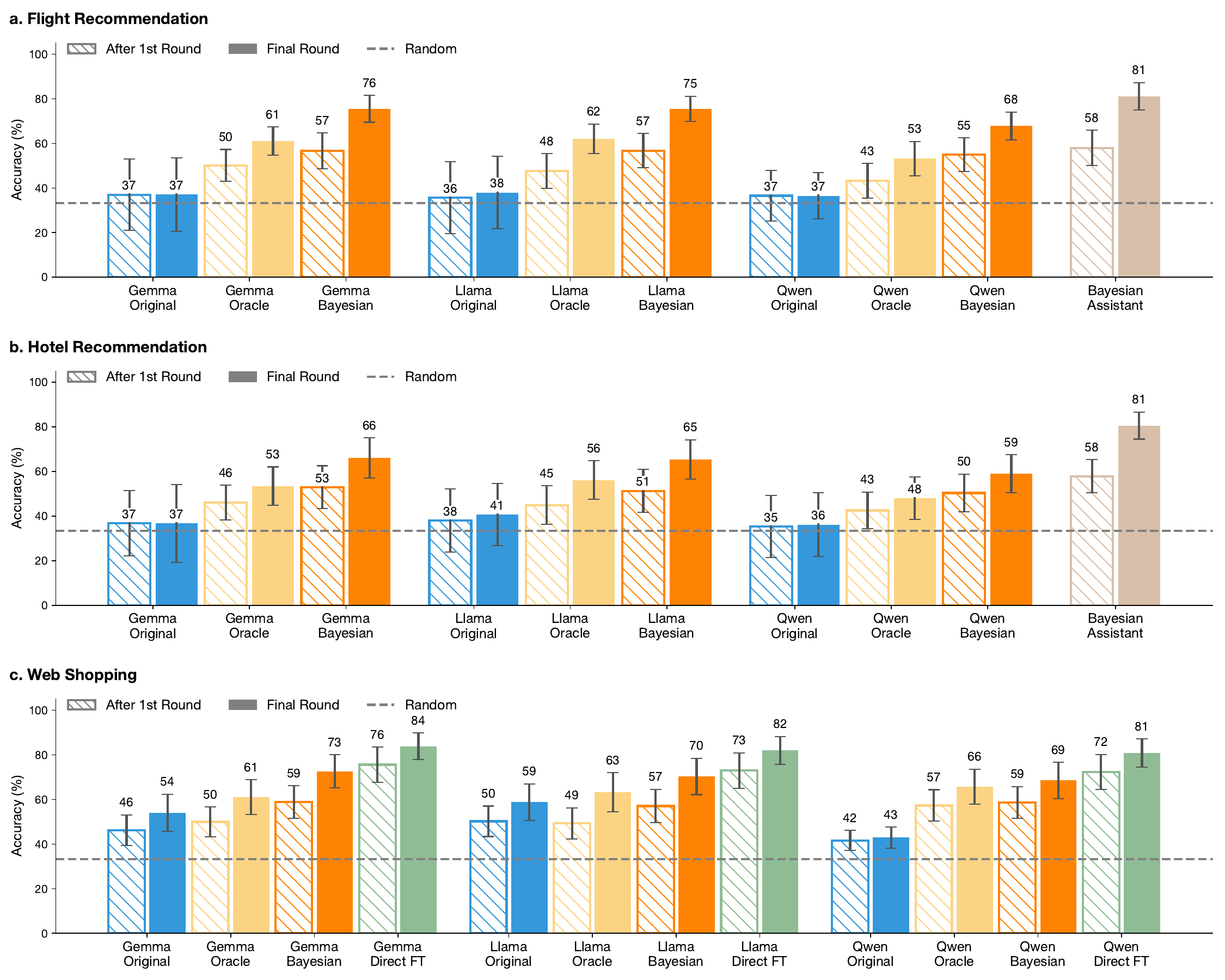}
    \caption{\textbf{Variability across simulated users.} We show accuracy after the first and final (fifth) rounds. \textbf{(a)} We compare the original LLMs, fine-tuned LLMs, and the upper bound (the Bayesian Assistant) on flight recommendation. \textbf{(b)} Comparison of LLMs and the upper bound (the Bayesian Assistant) on hotel recommendation. \textbf{(c)} Comparison of LLMs and the upper bound (LLMs fine-tuned directly on the task) for web shopping. Error bars indicate the standard deviation across reward functions (for flight and hotel recommendations) or product categories (for web shopping).}
    \label{fig:app_figc6_ft_results_by_reward_fn}
\end{figure*}

\subsection{Variability in LLM Accuracy Across Simulated Users}
\label{sec:app_c2_reward_variability}

In our main experiments, we show results averaged over all simulated users. Here, we explore how the LLM's accuracy varies by user. As before, for flight and hotel recommendations, the user is characterized as a reward function. For web shopping, we have 10 users with different goals (i.e. preferred attributes) for each category; we average their performance and compute the standard deviation across 100 product categories (see Table~\ref{tab:app_tabc1_webshop_example} for examples). All methods exhibit high variance as shown in Fig.~\ref{fig:app_figc6_ft_results_by_reward_fn}.

\begin{table}[t!]
\centering
\small
\caption{\textbf{Example product categories and their corresponding goals of different users.}}
\label{tab:app_tabc1_webshop_example}
\begin{tabularx}{0.8\textwidth}{lX} 
\toprule
Product Category                      & User's Goals (Preferred Attributes)     \\
\midrule
\multirow{3}{*}{Beds}                 & eco friendly, twin with drawers \\
                                      & wood frame, easy assemble, twin \\
                                      & memory foam, solid wood         \\
\midrule
\multirow{3}{*}{Men's athletic shoes} & running shoes, lace up          \\
                                      & non slip, mesh                  \\
                                      & daily wear, color back, size 14 \\
\midrule
\multirow{3}{*}{Food \& beverage}     & simple ingredients              \\
                                      & gluten free                     \\
                                      & low sodium                      \\
\bottomrule
\end{tabularx}
\end{table}

\begin{figure*}[t!]
    \centering
    \includegraphics[width=0.9\linewidth]{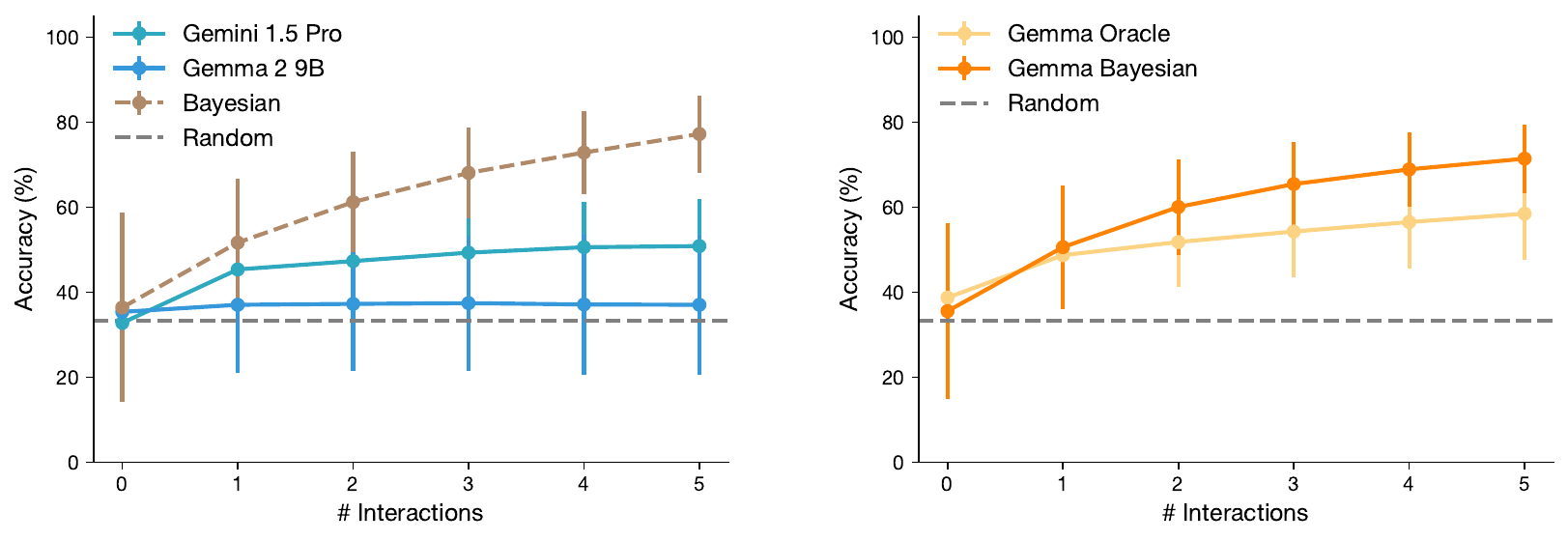}
    \caption{\textbf{Variability across reward functions over rounds.} Error bars indicate standard deviation across reward functions.}
    \label{fig:app_figc7_acc_by_rounds}
\end{figure*}

We additionally show results over rounds in Fig.~\ref{fig:app_figc7_acc_by_rounds}. We find that both original LLMs and the Bayesian Assistant display high variance across reward function.  While the variance of the Bayesian Assistant decreases as the number of interactions increases, as does that of the fine-tuned LLMs, the variance of the original LLM remains largely constant across interactions. Notably, Gemma Bayesian has lower variance while maintaining similar performance to the Bayesian Assistant.

In particular, we hypothesize that reward functions that more strongly deviate from the LLM's prior (Supplementary Section~\ref{sec:app_c1_llm_priors}) may be harder to infer. For example, the LLM may assume most people prefer shorter flights over long ones, making it more difficult to infer the preferences of an ``abnormal'' user who prefers longer flights.
To test the hypothesis that the variability across reward functions is due in part to the prior, we fit linear regression models predicting a reward function's final-round accuracy from its L2 distance to the mean of the prior reward distribution, focusing on Gemma in this experiment. We elicit the priors separately for Gemma Original, Gemma Bayesian and Gemma Oracle.
The prior of the Bayesian Assistant is uniform, as before.
Before computing distances we normalize the reward functions (divide them by their sum) to account for the fact that some functions are equivalent; for example, the reward function $[-1, -1, -1, -1]$ is equivalent to the function $[-0.5, -0.5, -0.5, -0.5]$ as both will always lead the user to prefer the same flights. 

\begin{figure*}[t!]
\centering
\includegraphics[width=\linewidth]{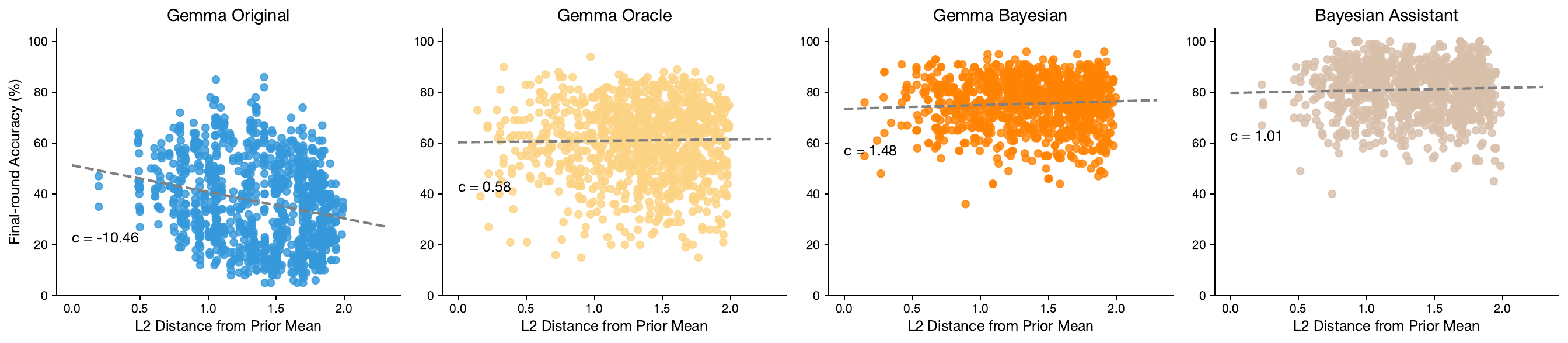}
\caption{\textbf{The relationship between the final-round accuracy and the normalized L2 distance to the mean of the prior reward distribution} (1000 randomly sampled points for readability). $c$ refers to the coefficient in a linear regression predicting accuracy from L2 distance. The impact of L2 distance on final-round accuracy is significant (p $<$ 0.001) for Gemma Original, Gemma Bayesian, and Bayesian Assistant, but not for Gemma Oracle (p = 0.24).}
\label{fig:app_figc8_acc_vs_l2}
\end{figure*}

In line with this hypothesis, we find negative regression coefficients for Gemma Original, indicating it performs worse when the reward function deviates from its priors (Fig.~\ref{fig:app_figc8_acc_vs_l2}). The absolute coefficients for the Bayesian Assistant and Gemma Bayesian are similar, and much smaller than that of Gemma Original. For these three models, the impact of L2 distance from prior on the final-round accuracy is all significant (p $<$ 0.001). The Gemma Oracle does not show sensitivity to this distance (p = 0.24).

\subsection{Interacting with Non-deterministic Users}
\label{sec:app_c3_nondeterministic_users}

\begin{figure*}[t!]
\centering
\includegraphics[width=0.45\linewidth]{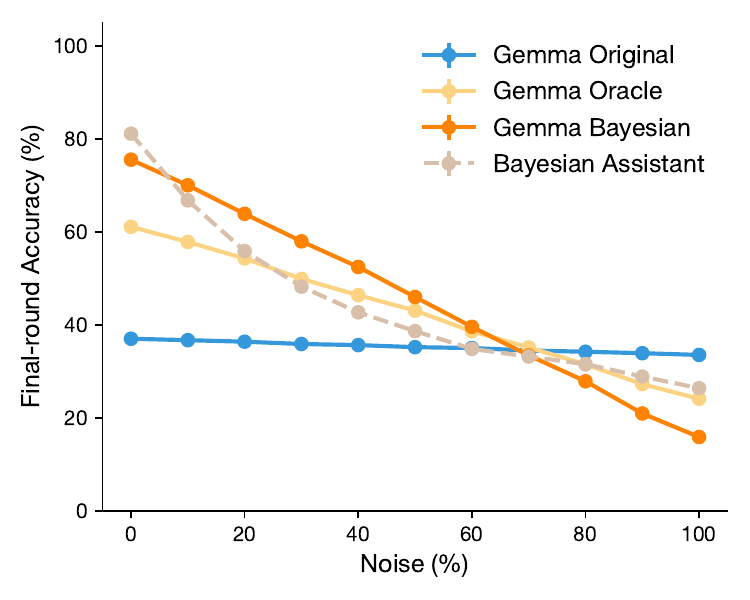}
\caption{\textbf{Final-round accuracy when interacting with a noisy simulated user.} We add noise to the simulated user's choice such that with some probability the user chooses an option that is different from the one that maximizes its reward. We plot fine-round accuracy with respect to the amount of noise. While all models show a decrease in performance as noise increases, Gemma Bayesian demonstrates greater robustness for noise levels between 10\% and 60\%. Error bars (too small to be visible in the plot) show standard errors across three random seeds (and three training runs).}
\label{fig:app_figc9_user_noise}
\end{figure*}

Our main experiments assume the simulated user always makes decisions that are consistent with its reward function. By contrast, as we show in the section \textit{Generalization to interactions with human users}, humans may behave inconsistently with their stated preferences. To simulate this real-world stochasticity, we evaluate a setting where the LLM interacts with a non-deterministic user. We add noise to the user's behavior, such that with a certain probability they select a non-optimal choice, that is a choice that does not maximize their reward. The relationship between the percentage of noise and final-round accuracy is shown in Fig.~\ref{fig:app_figc9_user_noise}. We experiment with the three variants of Gemma, and with the Bayesian Assistant. As expected, performance decreases across the board as the amount of noise increases. For realistic noise values in the 10--60\% range, we find that Gemma Bayesian is more robust to noise compared not only to Gemma Original and Gemma Oracle, but also to the Bayesian Assistant, which is the best model in the noiseless setting. This robustness to noise illustrates an advantage of an LLM fine-tuned to mimic a symbolic model compared to the original symbolic model (see \textit{Discussion}).

\subsection{What Makes Bayesian Teaching Effective?}
\label{sec:app_c4_whybayesian}

We have shown that it is more effective to fine-tune our LLMs on the Bayesian Assistant's predictions than on the user's true choices. In this section, we explore and rule out two deflationary hypotheses that might explain the effectiveness of this method, and tentatively conclude that the advantage of Bayesian teaching is in fact due the probabilistically optimal predictions made by the Bayesian Assistant. In all of the experiments described in this section, we focus on Gemma~2 9B. We use the same list of flight option sets for all models, and vary only the supervision we provide during fine-tuning (i.e. the assistant's recommendations).

\begin{figure*}[t!]
    \centering
    \includegraphics[width=\linewidth]{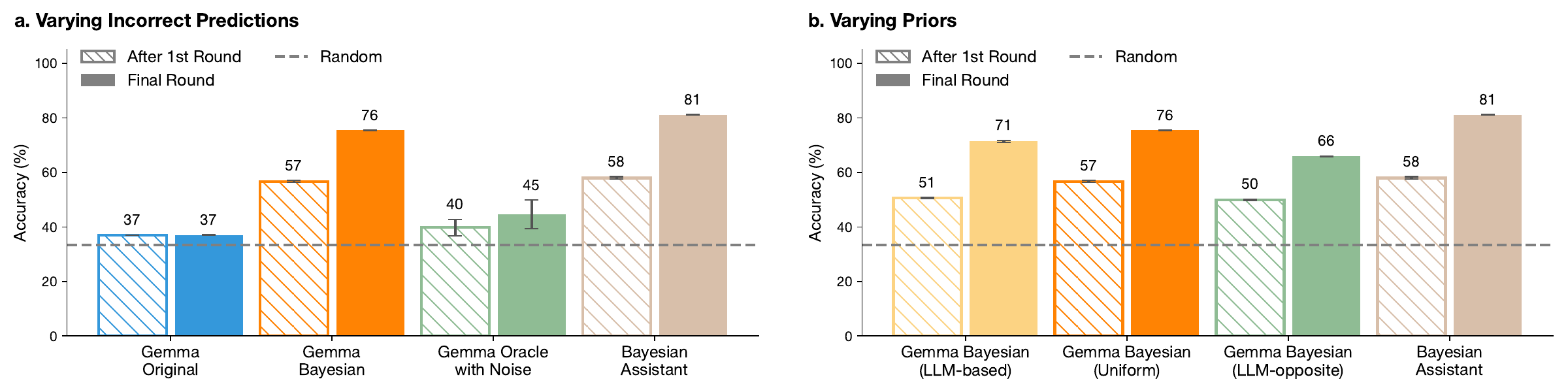}
    \begin{subfigure}[b]{0pt}
        \phantomsubcaption
        \label{fig:app_figc10a_bayesian_noise}
    \end{subfigure}
    \begin{subfigure}[b]{0pt}
        \phantomsubcaption
        \label{fig:app_figc10b_bayesian_prior}
    \end{subfigure}
    \vspace{-1em}
    \caption{\textbf{Final-round accuracy of LLMs fine-tuned with different data variants.} \textbf{(a)} Accuracy of the model using Bayesian teaching and the model using oracle teaching with random noise.  \textbf{(b)} Accuracy of models fine-tuned on predictions from variants of the Bayesian Assistant, initialized with different priors. Error bars show standard errors across three random seeds (and three training runs).}
    \label{fig:app_figc10_whybayesian}
\end{figure*}

\paragraph{Hypothesis: Incorrect predictions regularize training.} The Bayesian Assistant can make incorrect predictions, especially in the first few rounds, due to the fact that it only has limited information about the user (see the Bayesian Assistant's accuracy over rounds in Fig.~\ref{fig:app_figg18_eval_details}). Could these incorrect predictions regularize training and prevent overfitting, accounting for the effectiveness of Bayesian teaching? To test this hypothesis, we fine-tune the LLM using oracle teaching injected with random noise: 40\% of the time, instead of predicting the user's choice, the assistant recommends one of the incorrect options at random. The proportion of incorrect predictions in this control roughly matches that of the Bayesian predictions averaged across all five interactions. Contrary to the regularization hypothesis, we find that incorrect predictions do not necessarily improve performance: the model fine-tuned on noisy user's choices (Gemma Oracle with Noise) barely outperforms the original LLM and has high standard error (Fig.~\ref{fig:app_figc10a_bayesian_noise}). This suggests that random noise alone cannot explain why Bayesian predictions are more effective; rather, the Bayesian's educated mistakes are more valuable than random errors.

\paragraph{Hypothesis: The LLM benefits from the correct prior.}
We initialize the Bayesian Assistant using the uniform prior, which assigns equal probability to all reward functions, and therefore aligns with the data generation process of our evaluation. One hypothesis is that the LLM benefits from this correct prior (in the sense that it is calibrated to the distribution of simulated users in our experiment), which makes the predictions of the Bayesian Assistant more effective for supervised fine-tuning.

To test this hypothesis, we fine-tune Gemma three times, using the predictions of three variants of the Bayesian Assistant, initialized with three different priors: the uniform prior, the LLM-based prior obtained from Gemma Original (see Fig.~\ref{fig:app_figc5_prior}), and the prior that is contrary to the LLM-based one (e.g., if Gemma's prior favors cheaper flights, this prior would instead prefer more expensive flights).
The results are shown in Fig.~\ref{fig:app_figc10b_bayesian_prior}. LLMs fine-tuned on predictions from all three Bayesian models perform very well and dramatically better than the original LLM. The choice of prior does influence the performance of the fine-tuned LLMs. The model fine-tuned on Bayesian predictions using the uniform prior, which matches the distribution of users in our sample, achieves the best accuracy. The LLM-based prior, despite being biased and spiky, leads to accuracy that is only slightly worse. The LLM-opposite prior, which is both biased and mismatches the LLM's beliefs, leads to a more significant performance drop. That being said, the vast gap between all three LLMs fine-tuned on Bayesian predictions and Gemma Original suggests that the correct prior alone does not fully explain the effectiveness of Bayesian teaching.

\subsection{Qualitative Example}
\label{sec:app_c5_qualitative_example}
\begin{figure*}[t!]
\centering
\includegraphics[width=\linewidth]{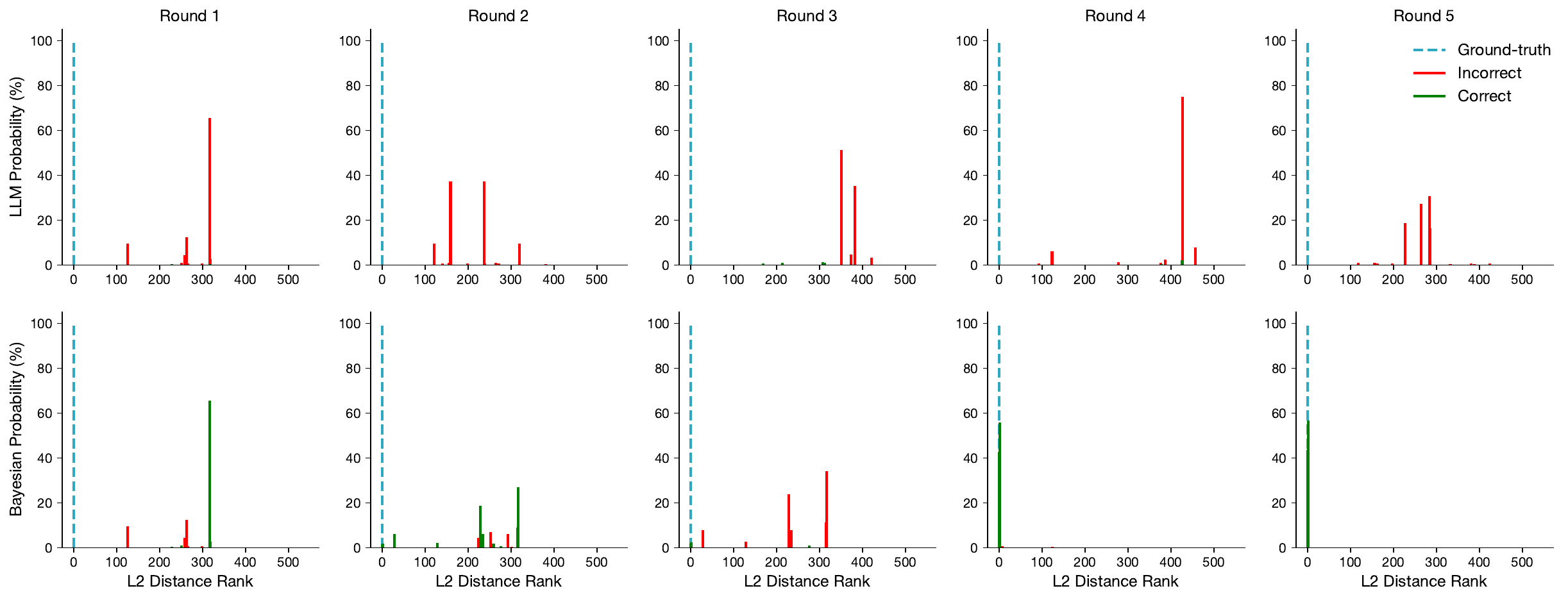}
\caption{\textbf{The reward distributions of Gemma Original (top) and the Bayesian Assistant (bottom) over multiple rounds.} The reward functions are sorted by their normalized L2 distance from the ground-truth (GT) reward function, indicated by the blue dashed line at $x=0$. Red indicates that the reward function's prediction on the given options is incorrect, while green indicates that its prediction is correct.}
\label{fig:app_figc11_reward_dist}
\end{figure*}

In Fig.~\ref{fig:app_figc11_reward_dist}, we show a qualitative example of the evolution of the reward distributions of Gemma Original and the Bayesian Assistant over interactions. In this case, since the user's true reward function differs significantly from the LLM's prior, both Gemma Original and the Bayesian Assistant perform poorly at the start of the interactions. However, while the Bayesian Assistant gradually converges toward the ground-truth reward function after a few rounds, Gemma Original continues to assign high probability to reward functions that are inconsistent with its observations.

\section{Sensitivity to the Informativeness of Option Sets}
\label{sec:app_d_informativeness}

In each round of the flight recommendation task, we present the model with a set of three flight options, and the user's choice among those options. The amount of information that can be gained through this process varies from round to round. For example, a choice between two flight options that differ in exactly one feature could be more informative than the choice between options that differ along multiple dimensions: the minimal pair of options provides direct evidence for the user's preference for the particular feature. We expect a strong probabilistic reasoner to be sensitive to this factor: when the user's choice between a particular set of options provides more information about their preferences, we expect the system to update its beliefs more substantially. 

In this section we test whether LLMs display this behavior. In contrast with the main experiments, where we sample the option sets randomly, here we sample them based on their informativeness. To measure the amount of information contained in a set of options $\mathcal{O}$, we define the \textit{ground truth information gain} as
\begin{align}
    g(\mathcal{O}, o^*, p(\bm{\theta}), q(\bm{\theta})) &=  \mathrm{KL} (p(\bm{\theta}) || q(\bm{\theta})) - \mathrm{KL} (p(\bm{\theta}) || q(\bm{\theta} | \mathcal{O}, o^*)) \\
    &= \log q(\bm{\theta}^* | \mathcal{O}, o^*) - \log q(\bm{\theta}^*),
\label{eq:acc_gain}
\end{align}

where $p(\bm{\theta}) =\delta(\bm{\theta}^*)$ and $q(\bm{\theta})$ is either $q_B(\bm{\theta})$ or $q_{LLM}(\bm{\theta})$.
This metric captures the increase in the posterior probability of the ground-truth reward function (that is, the user's true reward function) after this set of options has been observed. Note that $g$ is relative to the model that is used to update the probability distribution; we use $g_{\textit{B}}$ and $g_{\textit{LLM}}$ to refer to the gain derived from the Bayesian Assistant and the LLM, respectively.

\subsection{Experimental Setup}
\label{sec:app_d1_setup}
We randomly sample 5,000 candidate option sets, compute the ground truth information gain of each one based on the Bayesian Assistant, and select the option set that leads to the desired value of $g_{\textit{B}}$. The performance is evaluated at the end of a five-round interaction, and the ground truth information gain is averaged over these five rounds. We evaluate the Bayesian Assistant as well as Gemma Original, Gemma Oracle, and Gemma Bayesian; as in our main experiments, the Bayesian Assistant is initialized with the uniform prior.

\begin{figure*}[t!]
\centering
\includegraphics[width=0.85\textwidth]{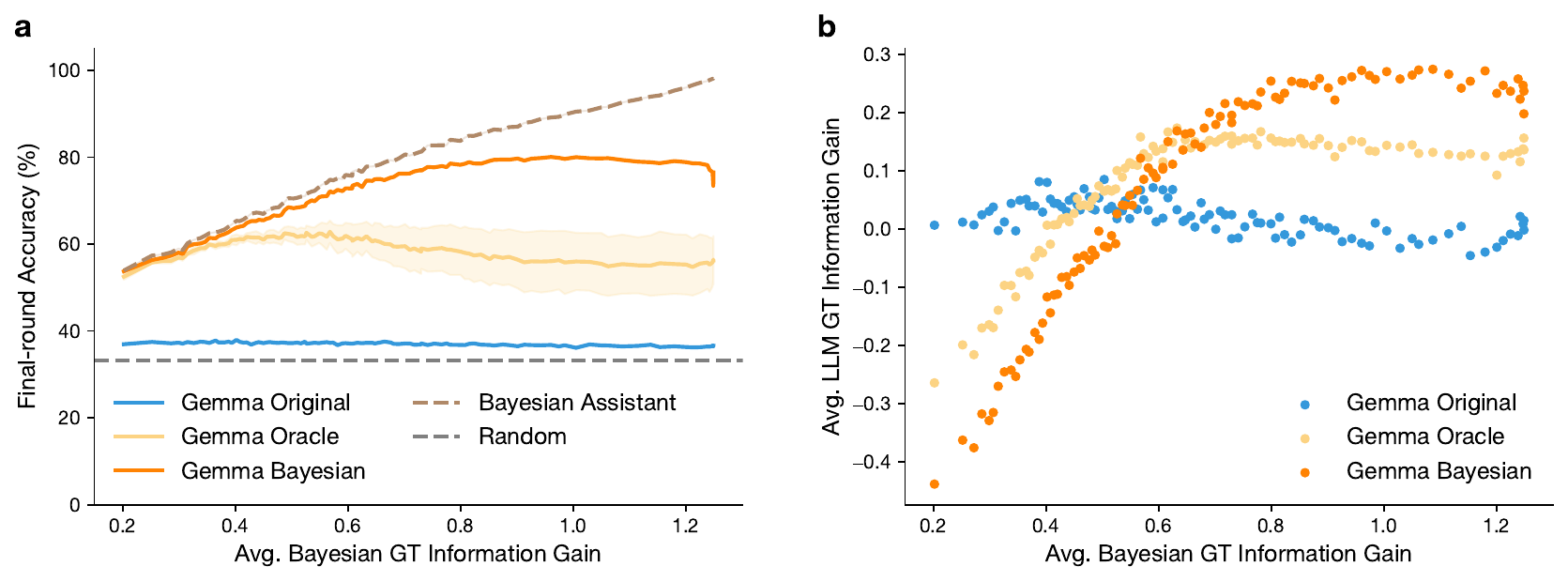}
\begin{subfigure}[b]{0pt}
    \phantomsubcaption
    \label{fig:app_figd12a_acc_vs_info_gain}
\end{subfigure}
    \begin{subfigure}[b]{0pt}
    \phantomsubcaption
    \label{fig:app_figd12b_info_gain_corr}
\end{subfigure}
\vspace{+0.5em}
\caption{\textbf{Analysis of sensitivity of LLMs to the informativeness of option sets.} \textbf{(a)} Effect of option set informativity on model performance. Option set informativity is defined by ground-truth information gain, i.e., the increase in the log probability assigned by the Bayesian Assistant to the ground truth reward function after observing the provided options. We plot accuracy after five interactions as a function of option set informativity averaged over the five interactions. \textbf{(b)} The relationship between ground-truth information gain computed from the Bayesian Assistant and from LLMs.}
\label{fig:app_figd12_info_gain}
\end{figure*}

\subsection{Results} 
\label{sec:app_d2_results}
The Bayesian Assistant's performance consistently improves as option sets become more informative: after observing highly informative options, its performance is almost perfect (Fig.~\ref{fig:app_figd12a_acc_vs_info_gain}). Gemma Original does not show sensitivity to option set informativity, but the fine-tuned models are much more sensitive to this factor: their performance positively correlates with the Bayesian ground-truth information gain up to a certain point. Gemma Bayesian saturates later than Gemma Oracle, and achieves higher final accuracy, especially in the highly informative regime.

\subsection{Comparing LLM-derived and Bayesian Information Gain} 
\label{sec:app_d3_comparison}

Recall that information gain is relative to the model that is used to update the probability distributions: $g_{\textit{LLM}}$ quantifies the amount of information the LLM can absorb from a particular set of options, whereas $g_{\textit{B}}$ quantifies the amount that the ideal Bayesian reasoner can absorb. How does $g_{\textit{LLM}}$ related to $g_{\textit{B}}$ for each of the variants of Gemma? We find that the correlation between the two measures is weak for Gemma Original (Fig.~\ref{fig:app_figd12b_info_gain_corr}). For  Gemma Oracle and Gemma Bayesian, we observe a more complex pattern. When $g_{\textit{B}}$ is small, there is a positive relationship between the two metrics, indicating that options that are informative from the Bayesian perspective are beneficial for the fine-tuned LLMs. In this part of the range, the information gain derived from Gemma Bayesian shows a stronger correlation with $g_{\textit{B}}$ compared with Gemma Oracle. When $g_{\textit{B}}$ is large, however, the relationship levels off and we no longer see a correlation between $g_{\textit{B}}$ and $g_{\textit{LLM}}$ for either of the fine-tuned models. This suggests that even Gemma Bayesian only approximates and does not fully implement the normative Bayesian reasoning strategy.

\section{Human Experiments}
\label{sec:app_e_human_exps}

\subsection{Humans As Assistants}
\label{sec:app_e1_human_as_assistant}

\paragraph{Participants.}
For the experiment where human participants acted as the assistant to a simulated user, we recruited 720 participants through the Prolific platform~\citep{palan2018prolific}. Each human participant interacted with one simulated user from a subset of 48 simulated users (out of the total 624 users), which we sampled based on the L2 distance of their reward function from the origin. The average age of human participants was 37.2 (SD=12.5). Of those, 54.9\% identified as male (395), 44.6\% as female (321), and 0.6\% preferred not to say (4). The major nationalities of human participants were the United States at 32.5\% (234), United Kingdom at 23.2\% (167), South Africa at 10.3\% (74), Canada at 7.6\% (55), and Poland at 4.4\% (32). By ethnicity, 62.5\% (450) were White, 17.4\% (125) were Black, 11.9\% (86) were Asian, and 5.6\% (40) were Mixed. All participants reported using English as their primary language. 

\paragraph{Procedure.}

At the beginning of the experiment, each participant was asked to complete a preference questionnaire to indicate their initial guess of the user's preferences for each individual feature. The participant subsequently proceeded to the annotation round, where they made recommendations from three flight options. After the selection, the human annotator received feedback indicating whether their choice was correct. They were then redirected to the preference questionnaire to report their updated beliefs about the user's preferences. This completed one round. The annotator repeated the same procedure for five rounds. Following these five rounds, we also implemented a quality control annotation round where the annotator interacted with a typical user with a highly informative option list (differing only in one feature dimension). We expected this quality control round to be very easy for participants who were paying close attention to task. We filtered out participants who failed the quality control annotation. The mean and median completion time (including the quality control annotation) was 9.35 minutes and 7.90 minutes, respectively, with a standard deviation of 5.08 minutes.

\paragraph{Additional Results.}

\begin{figure*}[t!]
    \centering
    \includegraphics[width=0.8\textwidth]{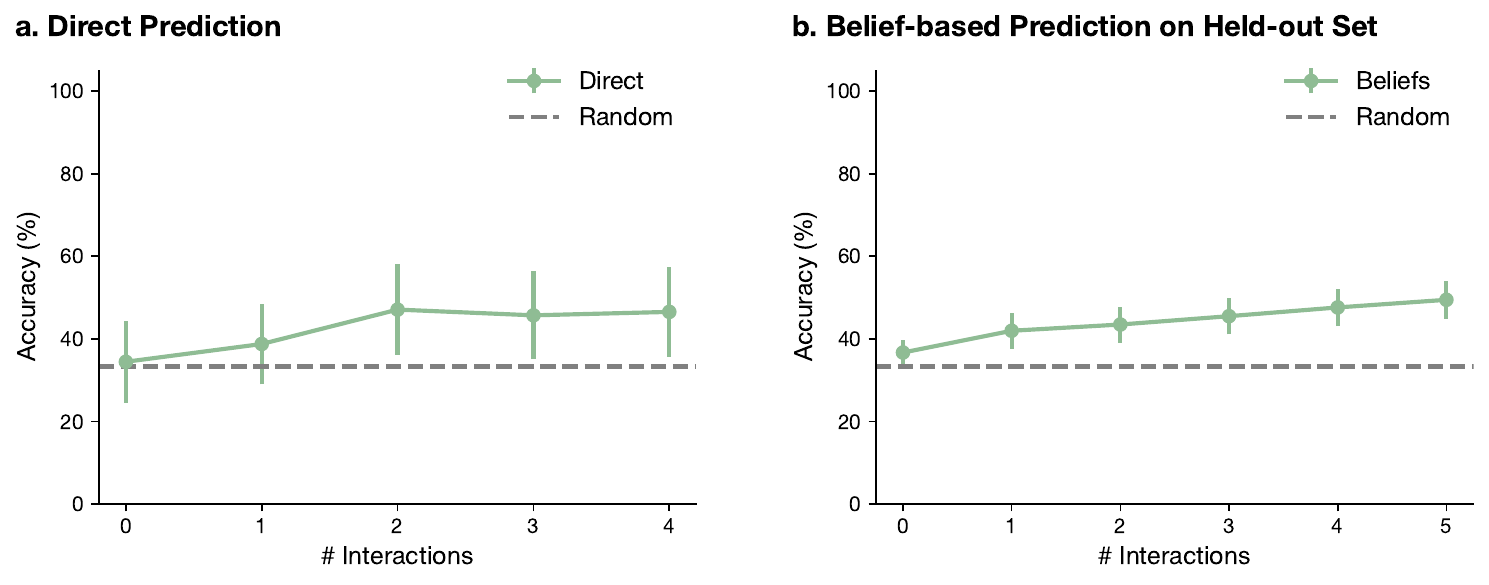}
    \caption{\textbf{Accuracy of the human assistant over rounds.} \textbf{(a)} Based on the human's direct predictions on provided option sets. \textbf{(b)} Based on the human's beliefs about the user's preferences on held-out option sets. Error bars show the averaged standard error across human participants.}
    \label{fig:app_fige13_human_assistant}
\end{figure*}

Our main results show the accuracy of human assistants using their direct predictions of the user's preferred choices. Since we also ask the annotator to rate their beliefs about the user's preferences after each round, we can also use these estimated preferences to make recommendations---the same procedure we use in Section~\ref{sec:app_a_verbalize_beliefs}. This allows us to evaluate on the larger held-out set and reduce the noise. As shown in Fig.~\ref{fig:app_fige13_human_assistant}, we find that while the accuracy of the human annotators' direct prediction may not monotonically improve from one round to the next, their beliefs about the user's preferences become consistently more accurate over rounds.

\paragraph{Qualitative Analysis.}

One pattern we observe in human assistants is that they tend to favor simpler heuristics when there is limited evidence. For example, in Table~\ref{tab:app_tabe2_human_qualitative}, we show that when there are multiple valid user preferences, human assistants may rely on simpler heuristics, e.g. in this example, always choosing the cheapest flight. In contrast, the fine-tuned Gemma Bayesian model does not seem to exhibit this behavior.

\begin{table}[ht!]
\centering
\caption{\textbf{Qualitative examples of LLM and human predictions.} Here, the user strongly prefers an early departure time, weakly prefers a short flight duration, and has no preference for the number of stops and the price. Most human participants tend to favor a simpler heuristic, i.e., always choosing the cheapest flight, while Gemma Bayesian does not seem to exhibit this behavior.}
\label{tab:app_tabe2_human_qualitative}
\footnotesize
\begin{tabularx}{\textwidth}{p{1.5cm}p{1.5cm}p{1.5cm}p{1.4cm}p{1.5cm}p{1.5cm}p{1.5cm}p{2.5cm}}
\toprule
\thead{Flight} & \thead{Departure} & \thead{Duration} & \thead{\# Stops} & \thead{Price} & \thead{User's \\ Choice} & \thead{Gemma \\ Bayesian's \\ Prediction} & \thead{Human \\ Predictions} \\
\midrule
Flight 1 & 05:12 PM & 30 min & 1 & \textcolor{blue}{\$190} & \multirow{3}{*}{Flight 1} & \multirow{3}{*}{Flight 1} & Flight 1: 66.7\% \\
Flight 2 & 03:36 PM & 12 hr 12 min  & 2 & \$460 &                           &                           & Flight 2: 26.7\% \\
Flight 3 & 10:00 PM & 10 hr 15 min  & 2 & \$640 &                           &                           & Flight 3: 6.7\%  \\
\midrule
Flight 1 & 06:48 PM & 4 hr 24 min & 1 & \$370 & \multirow{3}{*}{Flight 2} & \multirow{3}{*}{Flight 2} & Flight 1: 40.0\%  \\
Flight 2 & 07:36 AM & 16 hr 6 min & 1 & \textcolor{blue}{\$100} &                           &                           & Flight 2: 33.3\% \\
Flight 3 & 10:00 PM & 20 hr & 0 & \$550 &                           &                           & Flight 3: 26.7\%  \\
\midrule
Flight 1 & 10:00 PM & 30 min & 1 & \textcolor{blue}{\$280} & \multirow{3}{*}{Flight 3} & \multirow{3}{*}{Flight 3}  & Flight 1: 60\%  \\
Flight 2 & 08:24 PM & 30 min & 0 & \$910 &                          &                            & Flight 2: 0.0\% \\
Flight 3 & 06:00 AM & 8 hr 18 min & 0 & \$370 &                           &                            & Flight 3: 40\%  \\
\bottomrule
\end{tabularx}
\end{table}

\subsection{Humans As Users}
\label{sec:app_e2_human_as_user}

\paragraph{Participants.}

For the experiment where human participants acted as the users, we recruited 500 participants through the Prolific platform. The average age of the participants was 38.7 (SD=13.6); 51.0\% identified as male (255), 48.4\% as female (242), and 0.6\% preferred not to say (3). The major nationalities of human participants were the United States at 40.0\% (200), United Kingdom at 16.0\% (80), South Africa at 9.0\% (45), Canada at 7.8\% (39), and Australia at 5.6\% (28), with smaller representations from other countries. In terms of ethnicity, 65.2\% (326) identified as White, 15.0\% (75) as Black, 8.4\% (42) as Asian, 7.2\% (36) as Mixed, and 4.0\% (20) as Other. All participants reported that English is their primary language.

\paragraph{Procedure.}
Each participant in this experiment was first asked to complete a preference questionnaire to indicate their own preferences for different flight features. They then proceeded to the annotation rounds, where they needed to select their preferred option out of three flight options. To ensure quality, we required annotators to think for at least 30 seconds before making their selection. The procedure continued for five rounds. Participants were told to make choices consistent with their initially stated preferences throughout all five rounds. The mean and median completion times were 6.43 minutes and 5.18 minutes, respectively, with a standard deviation of 3.51 minutes.

\paragraph{Additional Results.}

\begin{figure*}[t!]
    \centering
    \includegraphics[width=\textwidth]{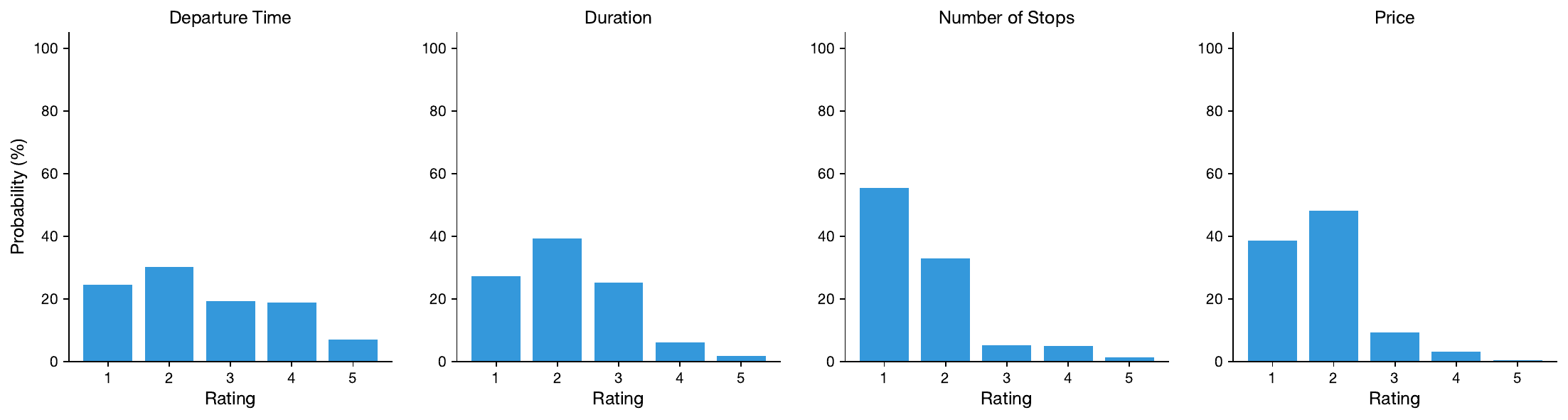}
    \caption{\textbf{The distributions of human participants' initial preferences for different flight features.} A rating of 1 indicates the strongest preference for the earliest departure time, the shortest duration, the fewest number of stops, and the lowest price, while a rating of 5 indicates the opposite. A rating of 3 indicates no preference.}
    \label{fig:app_fige14_human_preference}
\end{figure*}

\begin{figure*}[t!]
    \centering
    \includegraphics[width=0.8\textwidth]{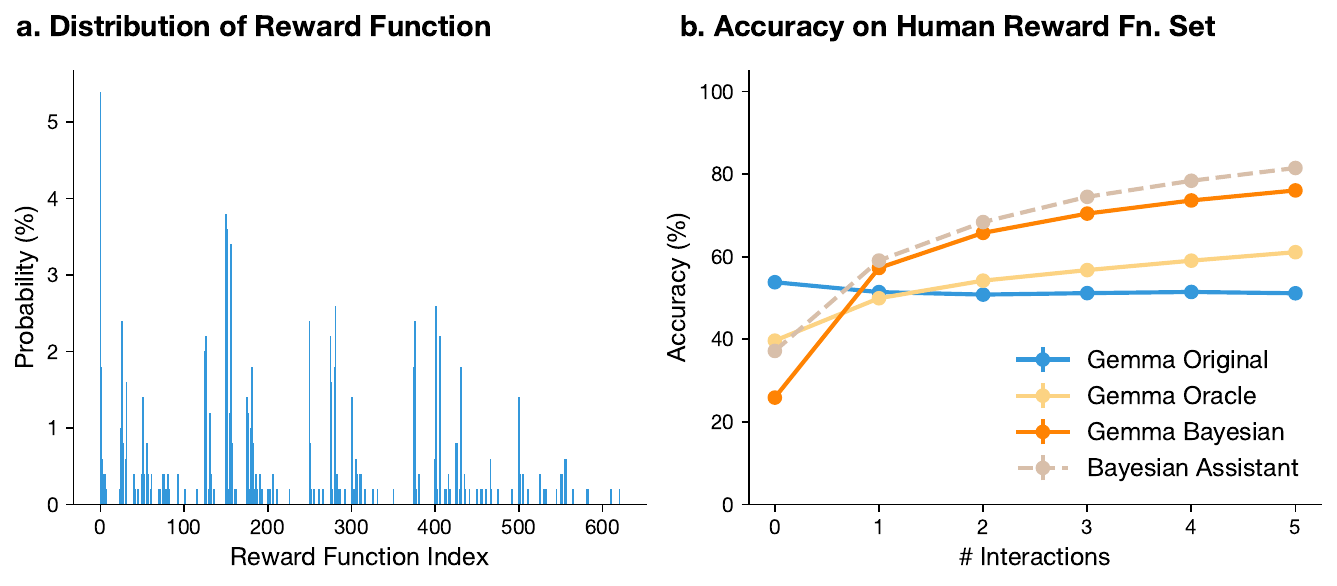}
    \begin{subfigure}[b]{0pt}
        \phantomsubcaption
        \label{fig:app_fige15a_human_reward_fn}
    \end{subfigure}
    \begin{subfigure}[b]{0pt}
        \phantomsubcaption
        \label{fig:app_fige15b_human_set_acc}
    \end{subfigure}
    \vspace{+0.5em}
    \caption{\textbf{Analysis of human reward functions.} \textbf{(a)} Distribution of human reward functions. \textbf{(b)} Accuracy over rounds on the subset of the original data where the simulated user's reward function is set of reward functions stated by the human participants. Error bars show standard errors across three random seeds (and three training runs).}
    \label{fig:app_fige15_human_set_analysis}
\end{figure*}

\begin{figure*}[t!]
    \centering
    \includegraphics[width=0.76\textwidth]{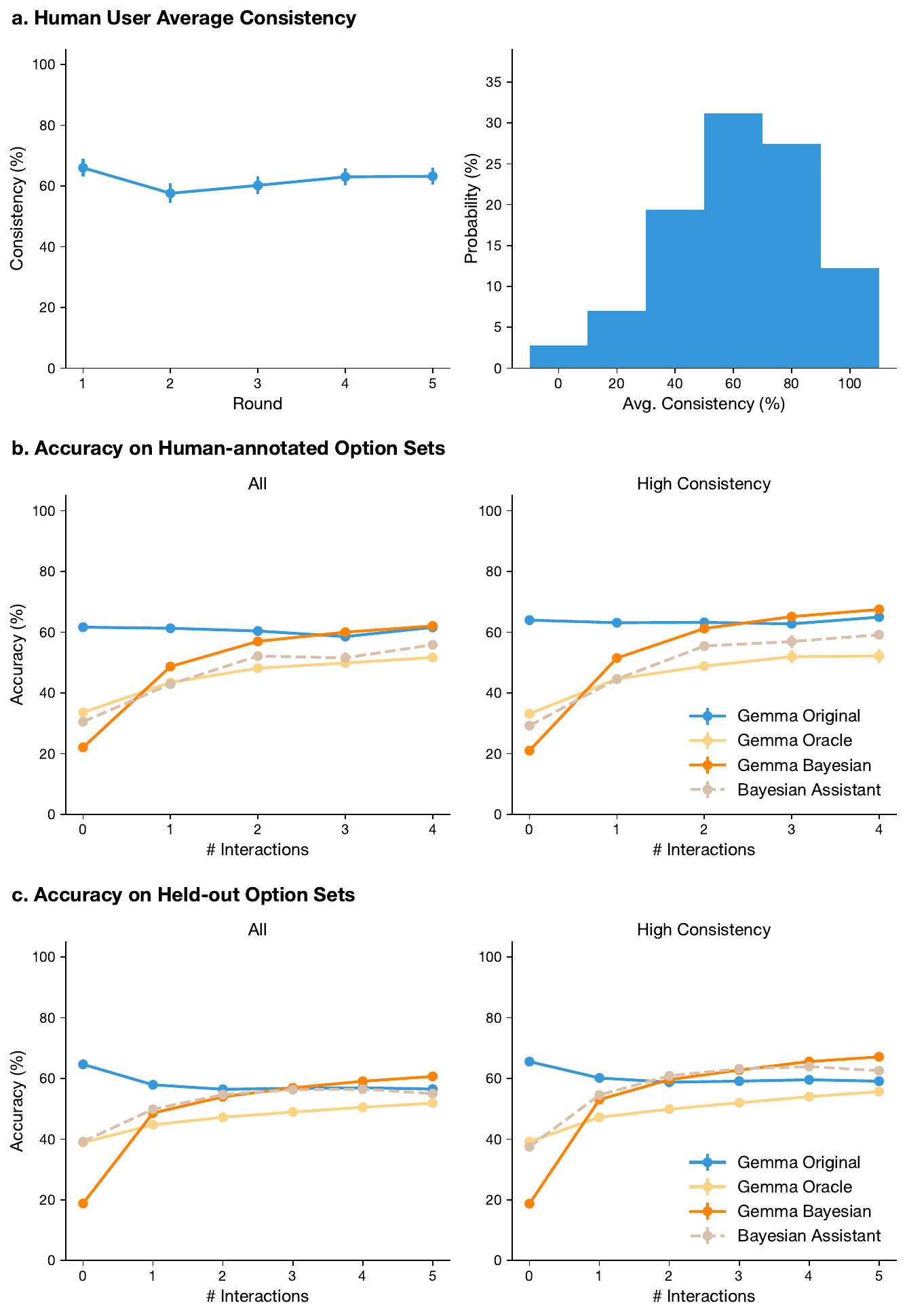}
    \begin{subfigure}[b]{0pt}
        \phantomsubcaption
        \label{fig:app_fige16a_human_user_consistency}
    \end{subfigure}
    \begin{subfigure}[b]{0pt}
        \phantomsubcaption
        \label{fig:app_fige16b_human_user_acc}
    \end{subfigure}
    \begin{subfigure}[b]{0pt}
        \phantomsubcaption
        \label{fig:app_fige16c_human_user_heldout_acc}
    \end{subfigure}
    \vspace{+0.5em}
    \caption{\textbf{Results on interactions with real human users.} \textbf{(a)} Consistency between the human users'
    choices and the predictions derived from their initially stated preferences. We show user consistency over rounds and the distribution of the average user consistency. Error bars show standard errors across five-round option sets list. \textbf{(b)} Accuracy over rounds on human-annotated option sets. We show the results of all human users and users with high consistency, i.e., their choices matched their initially stated preferences in 4 or 5 of the rounds (40.4\% of the data). \textbf{(c)} Accuracy over rounds on the held-out set, where the preferred choices are deterministically computed based on the human user's preferences. Error bars show standard errors across three random seeds (and three training runs).}
    \label{fig:app_fige16c_human_user_analysis}
\end{figure*}

In the main paper we report results for this more realistic setting where the model interacts with real human users on the flight recommendation task. Surprisingly, we also find that the original LLMs achieved good performance, unlike what was observed in earlier experiments.

We hypothesize that two factors may contribute to its improved performance. First, unlike our simulated users whose preferences are uniformly sampled from the space of possible reward functions, human preferences are biased towards particular types of functions, and in Fig.~\ref{fig:app_fige15a_human_reward_fn} we show that some reward functions are considerably more common than others in our sample of human participants. For example, most participants report preferring cheaper flights (see Fig.~\ref{fig:app_fige14_human_preference}). As such, a viable strategy for the original LLM could be to rely on its prior knowledge about user preferences to make relatively good recommendations.
To investigate this further, we filter results for simulated users based on reward functions stated by the human participants in Fig.~\ref{fig:app_fige15b_human_set_acc}.
We observe that also in this case, Gemma Original achieves a higher accuracy of around 60\% (as opposed to 37\% in Fig. 2, matching the high accuracy it obtained in Fig. 6. This makes it clear that the bias among the human preferences in this experiment contributes to the stronger performance of original LLMs.

Secondly, human users may not necessarily behave consistently with their preferences, i.e., their choices may differ from those that would reflect their initially stated preferences.
Indeed, note how in Fig.~\ref{fig:app_fige15b_human_set_acc} the gap between the original LLM and the Bayesian LLM increases significantly when evaluating on consistent simulated users.
To quantify this potential discrepancy, we compute the consistency between the human user's choices and the predictions derived from their preferences.
The latter are obtained by mapping their stated preferences to corresponding reward functions and selecting the option with the highest reward accordingly.
In line with our hypothesis, the average consistency is relatively low at 60\%, with chance performance being 33.3\% (Fig.~\ref{fig:app_fige16a_human_user_consistency}).

We further break down the performance by user consistency over rounds and show results for high-consistency users; that is, users whose choices were consistent with their stated preferences in 4 or 5 of the rounds (Fig.~\ref{fig:app_fige16b_human_user_acc}). We find that all models perform better for the high-consistency users. Specifically, when user consistency is high, the improvement of Gemma Bayesian over Gemma Original increases.

Finally, to limit the effect of such inconsistencies, while still retaining the real interactions between human users and the model, we also evaluate the LLMs on a held-out set of 100 randomly sampled options that simulate perfectly consistent users; to do so we use the preferred options derived from the participants' initially stated preferences rather than the participants' actual choices.
As shown in Fig.~\ref{fig:app_fige16c_human_user_heldout_acc}, when removing inconsistency from the evaluation data, Gemma Bayesian achieves the best performance. Gemma Original performs best initially, likely due to its correct prior about human users, but its performance decreases over rounds, indicating its limited ability to incorporate the simulated user's feedback.

\subsection{Human Annotation Interface}
\label{sec:app_e3_human_interface}
\begin{figure*}[t!]
    \centering
    \includegraphics[width=0.47\linewidth]{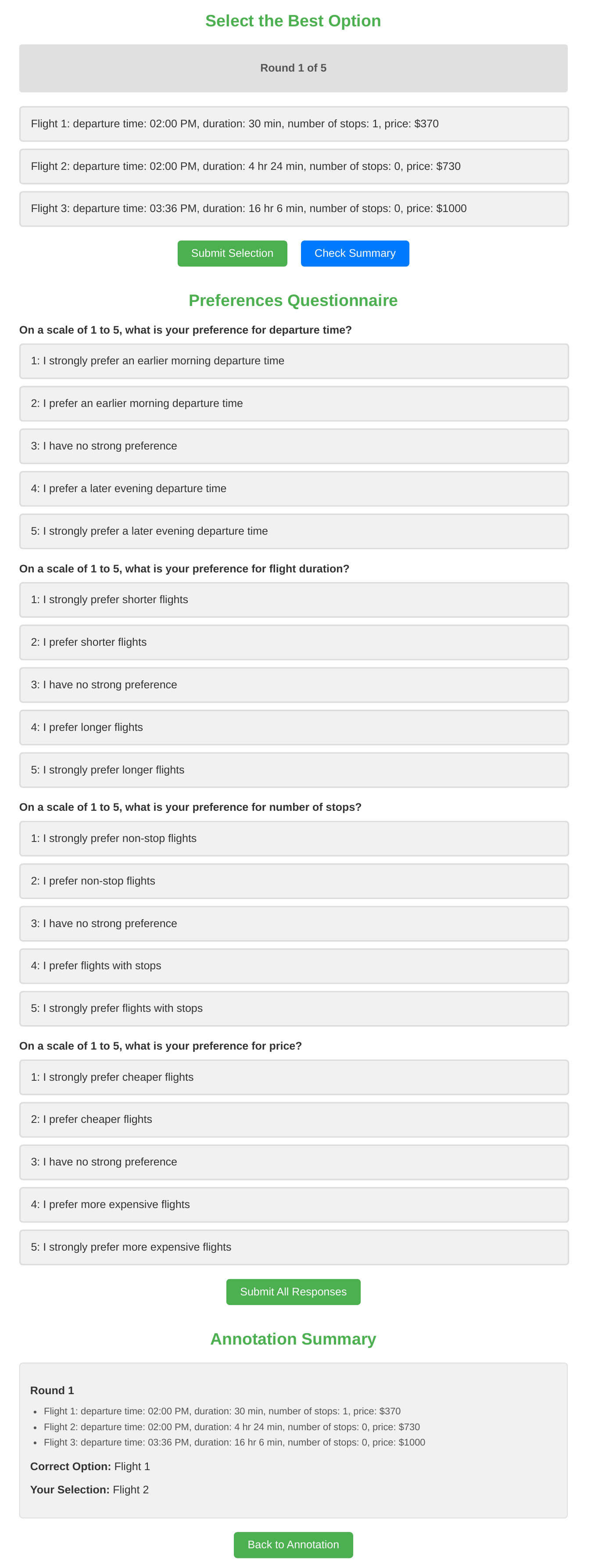}
    \vspace{+0.5em}
    \caption{\textbf{Example of annotation interface where humans act as the flight recommendation assistant.} The human annotator was asked to select the best option and rate their estimation of the user's preferences. We also allow the annotator to check a summary of previous flight booking history. The annotation interface where humans act as the user is similar.}
    \label{fig:app_fige17_interface}
\end{figure*}

We show the human annotation interface where humans act as the assistant in Fig.~\ref{fig:app_fige17_interface}. The interface allows the human annotator to select the best option from three flight options, rate their estimation of the user's preferences, and check the flight booking history from previous rounds. The annotation interface where humans act as the user is similar.

\section{Statistical Analyses}
\label{sec:app_f_stats}

This Supplementary reports analyses that test whether Bayesian teaching leads to statistically significant improvement over the baselines.
We fit linear mixed-effects models treating each method (Bayesian teaching, oracle teaching, and the original) and model family (Gemma, Llama, and Qwen) as fixed effects while controlling for various sources of randomness. For flight and hotel recommendation, we include training run, evaluation random seed, and reward function as random effects. For web shopping, we treat training run ad product category as random effects. Overall, the models demonstrate statistically significant differences between methods across all domains and all three model families (Gemma, Llama, and Qwen).

In flight recommendation, the original LLM achieves a baseline accuracy of 37.0\% (95\% CI: 30.6–43.5\%). The Oracle LLM performs significantly better with a 24.0\% increase (95\% CI: 16.6–31.4\%, p $<$ 0.001), while the Bayesian LLM shows an even more substantial 38.5\% increase (95\% CI: 31.1–45.9\%, p $<$ 0.001). Model family shows no significant effect on performance, with differences between model families all non-significant. The interaction between method and model family was not statistically significant (minimum p = 0.19). Within each model family, improvements between all methods are significant (p $<$ 0.001), with the exception of Qwen Oracle versus Qwen Original which shows slightly weaker but still significant improvement (p = 0.002).

In hotel recommendation, the original LLM achieves a baseline accuracy of 36.7\% (95\% CI: 32.1–41.3\%). The Oracle LLM performs significantly better with a 16.7\% increase (95\% CI: 11.4–22.0\%, p $<$ 0.001), while the Bayesian LLM shows a 29.4\% increase (95\% CI: 24.1–34.7\%, p $<$ 0.001). Model family shows no significant main effect on performance. The interaction between method and model family is not statistically significant (all interaction p-values $>$ 0.11). Within each model family, most pairwise comparisons show p-values $<$ 0.001, with two exceptions: Llama Bayesian versus Llama Oracle shows weak significance (p = 0.001), and Qwen Oracle versus Qwen Original shows weaker significance (p = 0.002).

In web shopping, the original LLM achieves a baseline accuracy of 54.0\% (95\% CI: 49.6–58.4\%). The Oracle LLM performs significantly better with a 7.1\% increase (95\% CI: 2.3–11.8\%, p = 0.013), while the Bayesian LLM shows a more substantial 18.6\% increase (95\% CI: 13.8–23.4\%, p $<$ 0.001). Unlike the other domains, model family shows a significant effect, with Qwen showing a significant decrease of -11.1\% (95\% CI: -17.0–5.3\%, p = 0.003) compared to the baseline. There is also a significant interaction between Oracle method and Qwen (15.8\%, 95\% CI: 9.0–22.6\%, p = 0.001). Within-family pairwise comparisons show different patterns: for Gemma, all method comparisons are significant (Original-Oracle: p = 0.033; others p $<$ 0.001); for Llama, Original-Oracle is non-significant (p = 0.199) while Original-Bayesian (p = 0.001) and Oracle-Bayesian (p = 0.004) are significant; for Qwen, Original-Oracle and Original-Bayesian are highly significant (p $<$ 0.001), but Oracle-Bayesian is non-significant (p = 0.282).

\section{Results Details}
\label{sec:app_g_details}

\begin{figure*}[t!]
    \centering
    \includegraphics[width=\linewidth]{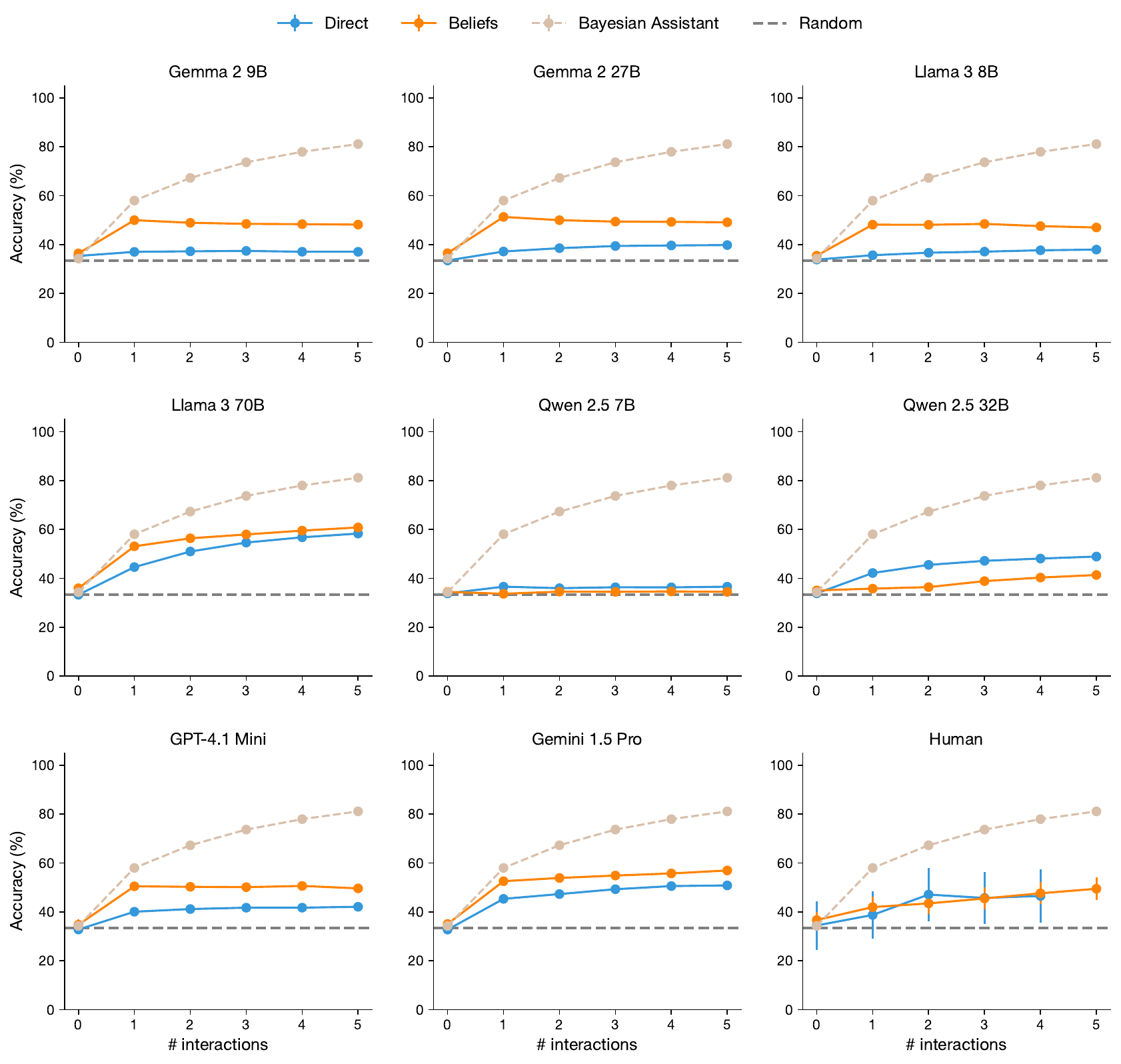}
    \caption{\textbf{Accuracy over rounds for different original LLMs.} We show accuracy based on direct predictions and accuracy based on predictions derived from their beliefs about users' preferences. Error bars show standard errors across three random seeds (and three training runs).}
    \label{fig:app_figg18_eval_details}
\end{figure*}

\begin{figure*}[t!]
    \centering
    \includegraphics[width=\linewidth]{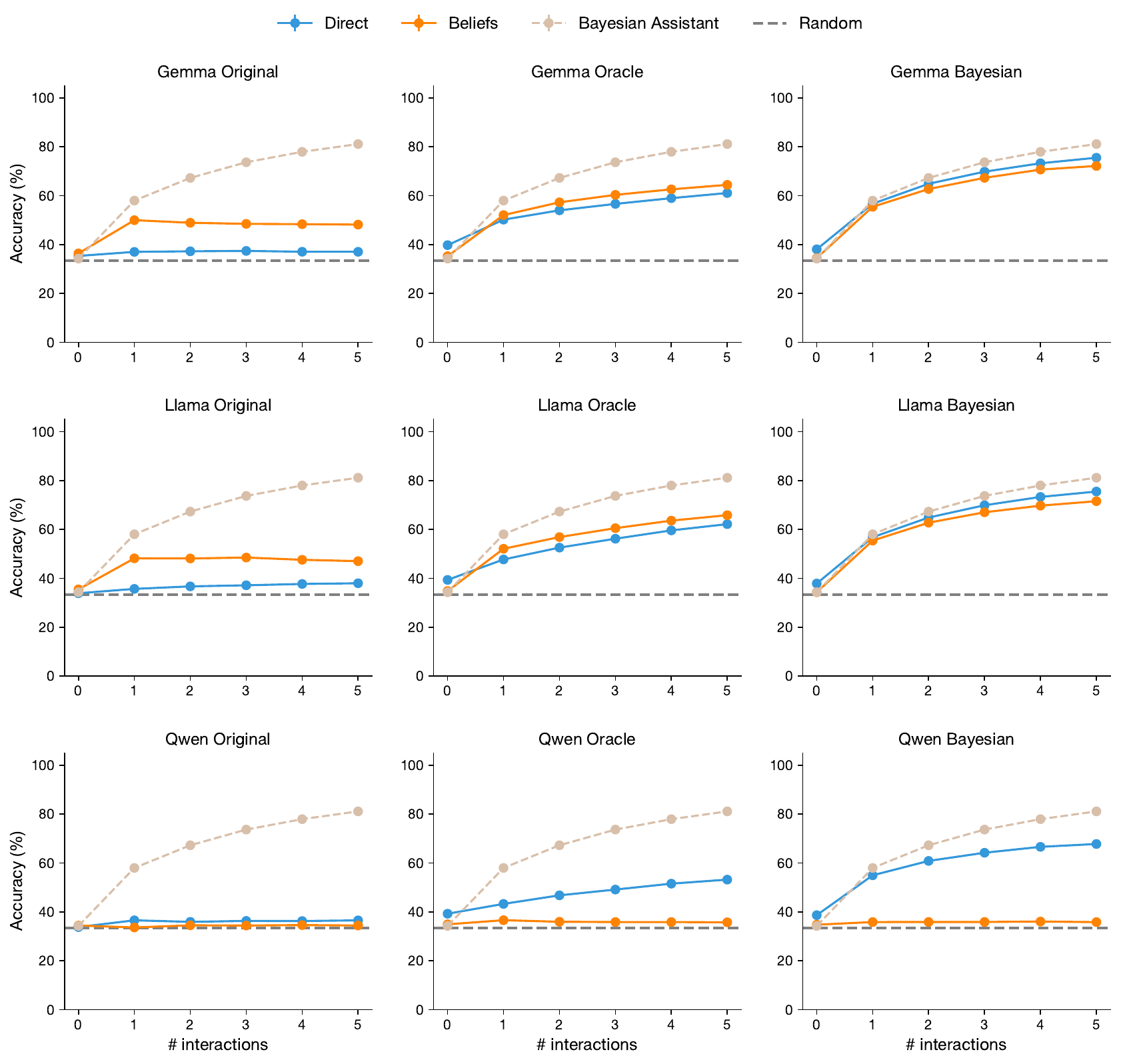}
    \caption{\textbf{Accuracy over rounds for different original LLMs and fine-tuned LLMs on the flight recommendation task.} We show accuracy based on direct predictions and accuracy based on predictions derived from their beliefs about users' preferences. Error bars show standard errors across three random seeds (and three training runs).}
    \label{fig:app_figg19_ft_flight_details}
\end{figure*}

\begin{figure*}[t!]
    \centering
    \includegraphics[width=\linewidth]{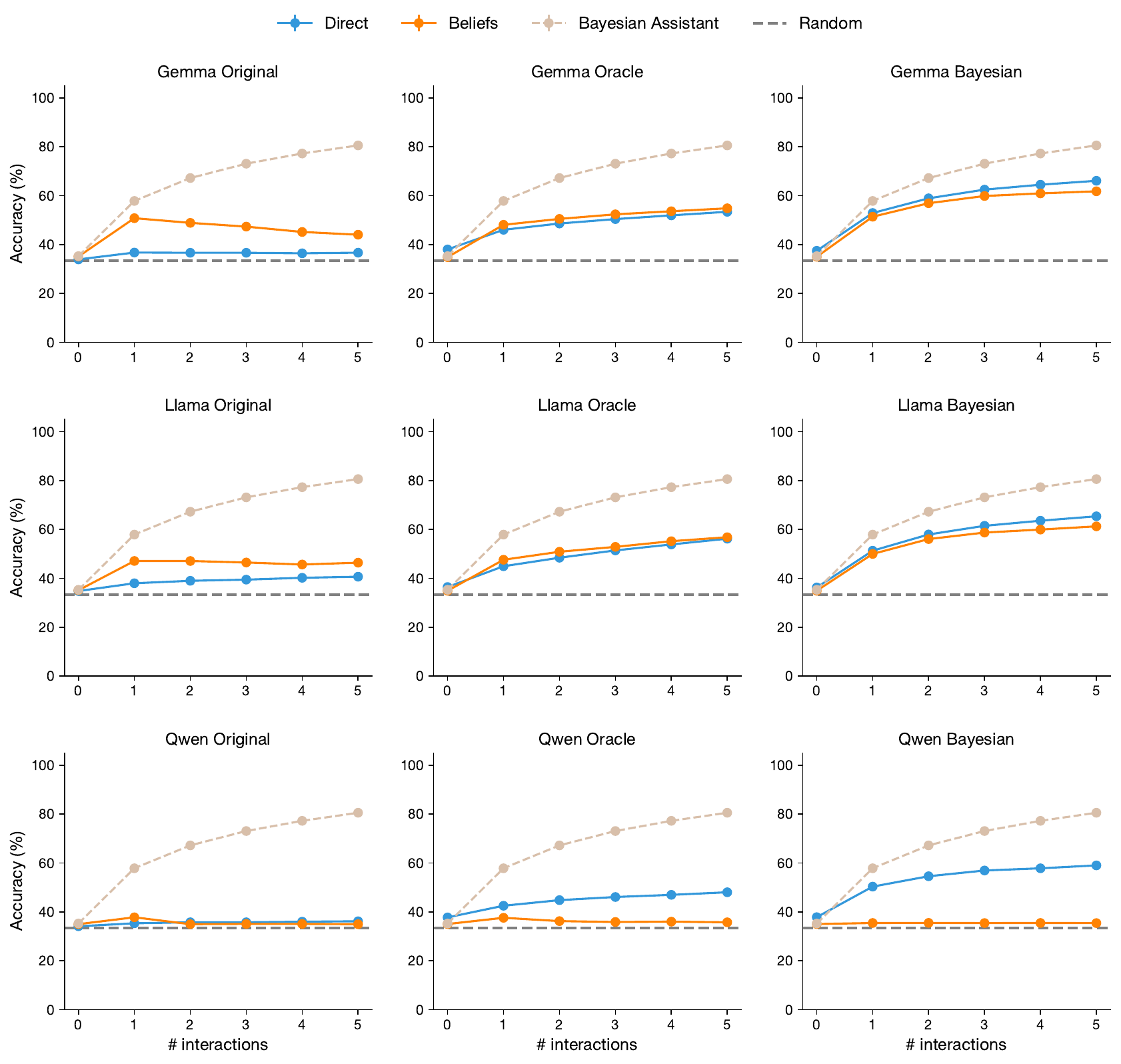}
    \caption{\textbf{Accuracy over rounds for different original LLMs and fine-tuned LLMs on the hotel recommendation task.} We show accuracy based on direct predictions and accuracy based on predictions derived from their beliefs about users' preferences. Error bars show standard errors across three random seeds (and three training runs).}
    \label{fig:app_figg20_ft_hotel_details}
\end{figure*}

\begin{figure*}[t!]
    \centering
    \includegraphics[width=\linewidth]{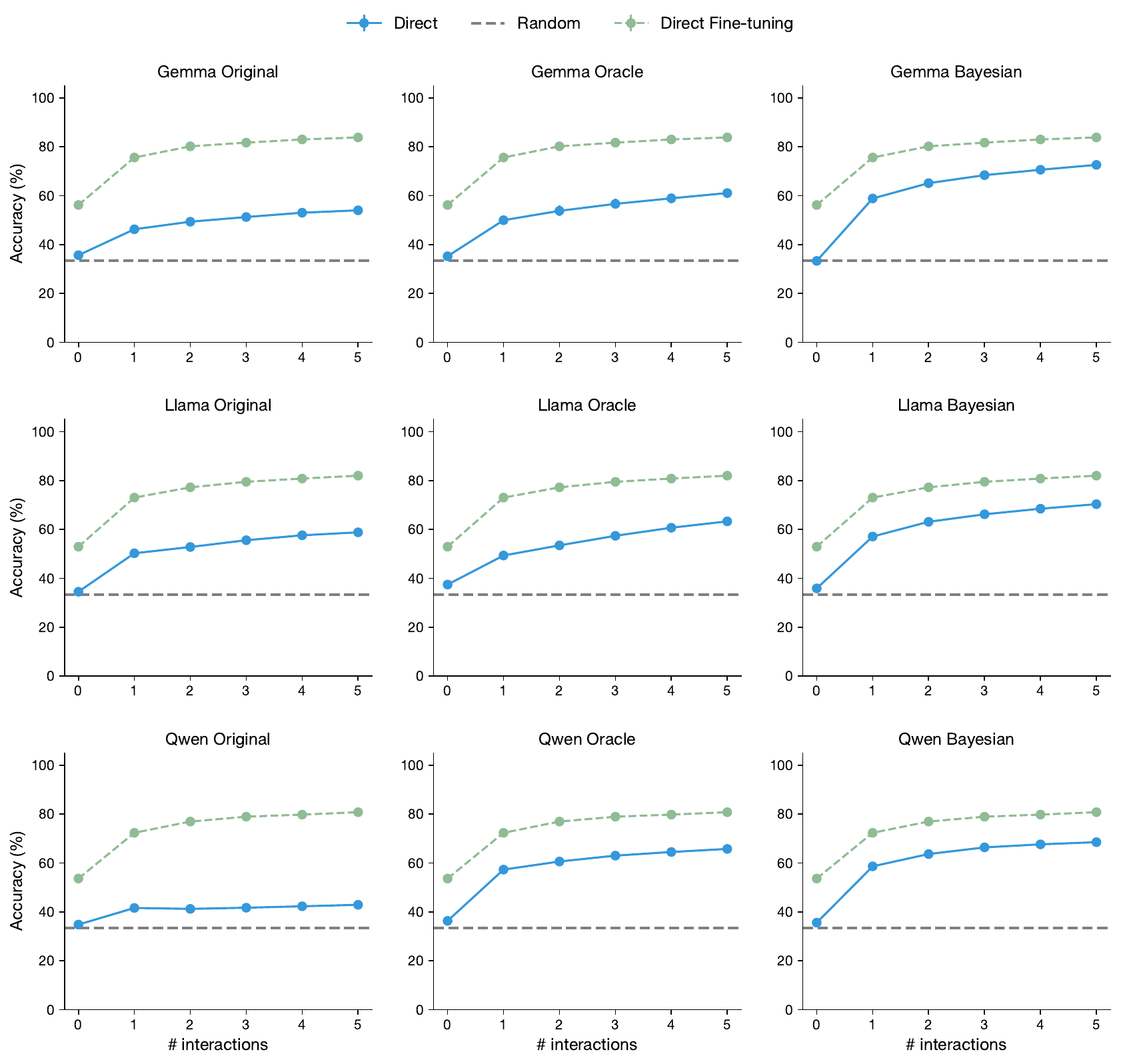}
    \caption{\textbf{Accuracy over rounds for different original LLMs and fine-tuned LLMs on the web shopping task.} We show accuracy based on direct predictions and accuracy based on predictions derived from their beliefs about users' preferences.  Error bars show standard errors across three training runs if any.}
    \label{fig:app_figg21_ft_webshop_details}
\end{figure*}

We show results over rounds for different models and methods in Fig.~\ref{fig:app_figg18_eval_details}–\ref{fig:app_figg21_ft_webshop_details}. For each, we show the accuracy based on the LLM's or human direct prediction (``direct'') and accuracy based on predictions derived from their beliefs about the user's preferences (``beliefs'') if available.

\section{Example Interactions}
\label{sec:app_h_prompts}

We show example interactions in Tables~\ref{tab:prompt_interaction}–\ref{tab:prompt_webshop_verbose}.
\clearpage
\begin{table*}[t!]
\centering
\caption{\textbf{Example of a two-round interaction between the user and the LLM.}}
\begin{tabular}{p{\textwidth}}
\toprule
\begin{minipage}[t]{\linewidth}
\texttt{\footnotesize \textbf{User:} Help me select the best flights for my trips. I have specific preferences for what I like and dislike in a flight, and  these preferences remain the same. You need to figure out my preferences and select the best flights for me. Use your best judgment if you are unsure. Do not say you need more information. \\
\\
Which flight is the best option? \\
\\
Flight 1: \\
departure time: 02:00 PM, duration: 2 hr 30 min, number of stops: 1, price: \$370 \\
Flight 2: \\
departure time: 10:00 PM, duration: 4 hr 24 min, number of stops: 0, price: \$730 \\
Flight 3: \\
departure time: 03:36 PM, duration: 16 hr 6 min, number of stops: 0, price: \$900 \\
\\
\textbf{Model:} The best option is Flight 1.\\
\\
\textbf{User:} Your option Flight 1 is incorrect. I prefer Flight 2.\\
\\
Which flight is the best option? \\
\\
Flight 1: \\
departure time: 04:00 PM, duration: 18 hr 3 min, number of stops: 2, price: \$280 \\
Flight 2: \\ 
departure time: 10:48 AM, duration: 6 hr 21 min, number of stops: 1, price: \$370 \\
Flight 3: \\
departure time: 06:48 PM, duration: 10 hr 5 min, number of stops: 1, price: \$810 \\
\\
\textbf{Model:} The best option is Flight 2.\\
\\
\textbf{User:} Your option Flight 2 is correct.} \\
\end{minipage} \\
\bottomrule
\end{tabular}
\label{tab:prompt_interaction}
\end{table*}

\begin{table*}[t!]
\centering
\caption{\textbf{Example of prompting the LLM for the user's preference on a single feature.}}
\begin{tabular}{l}
\toprule
\begin{minipage}[t]{\linewidth}
\texttt{\footnotesize\textbf{User:} Help me select the best flights for my trips. I have specific preferences for what I like and dislike in a flight, and  these preferences remain the same. You need to figure out my preferences and select the best flights for me. Use your best judgment if you are unsure. Do not say you need more information. \\
\\
Which flight is the best option? \\
\\
Flight 1: \\
departure time: 02:00 PM, duration: 2 hr 30 min, number of stops: 1, price: \$370 \\
Flight 2: \\
departure time: 10:00 PM, duration: 4 hr 24 min, number of stops: 0, price: \$730 \\
Flight 3: \\
departure time: 03:36 PM, duration: 16 hr 6 min, number of stops: 0, price: \$900 \\
\\
\textbf{Model:} The best option is Flight 1.\\
\\
\textbf{User:} Your option Flight 1 is incorrect. I prefer Flight 2.\\
\\
On a scale of 1 to 5, what is my preference for price? \\
\\
- 1: I strongly prefer a cheaper flight \\
- 2: I prefer a cheaper flight \\
- 3: I have no strong preference \\
- 4: I prefer a more expensive flight \\
- 5: I strongly prefer a more expensive flight  \\
\\
\textbf{Model:} Your preference for price is: 1. \\}
\end{minipage} \\
\bottomrule
\end{tabular}
\label{tab:prompt_reward}
\end{table*}

\begin{table*}[t!]
\centering
\caption{\textbf{Example of a two-round interaction between the user and the LLM using the numerical representation of flights.}}
\begin{tabular}{l}
\toprule
\begin{minipage}[t]{\linewidth}
\texttt{\footnotesize\textbf{User:} Help me select the best flights for my trips. I have specific preferences for what I like and dislike in a flight, and  these preferences remain the same. You need to figure out my preferences and select the best flights for me. Use your best judgment if you are unsure. Do not say you need more information. \\
\\
Which flight is the best option? \\
\\
Flight 1:\\
departure time: 0.7, duration: 0.5, number of stops: 1.0, price: 0.9\\
Flight 2:\\
departure time: 0.9, duration: 0.6, number of stops: 0.0, price: 0.7\\
Flight 3:\\
departure time: 0.5, duration: 0.9, number of stops: 0.5, price: 0.1\\
\\
\textbf{Model:} The best option is Flight 1.\\
\\
\textbf{User:} Your option Flight 1 is incorrect. I prefer Flight 2.\\
\\
Which flight is the best option? \\
\\
Flight 1: \\
departure time: 0.5, duration: 0.9, number of stops: 0.5, price: 0.1 \\
Flight 2: \\ 
departure time: 0.2, duration: 0.3, number of stops: 0.5, price: 0.2 \\
Flight 3: \\
departure time: 0.8, duration: 0.8, number of stops: 1.0, price: 0.9 \\
\\
\textbf{Model:} The best option is Flight 2.\\
\\
\textbf{User:} Your option Flight 2 is correct.} \\
\end{minipage} \\
\bottomrule
\end{tabular}
\label{tab:prompt_interaction_num}
\end{table*}

\begin{table*}[t!]
\centering
\caption{\textbf{Example of prompting the LLM for the user's preference on a single feature using the numerical representation of flights.}}
\begin{tabular}{l}
\toprule
\begin{minipage}[t]{\linewidth}
\texttt{\footnotesize\textbf{User:} Help me select the best flights for my trips. I have specific preferences for what I like and dislike in a flight, and  these preferences remain the same. You need to figure out my preferences and select the best flights for me. Use your best judgment if you are unsure. Do not say you need more information. \\
\\
Which flight is the best option? \\
\\
Flight 1:\\
departure time: 0.7, duration: 0.5, number of stops: 1.0, price: 0.9\\
Flight 2:\\
departure time: 0.9, duration: 0.6, number of stops: 0.0, price: 0.7\\
Flight 3:\\
departure time: 0.5, duration: 0.9, number of stops: 0.5, price: 0.1\\
\\
\textbf{Model:} The best option is Flight 1. \\
\\
\textbf{User:} Your option Flight 1 is incorrect. I prefer Flight 2.\\
\\
On a scale of 1 to 5, what is my preference for duration? \\
\\
- 1: I strongly prefer the minimum value \\
- 2: I prefer the minimum value \\
- 3: I have no strong preference \\
- 4: I prefer the maximum value \\
- 5: I strongly prefer the maximum value \\}
\end{minipage} \\
\bottomrule
\end{tabular}
\label{tab:prompt_reward_num}
\end{table*}

\begin{table*}[t!]
\centering
\caption{\textbf{Example of prompting the LLM for the user's preference on a single feature using the model's directly generated probability judgements.}}
\begin{tabular}{l}
\toprule
\begin{minipage}[t]{\linewidth}
\texttt{\footnotesize\textbf{User:} Help me select the best flights for my trips. I have specific preferences for what I like and dislike in a flight, and  these preferences remain the same. You need to figure out my preferences and select the best flights for me. Use your best judgment if you are unsure. Do not say you need more information. \\
\\
Which flight is the best option? \\
\\
Flight 1: \\
departure time: 02:00 PM, duration: 2 hr 30 min, number of stops: 1, price: \$370 \\
Flight 2: \\
departure time: 10:00 PM, duration: 4 hr 24 min, number of stops: 0, price: \$730 \\
Flight 3: \\
departure time: 03:36 PM, duration: 16 hr 6 min, number of stops: 0, price: \$900 \\
\\
\textbf{Model:} The best option is Flight 1.\\
\\
\textbf{User:} Your option Flight 1 is incorrect. I prefer Flight 2.\\
\\
On a scale of 1 to 5, what is my preference for price? \\
\\
- 1: I strongly prefer a cheaper flight \\
- 2: I prefer a cheaper flight \\
- 3: I have no strong preference \\
- 4: I prefer a more expensive flight \\
- 5: I strongly prefer a more expensive flight  \\
\\
Provide an integer between 0 and 100 (\%) that reflects the probability of each scale. Format your response exactly as follows:\\
\\
- 1: ??\%
...\\
\\
\textbf{Model:} The probabilities of each scale are: \\
\\
- 1: 70\% \\
- 2: 10\% \\
- 3: 15\% \\
- 4: 5\% \\
- 5: 0\% \\} 
\end{minipage} \\
\bottomrule
\end{tabular}
\label{tab:prompt_reward_generation}
\end{table*}

\begin{table*}[t!]
\centering
\caption{\textbf{Example of a first-round interaction between the user and the LLM using chain-of-thought prompting.}}
\begin{tabular}{l}
\toprule
\begin{minipage}[t]{\linewidth}
\texttt{\footnotesize \textbf{User:} Help me select the best flights for my trips. I have specific preferences for what I like and dislike in a flight, and these preferences remain the same. You need to figure out my preferences and select the best flights for me.\\
\\
First, infer my preferences by reasoning about each feature. For each feature, estimate the probability distribution of my preference across a 1-to-5 scale. For example, you might estimate a 30\% probability that I strongly prefer an earlier morning flight (scale 1), a 10\% probability that I prefer an earlier morning flight (scale 2), a 20\% probability that I have no strong preference (scale 3), and so on. Then, use these probabilities to determine the best flights for me.\\
\\
Use your best judgment if you are unsure. Do not say you need more information.\\
\\
Which flight is the best option? \\
\\
Flight 1: \\
departure time: 02:00 PM, duration: 2 hr 30 min, number of stops: 1, price: \$370 \\
Flight 2: \\
departure time: 10:00 PM, duration: 4 hr 24 min, number of stops: 0, price: \$730 \\
Flight 3: \\
departure time: 03:36 PM, duration: 16 hr 6 min, number of stops: 0, price: \$900 \\
\\
Let's think step by step. End your response with `The best option is Flight <your choice>.'. \\
\\
\textbf{Model:} Let's analyze each flight feature and estimate your preferences:\\
\\
...\\
<Model's reasoning chain> \\
...\\
\\
The best option is Flight 2.\\
\\
\textbf{User:} Your option Flight 2 is correct.} \\
\end{minipage} \\
\bottomrule
\end{tabular}
\label{tab:prompt_cot}
\end{table*}

\begin{table*}[t!]
\centering
\caption{\textbf{Example of a first-round interaction between the user and the LLM with verbalized reward posterior distribution.}}
\begin{tabular}{l}
\toprule
\begin{minipage}[t]{\linewidth}
\texttt{\footnotesize \textbf{User:} Help me select the best flights for my trips. I have specific preferences for what I like and dislike in a flight, and  these preferences remain the same. You need to figure out my preferences and select the best flights for me. Use your best judgment if you are unsure. Do not say you need more information. \\
\\
\textbf{Model:} Based on the current information, the probabilities for each preference scale across all features are:\\
\\
The probabilities for each scale of your preference for departure time are:\\
\\
- 1: I strongly prefer an earlier morning departure time, 0.3\% \\
- 2: I prefer an earlier morning departure time, 1.0\% \\
- 3: I have no strong preference, 97.6\% \\
- 4: I prefer a later evening departure time, 1.1\% \\
- 5: I strongly prefer a later evening departure time, 0.1\% \\
\\
The probabilities for each scale of your preference for duration are:\\
...\\
\\
\textbf{User:} Which flight is the best option? \\
\\
Flight 1: \\
departure time: 02:00 PM, duration: 2 hr 30 min, number of stops: 1, price: \$370 \\
Flight 2: \\
departure time: 10:00 PM, duration: 4 hr 24 min, number of stops: 0, price: \$730 \\
Flight 3: \\
departure time: 03:36 PM, duration: 16 hr 6 min, number of stops: 0, price: \$900 \\
\\
\textbf{Model:} The best option is Flight 2.\\
\\
\textbf{User:} Your option Flight 2 is incorrect. I prefer Flight 1. \\
\\
\textbf{Model:} Based on the current information, the probabilities for each preference scale across all features are:\\
 \\
The probabilities for each scale of your preference for departure time are: \\
 \\
- 1: I strongly prefer an earlier morning departure time, 0.1\% \\
- 2: I prefer an earlier morning departure time, 4.1\% \\
- 3: I have no strong preference, 2.5\% \\
- 4: I prefer a later evening departure time, 93.2\% \\
- 5: I strongly prefer a later evening departure time, 0.1\% \\
}
\end{minipage} \\
\bottomrule
\end{tabular}
\label{tab:prompt_posterior}
\end{table*}

\begin{table*}[t!]
\centering
\caption{\textbf{Example of a two-round interaction between the user and the LLM using the non-interactive setting.} The blue part highlights the difference compared to the interactive setting in Table~\ref{tab:prompt_interaction}.}
\begin{tabular}{p{\textwidth}}
\toprule
\begin{minipage}[t]{\linewidth}
\texttt{\footnotesize \textbf{User:} Help me select the best flights for my trips. I have specific preferences for what I like and dislike in a flight, and  these preferences remain the same. You need to figure out my preferences and select the best flights for me. Use your best judgment if you are unsure. Do not say you need more information. \\
\\
Which flight is the best option? \\
\\
Flight 1: \\
departure time: 02:00 PM, duration: 2 hr 30 min, number of stops: 1, price: \$370 \\
Flight 2: \\
departure time: 10:00 PM, duration: 4 hr 24 min, number of stops: 0, price: \$730 \\
Flight 3: \\
departure time: 03:36 PM, duration: 16 hr 6 min, number of stops: 0, price: \$900 \\
\\
\textcolor{blue}{\textbf{Model:} The best option is Flight 2.\\}
\\
\textcolor{blue}{\textbf{User:} Your option Flight 2 is correct.} \\
\\
Which flight is the best option? \\
\\
Flight 1: \\
departure time: 04:00 PM, duration: 18 hr 3 min, number of stops: 2, price: \$280 \\
Flight 2: \\ 
departure time: 10:48 AM, duration: 6 hr 21 min, number of stops: 1, price: \$370 \\
Flight 3: \\
departure time: 06:48 PM, duration: 10 hr 5 min, number of stops: 1, price: \$810 \\
\\
\textbf{Model:} The best option is Flight 2.\\
\\
\textbf{User:} Your option Flight 2 is correct.} \\
\end{minipage} \\
\bottomrule
\end{tabular}
\label{tab:prompt_icl}
\end{table*}

\begin{table*}[t!]
\centering
\caption{\textbf{Example of a first-round interaction between the user and the LLM on the hotel recommendation task.}}
\begin{tabular}{l}
\toprule
\begin{minipage}[t]{\linewidth}
\texttt{\footnotesize \textbf{User:} Help me select the best hotels for my trips. I have specific preferences for what I like and dislike in a hotel, and these preferences remain the same. You need to figure out my preferences and select the best hotels for me. Use your best judgment if you are unsure. Do not say you need more information. \\
\\
Which hotel is the best option?\\
\\
Hotel 1:\\
distance to downtown: 4 miles, price: \$550, rating: 3 stars, amenities: free parking and free breakfast \\
Hotel 2: \\
distance to downtown: 3 miles, price: \$820, rating: 2 stars, amenities: free parking, free breakfast, and pool \\
Hotel 3: \\
distance to downtown: 2.3 miles, price: \$370, rating: 1 stars, amenities: free parking \\
\\
\textbf{Model:} The best option is Hotel 3.\\
\\
\textbf{User:} Your option Hotel 3 is incorrect. I prefer Hotel 2.} \\
\end{minipage} \\
\bottomrule
\end{tabular}
\label{tab:prompt_hotel}
\end{table*}

\begin{table*}[t!]
\caption{\textbf{Example of a first-round interaction between the user and the LLM on the web shopping task.}}
\begin{tabular}{l}
\toprule
\begin{minipage}[t]{\linewidth}
\texttt{\footnotesize \textbf{User:} Help me select the best product. I have specific preferences for what I like and dislike in a product, and these preferences remain the same. You need to figure out my preferences and select the best products for me. Use your best judgment if you are unsure. Do not say you need more information. \\
\\
Which product is the best option? \\
 \\
Product 1: \\
Title: Chic D Independence Day Table Runner 72 Inches Long, Gnome Cotton Linen Spring Table Cloth Runners for Wedding Party Dinning Coffee Holiday, Farmhouse Style, USA Flag \\
Description: \\
- 14x72inch Dining Table Runner Size: 36x183cm, fits round, square or rectangular tables that seat 4 to 8 people. \\
- High Quality Cotton Linen Table Runner: 30\%cotton, 70\%burlap. triangle hem, wrinkle-free, no fade. easy to maintain and reusable. \\
- Our elegant table runner is perfect for holidays, catering, birthday party, dinning, brunches, potlucks, BBQ's, buffets, garden, bridal party etc \\
- Decorative Table Runner can be used as decor on kitchen tables; dining tables; entry tables; desks \& more \\
- Custom table runner, u also can personalized your tabletop decoration \\
Color: black white \\
Size: 13x108inch \\
 \\
Product 2: \\
Title: Ambesonne Orange Mandala Coffee Table, Pastel Colored Flourishes and Dark Toned Details Spring Bloom, Acrylic Glass Center Table with Wooden Frame for Offices Dorms, Small, Peach Dark Brown Cream \\
Description: \\
- Measurements - 24" Long x 18" Wide x 15" High coffee table with patterns that will add vitality to your living room. \\
- Made from - High Quality Beech Wooden Frame and Acrylic Glass Table Top. Durable and Elegant! \\
- Features- Light weight and easy to move around. You can place anywhere and enjoy its beauty! \\
- Easy To Use- It is easy to assembly with numbered parts and clear instructions. \\
- Versatile- Small coffee table, to keep your necessities within reach. Practical item for your living room! \\
Color: blue purple \\
Size: large \\
\\
Product 3: \\
Title: White Round Dining Table and 4 Chairs,Mid-Century Modern Coffee Table Round Kitchen  Table Small Office Table for Office Kitchen Dining Room (Round Table+4 Pale Grey Chairs) \\
Description: \\
- Dining Table and Chair Size:The dining table size is 35.4*35.4*29.5 inch, dining chairs is  36*17.5 inch.Our table and chairs suitable for 4-6 people. This glass table can also be used  as a conference table, coffee table \\
- Glass Dining Table and Chair Material:The table top is made of clear glass and equipped with  solid wood metal legs, with a modern design style and elegant appearance.The Chair is made of  velvet and solid wood metal cushions, the dining table and chair is an essential furniture \\
Size: round table with wood legs \\
 \\
\textbf{Model:} The best option is Product 3. \\
 \\
\textbf{User:} Your option Product 3 is incorrect. I prefer Product 2. \\
}
\end{minipage} \\
\bottomrule
\end{tabular}
\label{tab:prompt_webshop_verbose}
\end{table*}

\end{document}